\documentclass{article}
\usepackage{graphicx} 

\usepackage[dvipsnames]{xcolor}

\usepackage{parskip}
\usepackage{graphicx}
\usepackage{hyperref}
\usepackage[shortlabels]{enumitem}

\usepackage{algorithm}
\usepackage[noend]{algorithmic}
\usepackage{amsmath,amssymb,amsthm}
\usepackage[a4paper, total={6in, 8in}]{geometry}

\usepackage[authoryear,round]{natbib}

\newcommand{\rhs}{right hand side}
\newcommand{\lhs}{left hand side}
\newcommand{\T}{^{\top}}
\newcommand{\inv}{^{-1}}
\newcommand{\iid}{i.i.d.\@}
\newcommand{\st}{s.t.\@}

\newcommand{\infnorm}[1]{\|#1\|_{\infty}}

\newcommand{\onenorm}[1]{\|#1\|_{1}}
\newcommand{\twonorm}[1]{\|#1\|_{2}}
\newcommand{\pnorm}[2]{\|#1\|_{#2}}
\newcommand{\opnorm}[1]{\|#1\|_{op}}
\newcommand{\frobnorm}[1]{\|#1\|_{F}}
\newcommand{\abs}[1]{|#1|}

\newcommand{\Twonorm}[1]{\left\|#1\right\|_2}
\newcommand{\Infnorm}[1]{\left\|#1\right\|_{\infty}}

\newcommand{\Abs}[1]{\left|#1\right|}

\newcommand{\E}{\mathbb{E}}
\newcommand{\var}[2][]{\operatorname{Var}_{#1}(#2)}
\newcommand{\Var}[2][]{\operatorname{Var}_{#1}\Parent{#2}}
\newcommand{\cov}[2][]{\operatorname{Cov}_{#1}(#2)}
\newcommand{\Cov}[2][]{\operatorname{Cov}_{#1}\Parent{#2}}

\newcommand{\ent}[2][]{\operatorname{Ent}_{#1}(#2)}
\newcommand{\indicator}[1]{\mathbf{1}{\{#1\}}}
 
\newcommand{\LSI}[1]{{\operatorname{LSI}(#1)}}

\renewcommand{\P}{\mathbb{P}}

\newcommand{\R}{\mathbb{R}}
\newcommand{\N}{\mathbb{N}}
\newcommand{\Z}{\mathbb{Z}}

\newcommand{\diagonal}[1]{\operatorname{diag}(#1)}

\newcommand{\trace}{\operatorname{tr}}
\newcommand{\one}{\mathbf{1}}
\newcommand{\proj}[2]{\Pi_{#1}(#2)}

\newcommand{\parent}[1]{(#1)}
\newcommand{\bracket}[1]{[#1]}
\newcommand{\curly}[1]{\{{#1}\}}

\newcommand{\Parent}[1]{\left(#1\right)}
\newcommand{\Bracket}[1]{\left[#1\right]}
\newcommand{\Curly}[1]{\left\{{#1}\right\}}

\newcommand{\DP}{DP}
\newcommand{\LDP}{LDP}
\newcommand{\uLDP}{uLDP}
\newcommand{\uDP}{uDP}

\theoremstyle{plain}

\newtheorem{definition}{Definition}[section]
\newtheorem{lemma}[definition]{Lemma}
\newtheorem{theorem}[definition]{Theorem}
\newtheorem{corollary}[definition]{Corollary}
\newtheorem{assumption}[definition]{Assumption}
\newtheorem{remark}[definition]{Remark}

\theoremstyle{definition} 

\newtheorem{example}[definition]{Example}

\title{Differential Privacy with Dependent Data}
\author{
  Valentin Roth\thanks{Institute of Science and Technology Austria}
  \and
  Marco Avella-Medina\thanks{Department of Statistics, Columbia University}
}

\date{November 2025}

\begin{document}

\maketitle

\begin{abstract}

    Dependent data underlies many statistical studies in the social and health sciences, which often involve sensitive or private information. Differential privacy (DP) and in particular \textit{user-level} DP provide a natural formalization of privacy requirements for processing dependent data where each individual provides multiple observations to the dataset. However, dependence introduced, e.g., through repeated measurements challenges the existing statistical theory under DP-constraints. In \iid{} settings, noisy Winsorized mean estimators have been shown to be minimax optimal for standard (\textit{item-level}) and \textit{user-level} DP estimation of a mean $\mu \in \R^d$. Yet, their behavior on potentially dependent observations has not previously been studied. We fill this gap and show that Winsorized mean estimators can also be used under dependence for bounded and unbounded data, and can lead to asymptotic and finite sample guarantees that resemble their \iid{} counterparts under a weak notion of dependence. For this, we formalize dependence via log-Sobolev inequalities on the joint distribution of observations. This  enables us to adapt the stable histogram by \citet{karwa:vadhan2018} to a non-\iid{} setting, which we then use to estimate the private projection intervals of the Winsorized estimator. The resulting guarantees for our item-level mean estimator extend to \textit{user-level} mean estimation and transfer to the local model via a randomized response histogram. Using the mean estimators as building blocks, we provide extensions to random effects models, longitudinal linear regression and nonparametric regression. Therefore, our work constitutes a first step towards a systematic study of DP for dependent data.
    
\end{abstract}

\section{Introduction}

Differential privacy (DP) tools have been deployed at scale in numerous applications in industry and government agencies \citep{erlingsson:pihur:korolova2014, ding:kulkarni:yekhanin2017, tangetal2017, garfinkel:abowd:martindale2019}. The standard DP framework seeks to release statistics while protecting single data points as it assumes every individual contributes a single data point to a dataset of size $n$. We will follow the literature and call this standard approach \textit{item-level} DP. In this framework, privacy is achieved by releasing randomized outputs that are calibrated in a way which guarantees that the effect that any single data point can have in the computation of this output is hidden by the randomization \citep{dwork:mcsherry:nissim:smith2006,dwork:roth2014}.

A recent line of work called \textit{user-level} differential privacy (uDP) has studied the setting where each person/user contributes multiple observations to a dataset \citep{liu:theertha:yu:kumar:riley2020,levy:sun:amin:kale:kulesza:mohri:suresh2021,narayanan:mirrokni:esfandiari2022,acharya:liu:sun2023,bassily:sun2023,ghazi:kamath:kumar:manurangsi:meka:zhang2023,asi:liu2023}. For simplicity, we assume that every user contributes $T$ data points. Naturally, the goal now is to protect all $T$ points that a user contributes to the dataset. The challenge is that standard DP techniques protect only individual data points and either yield privacy guarantees that degrade as the users contribute more points or get the right privacy guarantee by adding too much noise if the algorithms are built using the group
property of DP \citep[Theorem 2.2]{dwork:roth2014}.

In this work, we introduce DP tools for dependent data and are particularly interested in longitudinal data, also commonly called panel data, consisting of multiple dependent observations collected from the same individuals over some time frame. Longitudinal data are immensely important in the social sciences and in medical applications \citep{diggle2002,baltagi2008,fitzmaurice:laird:ware2012,hsiao2022}. While user-level DP seems to be the natural definition of privacy for longitudinal data, this connection has not been exploited in the literature. In fact, there is relatively little work on modeling dependent data under differential privacy constraints. 

We were mainly inspired by \citet{karwa:vadhan2018}, as their algorithm serves as building block for all our procedures, as well as recent work in the emerging field of uDP, in particular \citet{levy:sun:amin:kale:kulesza:mohri:suresh2021} and \citet{kent:berrett:yu2024}.

Our main contributions can be summarized in the following key points:

\begin{enumerate}[(a)]

    \item DP for dependent and unbounded data: 
    our work seems to be the first to study multiple DP algorithms on dependent observations. In particular, we introduce \textit{log-Sobolev dependence}, a notion of dependence using log-Sobolev inequalities on the joint distribution of observations that replaces typical \iid{} assumptions. The building block of our methodology is a Winsorized mean estimation algorithm inspired by the one of \citet{karwa:vadhan2018}, developed for Gaussian \iid{} data. While additionally handling dependence, our algorithm inherits two appealing features from their work: it enable estimation with unbounded observations and does not require any knowledge about the unknown mean.

    \item DP for longitudinal data:
    we introduce various algorithms for uDP estimation for dependent and in particular longitudinal data, allowing for dependence across both users and time. This is a significant departure from the overwhelming majority of the theoretical analysis of DP algorithms which are designed and analyzed leveraging \iid{} assumptions, even in uDP settings \citep{levy:sun:amin:kale:kulesza:mohri:suresh2021, kent:berrett:yu2024, agarwal:kamath:majid:mouzakis:silver:ullman2025}. Our algorithms are shown to lead to optimal finite-sample error rates under log-Sobolev dependence. Thereby, the condition proves to be general enough to cover a wide range of interesting statistical models that had not been studied previously in the literature, including nonparametric regression, simple random effects models and linear regression for longitudinal data with dependent errors. 

    \item Histogram learning: 
    we extend the analysis of the histogram estimator by \citet{vadhan2017} to log-Sobolev dependent data using a Dvoretzky-Kiefer-Wolfowitz-type inequality by \citet{bobkov:götze2010}. This is a critical intermediate result in the construction of our mean estimation, because the private histogram is called as a first step when crudely estimating the midpoint of a private projection interval of length $O(\log n)$ that is used by our Winsorized mean estimator. This slight adjustment significantly extends the scope of the techniques introduced in \citet{vadhan2017} beyond \iid{} Gaussians and enables dependence even in the item-level setting.
    
    \item Item-level and user-level DP:
    a conceptually interesting contribution is to highlight how all the existing uDP algorithms are built on a known item-level DP algorithm. This connection allows us to give more perspective on the existing in-expectation minimax DP and uDP lower bounds. To do so, we provide an in-expectation analysis for projection estimators {\`a} la Karwa-Vadhan that seems to be novel even for \iid{} Gaussian data. The bound suggests that an impossibility result in uDP, saying that learning is impossible if only the number of observations per users $T \to \infty$ \citep[Theorem 8]{levy:sun:amin:kale:kulesza:mohri:suresh2021}, is connected to the impossibility of learning in the item-level setting when the variance tends to $0$ too quickly. We thereby identify a disconnect between existing item- and user-level lower bounds \citep{cai:wang:zhang2021, levy:sun:amin:kale:kulesza:mohri:suresh2021}. 

    \item Local DP: 
    while the focus of our presentation is on the central model of DP that assumes the existence of a trusted data curator, we extend all our results to the local model of DP where the central server is not trusted and the privacy mechanism is enforced in the data collection phase \citep{kasiviswanathan:lee:nissim:raskhodnikova:smith2011,duchi:jordan:wainwright2018}. In particular, we show how replacing the histogram in our central DP algorithms with a local analog leads to near-optimal estimators under the local model while keeping all the desirable properties that we highlighted in the previous points. This means, we can again allow for unbounded observations, dependence between users and among their observations in a user-level local DP (uLDP) framework.

\end{enumerate}

\subsection{Related Work}

Private estimation of location parameters like mean and median is a core statistical question frequently investigated in the \DP{} literature. The first DP mean estimator can be traced back to an application of noisy sums in \citet{dwork:mcsherry:nissim:smith2006} and asymptotic guarantees for private trimmed mean and median estimators were obtained in \citet{dwork:lei2009}. \citet{smith2011} seem to have been the first work to propose and study the asymptotic properties of a noisy two-step Winsorized mean estimator similar to the one we consider, i.e., one that first crudely estimates the mean, projects the data onto an interval around this estimate and then adds noise for privacy calibrated to the resulting finite sensitivity. Lower-bounds for mean estimation were derived in \citet{bun:ullman:vadhan2013,steinke:ullman2017,foygel:duchi214,steinke:ullman2017,bun:steinke2019,cai:wang:zhang2021}. In particular, \citet{cai:wang:zhang2021} provided in-expectation minimax upper and lower bounds matching the rate due to \citet{steinke:ullman2017}, for a Winsorized mean estimator lacking a data-driven projection interval. 
In contrast, \citet{karwa:vadhan2018} started a line of work on in-probability guarantees for Winsorized mean estimators that neither impose boundedness assumptions on observations nor on the mean. Their algorithm is based on a $(\varepsilon, \delta)$-\DP{} ``stable'' histogram estimator that is used to find a private confidence interval for the mean of univariate Gaussians. Subsequently, \citet{kamath:li:singhal:ullman2019} leveraged this approach to multivariate Gaussians with unknown covariance matrix and \citet{kamath:singhal:ullman2020} extended this approach to heavy-tailed distributions. 

We note that there are also numerous other works studying the problem of private mean and median estimation, in particular aiming to avoid assuming a bounded sample space, and in many cases making connections to robustness in the statistics community
\citep{avellamedina:brunel2020,avellamedina2020,avellamedina2021,avellamedina:bradshaw:loh2023,li:berrett:yu2023,yu:zhao:zhou2024,ramsay:jagannath:chenouri2022} as well as in computer science venues \citep{tzamosetal2020,liu:kong:kakade:oh2021,liu:kong:oh2022,hopkins:kamath:majid:narayanan2023,alabi:kothari:tankala:venkat:zhang2023,chhor:sentenac2023}. These constructions are not as closely related to our work.

User-level \DP{} was first introduced by \citet{mcmahan:ramage:talwar:zhang2017} to ensure privacy in the context of federated training of language models. Subsequently, the definition spread to other work in this context (\citet{wang:song:zhan:yang:wang:qi2019, augenstein:mcmahan:ramage:ramaswamy:kairouz:chen:mathews:aguera2020}) and user-level private data-aggregation in SQL-databases (\citet{wilson:zhang:lam:desfontaines:marengo:gipson2020}). Statistical results investigating the interplay between the number of users $n$, the number of observations per user $T$ and the magnitude of the noise necessary for user-level privacy in the context of empirical risk minimization were obtained in \citet{amin:kulesza:munoz:vassilvtiskii2019,epasto:mahdian:mao:mirrokni:ren2020,levy:sun:amin:kale:kulesza:mohri:suresh2021,narayanan:mirrokni:esfandiari2022,kent:berrett:yu2024}. Recently, \citet{agarwal:kamath:majid:mouzakis:silver:ullman2025} and \citet{zhao:lai:shen:li:wu:liu2024} derived the first user-level mean estimators for independent users with independent unbounded observations in the central model.

The problem of estimation under local differential privacy has been explored in a wide range of statistical problems including mean estimation, density estimation, nonparametric regression and hypothesis testing, just to cite a few \citet{wasserman:zhou2010, duchi:jordan:wainwright2018,gaboardi:rogers2018,butucea:dubois:kroll:saumard2020,berrett:butucea2020,berrett:gyorfi:walk2021,sart2023,pensia:asadi:jog:loh2024}. In particular user-level local differential privacy has recently been studied in \citet{girgis:data:diggavi2022,acharya:liu:sun2023,kent:berrett:yu2024}.

Finally, we note that there is some incipient literature on differential privacy with special temporal structure. This includes data problems such as change point detection \citep{zhang:krehbiel:tuo:mei:cummings2021,berrett:yu2021,li:berrett:yu2022} and multi-armed bandits \citep{mishra:thakurta2015,sajed:sheffet2019,hu:hedge2022,ou:avellamedina:cummings2024} which are typically studied under item-level DP and independent observation assumptions. There is also some initial work on time series with DP that again has tackled the problem from an item-level DP perspective. For example, \cite{zhang:khalili:mahdi:liu2022} considered a parametric approach with sequential AR$(1)$ data, \cite{amorino:gloter:halconruy2025} studied diffusion processes and 
\cite{kroll2024,butucea:klockman:krivobokova2025} considered the problem of nonparametric spectral density estimation.

\section{Preliminaries}
\label{sec:prelim}

We first present notation that will be used throughout this work and give basic background on differential privacy and log-Sobolev inequalities. At the end of this section we introduce the main dependence assumption that we use to model dependent and especially longitudinal data.

\subsection{Notation}

We define $[\mu \pm b] := [\mu-b,\mu+b]$, $[\mu \pm b) := [\mu-b, \mu+b)$ and $(\mu \pm b] := (\mu-b, \mu+b]$. Moreover, $[n]=\{1,2,\dots,n\}$, $\R_\geq := [0,\infty)$ and $\indicator{A}$ is the indicator function for a set $A$. For the minimum and maximum of $a,b \in \R$ we write $a \wedge b$ and $a \vee b$. When a sequence of random variables $X_1, X_1, X_2, \dots$ converges in probability to a random variable $X_*$, we write $X_n \overset{p}\to X_*$. When $\xi_1,\dots ,\xi_d \sim \operatorname{Lap}(l,s)$ \iid{}, we say $\Xi := (\xi_1,\dots ,\xi_d) \sim \operatorname{Lap}(l\one_d, sI_d)$. Here, let $\one_d$ be the all ones vector in $\R^d$. For matrices $A,B \in \R^{d\times d}$ we denote their positive definite and positive semi-definite order as $A\prec B$ and $A\preceq B$ and refer to them as Loewner orderings. Further, $\opnorm{A}$ and $\frobnorm{A}$ are the operator and Frobenius norm of $A$. For a matrix $M = [M_1^\top,\dots,M_n^\top]^\top\in\R^{n\times d}$ we denote its column vectors by $M_{\cdot j}$ for $j\in [d]$. We write the $L_2$-projection of a vector $x \in \R^d$ onto a convex set $C$ as $\proj{C}{x} := \arg\min_{y \in R} \twonorm{x - y}$. For two sequences $(a_n)_{n\in\N}$ and $(b_n)_{n\in\N}$ we use the following order symbols: $b_n = O(a_n)$ or $a_n \lesssim b_n$ if $\exists C > 0$ and $n_0 \in \N$ such that $\forall n \geq n_0, |\frac{b_n}{a_n}| \leq C$; if this holds up to log-factors we write $\tilde O$ or $a_n \lessapprox b_n$; $b_n = o(a_n)$ if $\lim_{n \to \infty} |\frac{b_n}{a_n}| = 0$. We write $a_n\asymp b_n$ if $a_n\lesssim b_n$ and $b_n\lesssim a_n$. For $x,y\in\mathcal{X}^n$, their Hamming distance is $d_H(x,y) := \sum_{i=1}^n \indicator{x_i \neq y_i}$, i.e., the number of elements which differ. Throughout the text we let $\kappa, c > 0$ denote absolute constants.

\subsection{Differential Privacy}

There are several competing notions of \DP{} (See \citet{mironov2017, dong:roth:su2022, dwork:rothblum2016, bun:steinke2016, dwork:mcsherry:nissim:smith2006}), but we will work with the most common notion $(\varepsilon, \delta)$-DP introduced in \citet{dwork:mcsherry:nissim:smith2006}. In its definition below, we denote the Borel $\sigma$-algebra on $\R^p$ by $\mathcal B(\R^p)$.

\begin{definition}{ $(\varepsilon, \delta)$-Differential Privacy (Central Model).}
    Let $\varepsilon, \delta > 0$ and let $\mathcal A : \mathcal{X}^n \to \R^d$ be a randomized algorithm, mapping $x \in \mathcal{X}^n$ to a $\mathcal B(\R^d)$-measurable random variable $\mathcal A(x)$. Then, $\mathcal A$ is $(\varepsilon, \delta)$-\DP{} if for all $x,y \in \mathcal{X}^n$ \st{} $d_H(x,y) \leq 1$ we have
    \begin{align*}
        \P\Parent{\mathcal A(x) \in S} 
        \leq e^\varepsilon \cdot \P\Parent{\mathcal A(y) \in S} + \delta,
        \quad \text{for all $S \in \mathcal B(\R^p)$}. 
    \end{align*}
    Furthermore, we say that 
    \begin{enumerate}
        \setlength\itemsep{-0.3em}
        \item $\mathcal A$ is \textit{item-level}-\DP{} if $x_i, y_i \in \R^{1 \times d}$ for all $i \in [n]$,
        \item $\mathcal A$ is \textit{user-level}-\DP{} if $x_i, y_i \in \R^{T \times d}$ for all $i \in [n]$.
    \end{enumerate}
\end{definition}

\begin{remark}
    The parameters $(\varepsilon, \delta)$ control the noise in the randomized algorithm and are commonly referred to as privacy budget parameters. If an estimator is $(\varepsilon, \delta)$-user-level DP, we write $(\varepsilon, \delta)$-\uDP{}.
\end{remark}

The so-called Laplace mechanism described in Algorithm \ref{alg:lapmech} is one of the most common ways to randomize an estimator to make it $(\varepsilon,0)$-DP.

\begin{theorem}{(Theorem 3.6, \citet{dwork:roth2014}).}
    \label{thm:lapmech}
    Algorithm \ref{alg:lapmech} is $(\varepsilon, 0)$-\DP{}. 
\end{theorem}

\begin{algorithm}[H]
    \caption{LaplaceMechanism$(f, x, \varepsilon)$:}
    \label{alg:lapmech}
    \begin{algorithmic}[1]
        \REQUIRE $f : \mathcal X^n \to \R^d$, $x \in \mathcal{X}^n\subset\R^n$, $\varepsilon > 0$    \STATE $D \gets \Curly{(x,y) : x,y\in \mathcal X^n, d_H(x,y)\leq 1}$
        \STATE $\Delta \gets \max_{(x,y) \in D} \onenorm{f(x) - f(y)}$
        \RETURN $f(x) + \Xi$ where $\Xi \sim \operatorname{Lap}(0, \frac{\Delta}{\varepsilon} I_d)$
    \end{algorithmic}
\end{algorithm}

The Laplace mechanism above and further mechanisms presented in \citet{dwork:roth2014} are the foundation of \DP{}. They can be combined to form more complex DP algorithms by composing the output of simple DP algorithms. The effective privacy budget of an algorithm that outputs a composition of various DP algorithms can be quantified by the composition theorems below. 

\begin{theorem}{(Corollary 3.21, Corollary B.2, \citet{dwork:roth2014}).}
    \label{thm:compthms}
    Let $\mathcal A$ be a privacy mechanism that composes $d$ mechanisms that are each $(\varepsilon^\prime, \delta/d)$-\DP{}.
    \begin{itemize}[itemindent=4cm]
        \setlength\itemsep{-0.3em}
        \item[\textup{(Basic Composition)}] If $\varepsilon, \delta > 0$ and $\varepsilon^\prime = \varepsilon/d$, then $\mathcal A$ is $(\varepsilon, \delta)$-DP.
        \item[\textup{(Advanced Composition)}] If $\varepsilon, \delta, \varrho \in (0,1)$ and $\varepsilon^\prime = \varepsilon /\sqrt{8d\log(1/\varrho)}$, then $\mathcal A$ is $(\varepsilon, \delta + \varrho)$-DP. 
    \end{itemize}
\end{theorem}

\subsection{Log-Sobolev Inequalities}

Our work uses log-Sobolev inequalities to control dependence along the columns of the data matrix $X^n \in \R^{n \times d}$. Log-Sobolev inequalities are functional inequalities extensively studied in the fields of optimal transport, Markov diffusions and concentration of measure (cf.\ \citet{villani2009, bakry:gentil:ledoux2014, ledoux2001, boucheron:lugosi:massart2013}). Log-Sobolev inequalities impose sub-Gaussian tail-decay and thus imply concentration as stated in Theorem \ref{thm:logsoblipconc}. Moreover, a log-Sobolev inequality controls the variance of any test function $f : \R^d \to \R$, because it implies a Poincar{\'e} inequality (Proposition 5.1.3, \citet{bakry:gentil:ledoux2014}). To be more precise in the treatment of log-Sobolev inequalities to follow, we define them below. 

\begin{definition}{(Log-Sobolev Inequality).}
    Let $\nu$ be a probability measure on $\R^d$. For an integrable function $f : \R^d \to \R_\geq$ \st{} $\int_{\R^d} f\abs{\log(f)}d\nu < \infty$ define the entropy of $f$ as the following functional:
    \begin{align*}
        \ent[\nu]{f} := \int_{\R^d} f \log(f) d\nu - \Parent{\int_{\R^d} f d\nu}\log\Parent{\int_{\R^d} fd\nu}. 
    \end{align*}
    Then, the probability measure $\nu$ fulfills a log-Sobolev inequality with constant $\rho > 0$ and we write $\nu \in \LSI{\rho}$ if for all $f : \R^d \to \R$ \st{} $\ent[\nu]{f^2} < \infty$ and $\nabla f \in L^2(\nu)$ we have
    \begin{align*}
        \ent[\nu]{f^2} \leq 2\rho \cdot \E[\twonorm{\nabla f(X)}^2]. 
    \end{align*}

\end{definition}

A generic approach to establish a log-Sobolev inequality is through its relation to the curvature of the density of a probability measure with respect to the Lebesgue measure. Specifically, strong log-concavity immediately implies a log-Sobolev inequality through the Bakry-Emery Criterion that can be found in Appendix \ref{app:aux}; see Theorem \ref{thm:bakryemery}. 

The following Lipschitz concentration inequality will be the main probabilistic workhorse for the derivation of our algorithms under dependence.

\begin{theorem}{(Log-Sobolev Lipschitz Concentration; Theorem 5.3, \citet{ledoux2001}).}
    \label{thm:logsoblipconc}
    Let the probability measure $\mathcal \nu$ on $\R^d$ have log-Sobolev constant $\rho$. Further, let $f : \R^n \to \R$ be an $L$-Lipschitz function. Then, when $X \sim \nu$ and $t \geq 0$ we have: 
    \begin{align*}
        \P\Parent{|f(X) - \E[f(X)]| \geq t}
        \leq 2\exp\Parent{-\frac{t^2}{2\rho L^2}}. 
    \end{align*}
\end{theorem}

\subsection{Log-Sobolev Dependence}

To the best of our knowledge, all the existing theoretical analysis of item-level differentially private estimators has relied on simplifying \iid{} assumptions similar or stronger to the one we state next.

\begin{assumption}
    \label{ass:iiddata}
    The matrix $X^n=(X_1^\top,\dots,X_n^\top)^\top\in \R^{n\times d}$ has i.i.d rows. 

\end{assumption}

Observe that Assumption \ref{ass:iiddata} does not impose a known bounded domain for the data as is frequently the case in the differential privacy literature. We will see that our estimators will allow for this relaxed \iid{} assumption and show that slight modifications of existing methods can be shown to work well for unbounded data domains. This is an interesting side product of our analysis. However, our main contribution will be to explicitly allow for dependent data. 

We introduce a \textit{log-Sobolev dependence} condition on the data matrix $X^n$ that imposes log-Sobolev inequality constraints to model the dependence of the entries of $X^n$. To capture the dependence between $n$ observations across $d$ dimensions we could require that $X^n \sim \mathcal D$ for some joint probability measure $\mathcal D$ that fulfills a log-Sobolev inequality. However, this would be an unnecessarily rigid condition for our purposes as we only need to control the variance and concentration of the column vectors $X^n_{\cdot 1},\dots,X^n_{\cdot d}$ for the design of our algorithms for histogram learning and mean estimation. This motivates our formulation of Assumption \ref{ass:logsobdep}.

\begin{assumption}{(Log-Sobolev dependence).}
    \label{ass:logsobdep}
    A matrix $X^n \in \R^{n \times d}$ fulfills this assumption with constants $(\rho, M)$ if for every column $X^n_{\cdot j}$: 
    \begin{enumerate}
        \setlength\itemsep{-0.3em}
        \item[\textup{(a)}] $X^n_{\cdot j} \sim \mathcal D_j$ on $\R^n$ where $\mathcal D_j$ fulfills $\LSI{\rho}$, 
        \item[\textup{(b)}] The average marginal cdf $\bar F_j(x) := \frac 1 n \sum_{i=1}^n \P(X_{ij} \leq x)$ is $M$-Lipschitz, 
    \end{enumerate}
\end{assumption}

\begin{remark}
    Lipschitzness of $\bar F_j$ is implied by $X_{ij}$ having marginal densities bounded by $M$. When we use constants $(\rho, \infty)$, we specify $M = \infty$ and mean that Condition (b) does not have to hold. 
\end{remark}

Assumption \ref{ass:logsobdep} allows us to use a Dvoretzky-Kiefer-Wolfowitz-type inequality in the analysis of the histogram estimator of Section \ref{sec:hist}. This is a key subroutine for our main mean estimators in Section \ref{sec:meanest}. Interested readers may find this inequality due to \citet{bobkov:götze2010} in Theorem \ref{thm:concavgmargCDF} in the appendix. Assumption \ref{ass:logsobdep}(b) is a technicality required in the proof of \citet{bobkov:götze2010}'s result. Besides its role in Theorem \ref{thm:concavgmargCDF}, we use Assumption \ref{ass:logsobdep}(a) as a model of dependence within the columns $X^n_{\cdot j} \in \R^n$. Corollary \ref{cor:varlogsobdes} shows that imposing it for all columns $X^n_{\cdot j}$ is enough to control each column's covariance and the variance of their mean. The proof can be found in Appendix \ref{app:relim}.

\begin{corollary}
    \label{cor:varlogsobdes}
    For $X^n \in \R^{n\times d}$ fulfilling Assumption \ref{ass:logsobdep} with constants $(\rho, \infty)$, 
    \begin{align*}
        \cov{X^n_{\cdot j}}
        \preceq \rho I_n, 
        \quad \text{and} \quad
        \E\bracket{\twonorm{\bar X_n - \E[\bar X_n]}^2}
        \leq \frac{d\rho}{n}, 
        \quad 
        \text{where $\bar X_n := \frac 1 n X\T\one_n$}. 
    \end{align*}
\end{corollary} 

When $\rho \asymp 1$, the variance of $\bar X_n$ qualitatively behaves as if the $X_1,\dots,X_n$ were independent and each had variance bounded by $\rho$. Since $\bar X_n$ is unbiased, its MSE vanishes as $n \to \infty$ as long as $\rho = o(n)$. The next lemma shows that Assumption \ref{ass:logsobdep}(a) implies sub-Gaussian concentration of the mean $\bar X_n \in \R^d$ over the $n$ observations. The result is well know but we give a proof for completeness in Appendix \ref{app:relim}.

\begin{lemma}
    \label{lem:logsobconcmean}
    Let $X^n\in \R^{n \times d}$ fulfill Assumption \ref{ass:logsobdep} with constants $(\rho, \infty)$ and $\E[X_i] = \mu$. Then,
    \begin{align*}
        \P\Parent{\Twonorm{\bar X_n - \mu} \leq \sqrt{\frac{2d\rho \log(2d/\alpha)}{n}}} 
        \geq \P\Parent{\Infnorm{\bar X_n - \mu} \leq \sqrt{\frac{2\rho \log(2d/\alpha)}{n}}}
        \geq 1 - \alpha. 
    \end{align*}
\end{lemma}

We will repeatedly exploit the $L_\infty$-norm concentration of Lemma \ref{lem:logsobconcmean} to derive our multivariate mean estimators. This lemma shows that, for the empirical mean, Assumption \ref{ass:logsobdep} trades off dependence and concentration via the log-Sobolev constant $\rho$. If $\rho \asymp 1$, the concentration in Lemma \ref{lem:logsobconcmean} resembles that of \iid{} sub-Gaussian random variables. In turn, if $\rho \asymp n$ there is no effect of averaging and the mean $\bar X_n$ behaves like each single $X_{i}$. 

\section{Histogram Estimator}
\label{sec:hist}

We now turn to histogram learning which is one of the building blocks of our mean estimator. The main feature of our histogram estimator is that it allows for dependent observations with an unbounded domain. The estimator in Algorithm \ref{alg:stablehist} is a stable histogram that estimates the probability mass in possibly infinitely many non-overlapping bins $(B_k)_{k \in \Z}$. It was introduced by \citet{korolova:krishnaram:mishra:ntoulas2009} and first fully described by \citet{bun:nissim:stemmer2016}. To fix ideas, in later sections we will work with bins of the form $B_k=(2h k \pm h]$ for $k\in\mathbb{Z}$ and some given bin width parameter $h$. In the current section we allow for general non-overlapping bins\footnote{Note that \citet[Lemma 2.4]{aliakbarpour:silver:steinke:ullman2024} considered an alternative to stable histograms which relies on truncated Laplace noise. We do not explore this approach in this work.}.

\begin{algorithm}
    \caption{StableHistogram$(X^n, (B_k)_{k\in\Z}, \varepsilon, \delta)$:}
    \label{alg:stablehist}
    \begin{algorithmic}[1]
        \REQUIRE $X^n \in \R^n $, $\varepsilon, \delta > 0$
        \FORALL{$k \in \Z$}
            \STATE $\hat p_k \gets \frac 1 n \sum_{i=1}^n \indicator{X_i \in B_k}$
            \STATE $\xi_k \gets \xi$ where $\xi \sim \operatorname{Lap}(0, \frac{2}{\varepsilon n})$ \hfill
            \STATE $\tilde p_k \gets 
                \begin{cases}
                    0 & \text{if $\hat p_k = 0$} \\
                    0 & \text{if $\hat p_k + \xi_k < \frac{2}{\varepsilon n} \log(\frac{2}{\delta}) + \frac 1 n$} \\
                    \hat p_k + \xi_k, & \text{otherwise}
                \end{cases}$
        \ENDFOR
        \RETURN $\boldsymbol{\tilde p} := (\dots, \tilde{p}_{-1},\tilde{p}_0,\tilde p_1, \tilde p_2,\dots )$
    \end{algorithmic}
\end{algorithm}

\subsection{Privacy and Utility Guarantees}

We note that while the privacy of Algorithm \ref{alg:stablehist} is known in the literature, we provide a proof of this result in the continuous setting for completeness; see Lemma \ref{lem:privhistlearner}. Indeed, this case is not formally covered in Theorem 7.3.5 \citet{vadhan2017}. 

\citet{karwa:vadhan2018} seem to have been the first authors to use stable histograms for DP mean estimation. More specifically they use this idea to privately estimate the mean and variance of \iid{} Gaussian random variables. Their results rely on the following histogram utility guarantee.

\begin{lemma}{(Adjusted from Lemma 2.3, \citet{karwa:vadhan2018}).}
    \label{lem:utilhistiid}
    Let $\eta, \varepsilon, \delta > 0$. Further, let $X^n \in \R^n$ have \iid{} entries, a $(B_k)_{k \in \Z}$ be a collection of disjoint bins covering $\R$, and $p_k := \P(X_1 \in B_k)$. Then, the output $\boldsymbol{\tilde p}$ of Algorithm \ref{alg:stablehist} satisfies 
    \begin{align*}
        \P\Parent{\max_{k \in \Z} |\tilde p_k - p_k| \leq \eta} \geq 1 - \Parent{1+\frac{e^\varepsilon}{\delta}} n \exp\Parent{-\frac{\varepsilon n\eta}{4}} - \exp\Parent{-\frac{n\eta^2}{2}}. 
    \end{align*}
\end{lemma}

Our analysis of the stability-based histogram estimator generalizes the one by \citet{karwa:vadhan2018} to dependent observations. Concretely, we extend their Lemma 2.3 for observations $X^n \in \R^n$ whose joint distribution fulfills a log-Sobolev inequality following essentially their proof strategy. This leads to Lemma \ref{lem:utilhistlogsob} below that is proven in Appendix \ref{app:hist}. 

\begin{lemma}
    \label{lem:utilhistlogsob}
    Let $\eta, \varepsilon, \delta > 0$ and the random matrix $X^n \in \R^{n}$ fulfill Assumption \ref{ass:logsobdep} with constants $(\rho, M)$. Let $(B_k)_{k \in \Z}$ be a collection of disjoint bins covering $\R$ and $p_k := \frac 1 n \sum_{i=1}^n \P(X_i \in B_k)$. Then, the output $\boldsymbol{\tilde p}$ of Algorithm \ref{alg:stablehist} satisfies
    \begin{align*}
        \P\Parent{\max_{k \in \Z} |\tilde p_k - p_k| \leq \eta} \geq 1 - \Parent{1+\frac{e^\varepsilon}{\delta}} n \exp\Parent{-\frac{\varepsilon n\eta}{4}} - \frac{16}{\eta} \exp\Parent{-\frac{1}{864} \frac{n\eta^3}{\rho M^2}}. 
    \end{align*}
\end{lemma}

The critical difference between the proof of Lemma \ref{lem:utilhistiid} and \ref{lem:utilhistlogsob} is that we replace the Dvoretsky-Kiefer-Wolfowitz (DKW) inequality used by \citet{karwa:vadhan2018} with the DKW-type inequality obtained by \citet{bobkov:götze2010} which allows for dependent observations. We restate the DKW-type inequality of \citet{bobkov:götze2010} in Theorem \ref{thm:concavgmargCDF}. It ensures ensures that $\hat p_k$ approximates well $p_k$ uniformly over all bins. When computing the private estimates $\tilde p_k$ from $\hat p_k$, the stability mechanism accumulates at most $n$ errors of the form $\abs{\tilde p_k - \hat p_k}$ that have fast exponential tail-decay. We note that the deviation in Lemma \ref{lem:utilhistlogsob} suggests a slower estimation error rate of $O(n^{-1/3})$ which can be sub-optimal in some dependent settings. However, this slower error will suffice for our purposes in the mean estimation algorithm described in the next section.

\subsection{Finding Private Data-Driven Projection Intervals}

Our main application of the histogram learning algorithm is to find data-driven projection intervals that will be used by our main mean estimator. The idea is to find a slowly diverging interval that contains all the observations with high-probability. This follows the blueprint of \citet{karwa:vadhan2018}. The projection interval will be the bin containing most of the mass and its two neighboring bins. This simple algorithm fits within a broader family of private midpoint algorithms used in the central and local item- and user-level \DP{} mean estimation literature \citep{smith2011,kamath:singhal:ullman2020,levy:sun:amin:kale:kulesza:mohri:suresh2021,kent:berrett:yu2024,agarwal:kamath:majid:mouzakis:silver:ullman2025}.

\begin{algorithm}
    \caption{ProjectionInterval$(X^n, \tau, \varepsilon, \delta)$:}
    \label{alg:privrangestabhist}
    \begin{algorithmic}[1]
        \REQUIRE $X^n \in \R^n$, $\tau, \varepsilon, \delta > 0$
        \STATE $B_k \gets (2\tau k \pm \tau]$ \hfill $\forall k \in \Z$
        \STATE $\boldsymbol{\tilde p} \gets \operatorname{StableHistogram} (X^n, (B_k)_{k\in \Z}, \varepsilon, \delta)$ \hfill $\boldsymbol{\tilde p} = (\dots,\tilde p_{-1},\tilde p_0,\tilde p_1, \tilde p_2,\dots )$
        \STATE $\hat k \gets \arg\max_{k \in \Z} \tilde p_k$
        \STATE $\hat m \gets 
            \begin{cases}
                2\tau\hat k & \text{if $\exists k \in \Z : \tilde p_k > 0$} \\
                0 & \text{otherwise}
            \end{cases}$
        \RETURN $\hat I := [\hat m \pm 3\tau]$
    \end{algorithmic}
\end{algorithm}

Our theoretical analysis will work with a notion of $(\tau, \gamma)$-concentration that is similar to the one introduced by \citet{levy:sun:amin:kale:kulesza:mohri:suresh2021} in the context of user-level privacy. While this is not really needed, it makes it easier to compare our results to previous work. It will also help highlighting how more observations per user can improve the estimation error even under the more stringent user-level DP setting. This will become very clear in the applications of Section \ref{sec:uDP}. 
 
\begin{assumption}{($(\tau, \gamma)_p$-Concentration, Adjusted from Definition 2, \citet{levy:sun:amin:kale:kulesza:mohri:suresh2021}).}
    \label{ass:taugammaconc}
    A matrix $X^n \in \R^{n \times d}$ is $(\tau, \gamma)_p^{x_0}$-concentrated or $(\tau, \gamma)_p$-concentrated around $x_0 \in \R^d$ if for all $i \in [n]$ with probability at least $1-\gamma/n$ we have $\pnorm{X_{i}\T - x_0}{p} \leq \tau$. We call $\tau$ the concentration radius.
\end{assumption}

Algorithm \ref{alg:privrangestabhist} returns an interval whose midpoint is the center of the bin of a stable histogram with the biggest estimated mass (cf.\ \citet{karwa:vadhan2018}, Algorithm 1). Our innovation is in the analysis of the algorithm which leads to Lemma \ref{lem:privmidpointstabhist} below, proven in Appendix \ref{app:privmidpoint}. The lemma generalizes Theorem 3.1 by \citet{karwa:vadhan2018} using $(\tau, \gamma)_\infty$-concentration and Lemma \ref{lem:utilhistlogsob} enabled by the DKW-type inequality due to \citet{bobkov:götze2010}. Most importantly, this allows for dependence between observations within the model of log-Sobolev dependence, but also enables non-identically distributed observations. Moreover, using the original DKW inequality, our proof strategy also generalizes the analysis of \citet{karwa:vadhan2018} for \iid{} Gaussians to general \iid{} and $(\tau, \gamma)_\infty$-concentrated observations. 

\begin{lemma}
    \label{lem:privmidpointstabhist}
    Algorithm \ref{alg:privrangestabhist} is $(\varepsilon, \delta)$-DP for $\varepsilon, \delta > 0$. Let $\gamma \in (0,1 \wedge \frac n 4)$, let $X^n \in \R^n$ be $(\tau, \gamma)_\infty$-concentrated around $x_0 \in \R$ and fulfill Assumption \ref{ass:iiddata} or \ref{ass:logsobdep} with constants $(\rho, M)$ \st{} $\rho M^2 \lesssim 1$. Let $\hat{m}$ denote the midpoint of the interval returned by Algorithm \ref{alg:privrangestabhist}. Then, with probability at least $1-O(\frac{n}{\delta} \cdot e^{-\kappa n \varepsilon} \vee e^{-\kappa^\prime n})$ for absolute constants $\kappa, \kappa^\prime > 0$, 
    \begin{align*}
        \hat m \in \Big[x_0 \pm 2\tau \Big]. 
    \end{align*}
\end{lemma}

\begin{remark}
    \label{rem:tauprime}
    We state Lemma \ref{lem:privmidpointstabhist} in terms of $(\tau, \gamma)^{x_0}_\infty$-concentration of $X^n \in \R^n$, because the analysis to follow heavily relies on this notion of concentration. However, note that the proof permits relaxing to $(\tau^\prime, \gamma)^{x_0}_\infty$-concentration of each single $X_i$, which can have considerably smaller $\tau^\prime$. This is exploited in our numerical experiments in Section \ref{sim:improving_constants} to sharpen rates.

\end{remark}

Since $(\tau, \gamma)_\infty$-concentration is an abstract notion of concentration, we instantiate Lemma \ref{lem:privmidpointstabhist} for $n$ \iid{} $\sigma^2$-sub-Gaussian random variables with mean $\mu \in \R$ to give an example. These are $(\tau, \gamma)^\mu_\infty$-concentrated with $\tau = \sqrt{2\sigma^2 \log(2n/\gamma)}$, yielding Corollary \ref{cor:privmedianiid} that is proven in Appendix \ref{app:privmidpoint}.

\begin{corollary}
    \label{cor:privmedianiid}
    Let $\varepsilon, \delta > 0$ and $X^n \in \R^n$ have \iid{} and sub-Gaussian entries with variance proxy $\sigma^2$ and mean $\mu \in \R$. Let $\hat{m}$ denote the midpoint of the interval returned by Algorithm \ref{alg:privrangestabhist}. Then, for $\gamma \in (0,1 \wedge \frac n 4)$ and with absolute constants $\kappa, \kappa^\prime > 0$ we have
    \begin{align*}
        \P\Parent{\hat m \in \Big[\mu \pm 2\sqrt{2\sigma^2\log(2n/\gamma)} \Big]}
        \geq 1 - O\Parent{\frac{n}{\delta} \cdot e^{-\kappa n \varepsilon} \vee e^{-\kappa^\prime n}}. 
    \end{align*}
    
\end{corollary}

The corollary exemplifies that we can estimate the true mean $\mu \in \R$ using the private midpoint $\hat m$ up to an error of size $\asymp \sigma$ with high probability, ignoring log-factors in $n, \gamma$.

\section{Item-Level DP Estimation with Dependent Data}
\label{sec:meanest}

We are ready to present our main algorithm. Our proposal follows a popular idea for private mean estimation: compute a Winsorized mean estimator and make it private with the Laplace mechanism. This approach is common both in the item- and user-level mean estimation literature (See, e.g., \citet{smith2011, karwa:vadhan2018, levy:sun:amin:kale:kulesza:mohri:suresh2021, kent:berrett:yu2024, agarwal:kamath:majid:mouzakis:silver:ullman2025}). Estimators within this framework restrict their global sensitivity by projecting observations onto an interval of fixed length. Given observations $X_1,\dots ,X_n \in \R$, we follow the one-dimensional item-level mean estimator construction of \citet{karwa:vadhan2018} given in three steps: 
\begin{enumerate}
    \setlength\itemsep{-0.3em}
    \item Compute a private crude mean estimator $\hat m$ and range $\hat I := [\hat m \pm \varsigma]$, 
    \item Project observations $X_i$ onto $\hat I$ and average them, 
    \item Privatize the projected average by adding Laplace noise with variance $\asymp \varsigma/(\varepsilon n)$. 
\end{enumerate}

In this construction, the choice of $\hat{m}$ and the parameter $\varsigma$ are critical for the theoretical analysis and empirical performance of the estimator. We will find both simultaneously using the private histogram-based projection estimator from the previous section. 
Item-level mean estimators in $\R^d$ can be obtained through coordinate-wise application of the one-dimensional estimator. Conceptually this is equivalent to first constructing an $L_\infty$-ball around a rough mean estimator $\hat m \in \R^d$, projecting observations $X_1,\dots ,X_n \in \R^d$ onto this hypercube, averaging them and adding isotropic Laplace noise in $\R^d$. The theory worked out in this section will be presented for dependent data in the item-level setting, but it will also be the cornerstone for all the user-level estimators in Section \ref{sec:uDP}. 

Our main theoretical results are the finite-sample error and in-expectation MSE bounds in Theorems \ref{thm:rMSEwinsmeanHD} and \ref{thm:inexpub}. These main results as well as all the guarantees derived for intermediate algorithms are obtained under log-Sobolev dependence. An interesting side product of our analysis is that we generalize some known results even in the \iid{} setting by allowing for both unbounded data domains and unbounded parameter spaces.

\subsection{Mean Estimators}

Algorithm \ref{alg:winsmean1D} presents our main mean estimation routine. It is a one-dimensional noisy Winsorized mean estimator that is shown to work well under log-Sobolev dependence. The estimator relies on projecting the data to the private range $\hat I := \bracket{\hat m \pm 3\tau}$ provided by Algorithm \ref{alg:privrangestabhist}. 

\begin{algorithm}
    \caption{WinsorizedMean1D($X, \tau, \varepsilon, \delta,\dots $)}
    \label{alg:winsmean1D}
    \begin{algorithmic}[1]
        \REQUIRE $X \in \R^n$, $\tau, \varepsilon, \delta > 0$
        \STATE $\hat I \gets \operatorname{ProjectionInterval}(X, \tau, \frac \varepsilon 2, \delta,\dots)$ where $\hat I = \bracket{\hat m \pm 3\tau}$ and $| \hat I| = 6\tau$
        \RETURN $\frac 1 n \sum_{i=1}^n \proj{\hat I}{X_i} + \xi$ with $\xi \sim \operatorname{Lap}(0, \frac{12\tau}{n\varepsilon})$ 
    \end{algorithmic}
\end{algorithm}

By construction, all observations $X_1,\dots ,X_n \in \R$ lie in the private projection interval $\hat I$ with high probability. This guarantees that with high probability the noisy Winsorized mean estimator behaves like a noisy empirical mean. Lemma \ref{lem:winsmean1D} formalizes this behavior and is proven in Appendix \ref{app:meanest}.

\begin{lemma}
    \label{lem:winsmean1D}
    Algorithm \ref{alg:winsmean1D} denoted by $\mathcal A$ is $(\varepsilon,\delta)$-\DP{} for $\varepsilon, \delta > 0$. Assume $X^n \in \R^n$ is $(\tau, \gamma)^{x_0}_\infty$-concentrated and make Assumption \ref{ass:iiddata} or \ref{ass:logsobdep} \st{} $\rho M^2 \lesssim 1$. Then, for absolute constants $\kappa, \kappa^\prime > 0$ the event $\mathcal E := \curly{\forall i \in [n] : \proj{\hat I}{X_i} = X_i}$ has probability at least $1 - \gamma - O(n/\delta \cdot e^{- \kappa n \varepsilon} \vee e^{-\kappa^\prime n})$ and 
    \begin{align*}
        \mathcal A(X^n) \cdot \indicator{\mathcal E} = \Parent{\bar X_n + \xi} \cdot \indicator{\mathcal E}, 
        \quad \text{where} \quad 
        \xi \sim \operatorname{Lap}\Parent{0, \frac{12\tau}{n\varepsilon}}. 
    \end{align*}
\end{lemma}

While Lemma \ref{lem:winsmean1D} does not itself provide a statistical error guarantee, it is helpful for deriving the finite-sample result of Subsection \ref{subsec:finitesample} as it lets us analyze Algorithm \ref{alg:winsmean1D} via the proxy $\bar X_n + \xi$. 

A multivariate version of Lemma \ref{lem:winsmean1D} is presented in Lemma \ref{lem:winsmeanHD} in Appendix \ref{app:meanest}. It is analogous to Theorem 2 by \citet{levy:sun:amin:kale:kulesza:mohri:suresh2021} and obtained through our multivariate mean estimator given in Algorithm \ref{alg:winsmeanHD}. The estimator is a simple coordinate-wise application of Algorithm \ref{alg:winsmean1D}. Therefore, the assumptions of $(\tau, \gamma)_\infty$-concentrated data $X^n \in \R^{n \times d}$ around a point $x_0 \in \R^d$ in combination with log-Sobolev dependence or \iid{} data are again natural in this context. Note that $(\tau_p, \gamma)_p$-concentration for all $p \in \N$ are equivalent in $\R$ but not in $\R^d$ since for a fixed $\gamma$ the order of $\tau_p$ as a function of $d$ might differ significantly (cf. \citet{levy:sun:amin:kale:kulesza:mohri:suresh2021, kent:berrett:yu2024}).

\begin{algorithm}[H]
    \caption{WinsorizedMeanHD($X^n, \tau, \varepsilon, \delta, \varrho, \dots $)} 
    \label{alg:winsmeanHD}
    \begin{algorithmic}[1]
        \REQUIRE $X^n \in \R^{n \times d}$, $\tau > 0$, $\delta, \varrho, \varepsilon \in (0,1)$
        \STATE $\varepsilon^\prime \gets {\varepsilon}/{\sqrt{8d\log(1/\varrho)}}$
        \STATE $\delta^\prime \gets \delta/d$
        \STATE $\bar X_j \gets \operatorname{WinsorizedMean1D}(X_{\cdot j}, \tau, \varepsilon^\prime, \delta^\prime,\dots )$ \hfill $\forall j \in [d]$
        \RETURN $(\bar X_1,\dots , \bar X_d)\T$
    \end{algorithmic}
\end{algorithm}

\subsection{Theoretical Guarantees}
\label{subsec:guarantees}

\subsubsection{High-Probability Bound}
\label{subsec:finitesample}

We translate the utility guarantee of Theorem \ref{lem:winsmeanHD} into a finite-sample upper bound on the MSE of the estimator in Algorithm \ref{alg:winsmeanHD}. The estimation error in Theorem \ref{thm:rMSEwinsmeanHD} below is quantified by the sum of two terms: the statistical error and the cost of privacy. The theorem is proven in Appendix \ref{app:finitesample}.

\begin{theorem}
    \label{thm:rMSEwinsmeanHD}
    Algorithm \ref{alg:winsmeanHD} denoted by $\mathcal A$ is $(\varepsilon,\delta+\varrho)$-DP for $\delta, \varrho, \varepsilon \in (0,1)$. Furthermore, let $\gamma \in (0, 1 \wedge \frac n 4)$ and $X^n \in \R^{n \times d}$ be $(\tau, \gamma)_\infty^\mu$-concentrated and fulfill Assumption \ref{ass:iiddata} or \ref{ass:logsobdep} \st{} $\rho M^2 \lesssim 1$. Then, for absolute constants $\kappa, \kappa^\prime > 0$, with probability at least $1-2d\gamma-O(d^2n/\delta \cdot e^{-\kappa n \varepsilon^\prime})$,
    \begin{align*}
        \Twonorm{\mathcal A(X^n) - \mu} 
        &\lesssim \Twonorm{\bar X_n - \mu} + \tau \sqrt{\frac{d^2\log(1/\varrho)\log(3/(d\gamma))^2}{n^2 \varepsilon^2}}.
    \end{align*}
\end{theorem}

As shown in Lemma \ref{lem:logsobconcmean}, under log-Sobolev dependence with constant $\rho$, the statistical rate $\twonorm{\bar X_n - \mu}$ is upper bounded by $\sqrt{2d\rho\log(2d/\alpha)/n}$. Thus, up to constants and logarithmic terms Theorem \ref{thm:rMSEwinsmeanHD} yields an MSE bound for mean estimation of order $d/n + d^2/(n\varepsilon)^2$ in probability. Expressing the theorem in terms of sample complexity bounds recovers \citet[Theorem 4.1]{kamath:li:singhal:ullman2019} in the \iid{} Gaussian setting when $X^n \sim \mathcal N(0,\sigma^2I_n)$. It is tempting to apply the integral identity of the expectation to translate an in-probability bound like Theorem \ref{thm:rMSEwinsmeanHD} into an in-expectation bound. However, this will not work becuase of the term $O(d^2n/\delta \cdot e^{- \kappa n \varepsilon^\prime})$ in the probability. Alternatively, splitting the expectation on the event in Theorem \ref{thm:rMSEwinsmeanHD} and its complement is equally troublesome, since we cannot easily control the MSE on the low probability event.

\subsubsection{In-Expectation Analysis}

While the main focus of this work is on non-asymptotic bounds like the finite-sample guarantee of Theorem \ref{thm:rMSEwinsmeanHD} and instantiations thereof in Section \ref{sec:uDP}, we also give an in-expectation MSE upper bound in Theorem \ref{thm:inexpub}. We do so to provide an analysis of our estimator that is comparable to the existing upper and in particular also lower bounds in the user-level DP literature like Corollary 5 in \citet{levy:sun:amin:kale:kulesza:mohri:suresh2021}, and Theorem 3.1 or Theorem 3.2 of \citet{cai:wang:zhang2021} in the item-level DP literature. This allows us to make direct comparisons to known results in the \iid{} setting. In particular, we can match the \iid{} rates under weak dependence as shown in Theorem \ref{thm:inexpub}. 

The procedure used for the in-expectation and consistency result in Theorem \ref{thm:inexpub} is again Algorithm \ref{alg:winsmeanHD}. However, we employ a sample splitting technique like in the proof of Theorem 6 by \citet{kent:berrett:yu2024}. For simplicity, we assume that we have $2n$ samples instead of just $n$ that are split into two independent halves $Z^n := [Z_1,\dots,Z_n]\T$ and $X^n := [X_1,\dots,X_n]\T$. We estimate the private midpoint $\hat m := \hat m(Z^n)$ with Algorithm \ref{alg:privrangestabhist} using only the first half and project the second half onto the hypercube $\hat I^d := \hat I_1 \times \dots \times \hat I_d$ where $\hat I_j = [\hat m_j \pm 3\tau]$ to then estimate the mean using $\proj{\hat I^d}{X_1},\dots,\proj{\hat I^d}{X_n}$. This lets us prove the statement conditional on $\hat I^d$. Through this conditioning, $\hat I^d$ is deterministic which makes $\proj{\hat I^d}{X_{i}}$ a contraction and significantly simplifies controlling the estimator's variance. More concretely, we propose the estimator:
\begin{equation}
    \label{eq:mean_data_split}
    \mathcal{A}_{\hat m(Z^n)}(X^n)=\frac{1}{n}\sum_{i=1}^n \proj{\hat I^d}{X_i}+\Xi, 
    \quad \text{where} \quad 
    \Xi\sim \operatorname{Lap}\Parent{0, \frac{12\tau}{\varepsilon n}I_d}. 
\end{equation}

\begin{theorem}
    \label{thm:inexpub}
    The estimator \eqref{eq:mean_data_split} is $(\varepsilon, \delta + \varrho)$-DP for $\varepsilon, \delta, \varrho \in (0,1)$. Further, let $X^n,Z^n \in \R^{n \times d}$ have rows with mean $\mu \in \R^d$. Make Assumption \ref{ass:iiddata} with $\rho$-sub-Gaussian entries or alternatively Assumption \ref{ass:logsobdep} \st{} $\rho M^2 \lesssim 1$. Then, choosing $\tau^2 \asymp \rho \log(4n)$ we have
    \begin{align*}
        \E[\twonorm{\mathcal A_{\hat m(Z^n)}(X^n) - \mu}^2]
        &\lesssim d\rho \log(4n) \Parent{\frac{1}{n} + \frac{d\log(1/\varrho)}{n^2\varepsilon^2} + \frac{dn}{\delta e^{\kappa n \varepsilon^\prime/2}}} + \frac{d\twonorm{\mu}^2}{\delta e^{\kappa n \varepsilon^\prime}}.
    \end{align*}
\end{theorem}

Theorem \ref{thm:inexpub} is proven in Appendix \ref{app:asymptotics}. In the \iid{} setting where the rows have  $\rho$-sub-Gaussian entries, the upper bound in Theorem \ref{thm:inexpub} recovers the error rates obtained in prior work such as Theorem 3.2 by \citet{cai:wang:zhang2021} or Corollary 5 by \citet{levy:sun:amin:kale:kulesza:mohri:suresh2021}. However, note that the construction of our estimator does not need to assume a known bound on $\twonorm{\mu}^2$. Unlike other results in the literature, $\twonorm{\mu}^2$ only shows up in the analysis of our estimator. 

\subsection{Minimax Optimality}

To assess the optimality of our mean estimators, we compare our upper bounds to the statistical minimax lower bound given in \citet{cai:wang:zhang2021}. Their results were obtained for \iid{} observations $X_1,\dots ,X_n \in \R^d$ each with \iid{} $\rho$-sub-Gaussian entries in the item-level DP setting. For sake of exposition, we restate their theorem below. We will see that this bound implies that our mean estimator is minimax optimal in the \iid{} setting of item-level \DP{}. 

\begin{theorem}{(Theorem 3.1, \citet{cai:wang:zhang2021}).}
    \label{thm:inexplb}
    Let $X^n \in \R^{n \times d}$ fulfill Assumption \ref{ass:iiddata}, have $\rho$-sub-Gaussian entries and rows with mean $\infnorm{\mu}^2 < \rho$. Call the class of such distributions $\mathcal P$. Let $\varepsilon \in (0,1)$ and $\delta \in ((ne^{\varepsilon n})\inv, n^{-(1+\omega)})$ for $\omega > 0$. Assume $\log(\delta)/\log(n)$ is non-increasing in $n$, and $d \gtrsim \log(1/\delta)$. Then, if $n\varepsilon \gtrsim \sqrt{d\log(1/\delta)}$ it holds that 
    \begin{align*}
        \inf_{\mathcal A \in \boldsymbol{\mathcal{A}}_{(\varepsilon, \delta)}^{\mathrm{item}} } \sup_{\nu \in \mathcal P} \E[\twonorm{\mathcal A(X^n) - \mu}^2] 
        \gtrsim d\rho \Parent{\frac 1 n + \frac{d \log(1/\delta)}{n^2\varepsilon^2}}, 
    \end{align*}
    where $\boldsymbol{\mathcal{A}}_{(\varepsilon, \delta)}^{\mathrm{item}}$ is the family of all $(\varepsilon,\delta)$-DP estimators. 
\end{theorem}

Under Assumption \ref{ass:iiddata} which subsumes Assumption \ref{ass:logsobdep} with \iid{} rows, the distributional conditions of Theorem \ref{thm:inexpub} are more general than those of Theorem \ref{thm:inexplb}, because the latter restricts the population mean to satisfy $\infnorm{\mu}^2 < \rho$. To see this, note that if we assume $\infnorm{\mu}^2 \lesssim \rho n e^{\kappa n \varepsilon^\prime/2}$ and $\delta \in (n\varepsilon^\prime/e^{\kappa n \varepsilon^\prime}, n\inv]$, the upper bound of Theorem \ref{thm:inexpub} will match the lower bound of Theorem \ref{thm:inexplb} up to logarithmic terms. This is slightly more restrictive than the condition on $\delta$ required in the lower bound, but given that our analysis does not need a known bound on the norm of $\mu$ and even allows the norm to grow exponentially in $n$, we consider this a minor inconvenience. For a formal statement of this discussion see Corollary \ref{cor:inexpminimax} in Appendix \ref{app:asymptotics}. 

\begin{remark}
    Interestingly, the rate in Theorem \ref{thm:inexpub} also matches the lower bound up to a $O(\log n)$-factor under log-Sobolev dependence of Assumption \ref{ass:logsobdep} as long as $\rho \asymp 1$. Therefore, in this weak dependence scenario, the rate of differentially private mean estimation in $n, d, \varepsilon, \delta$ is unchanged relative to \iid{} sub-Gaussian observations. However, as soon as $\rho \gtrsim n^{\omega}$ for $\omega > 0$, dependence decreases the rate in the upper bound and Theorem \ref{thm:inexpub} is not tight anymore. We leave deriving lower bounds under dependence incorporating the log-Sobolev constant $\rho$ as an \textup{open problem}. 
\end{remark}

\begin{remark}
    The lower bound in Theorem \ref{thm:inexplb} requires the condition $\infnorm{\mu}^2 < \rho$ which connects the mean and sub-Gaussian variance proxy $\rho$. In particular, the lower bound is not very interesting when $\rho \to 0$. In contrast, our upper bound suggests that the MSE does not vanish as $\rho \to 0$ but $n$ stays constant when we do not connect the norm of $\mu$ to $\rho$. Indeed, the fourth term in the upper bound of Theorem \ref{thm:inexpub} is exponentially small in $n$ but does not vanish when $\rho\to 0$. While this limit might not seem practically relevant in the item-level setting, it has important implications for user-level estimation related to the impossibility of learning if only the number of observations $T$ per user goes to infinity. See also Remark \ref{rem:imposibilityuDP}.
\end{remark}

\subsection{Extension to Nonparametric Regression}
\label{subsec:nonparametric}

Nonparametric regression aims to estimate a function $f : [0,1] \to \R$ from $n$ noisy observations of the form $Y_i = f(x_i) + \epsilon_i$ with $\epsilon^n=(\epsilon_1,\dots,\epsilon_n) \sim \mathcal N(0, \Sigma^n)$. Commonly, the function $f$ is assumed to be part of a family of functions $\mathcal F$ such as a Hilbert, Hölder, Sobolev or Besov space. While the $x_i$ may be random, we make a fixed design assumption and view $x_i$ as fixed points on a grid on $[0,1]$. 

We note that the problem of private nonparametric regression has been considered for \iid{} data in \citet{awan:reimherr:slavkovic2020,golowich2021,cai:wang:zhang2023} in the central \DP{} model and in \citet{berrett:gyorfi:walk2021,gyorfi:kroll2025} in the local DP model. We permit correlated observations and want to focus on their effect on estimation. For simplicity, we restrict ourselves to $\mathcal F$ being the space of bounded Lipschitz functions. This setting is formalized in Definition \ref{def:fixeddesregmod}. 

\begin{definition}{(Fixed Design Nonparametric Regression).}
    \label{def:fixeddesregmod}
    Suppose $f : [0,1] \to \R$ is $L_f$-Lipschitz and $\infnorm{f} < \infty$ and define the equidistant design points $x_i = i/n$ for $0 \leq i \leq n-1$. Let $\epsilon^n \sim \mathcal N(0, \Sigma^n)$ be noise with $\sigma^2_{\min} I_n \preceq \Sigma^n \preceq \sigma^2_{\max} I_n$ and $0 < \sigma^2_{\min} \leq \sigma^2_{\max}$. We define the nonparametric fixed design regression model through the observations $Y_i = f(x_i) + \epsilon_i$. 
\end{definition}

Given observations, the underlying function $f$ is commonly estimated using classical kernel-based estimators like the nearest neighbor, Priestley-Chao, Nadaraya-Watson, Gasser-Müller or local polynomial estimators. Alternatively, the task is translated into mean estimation using Fourier or wavelet transforms, or linear regression via smoothing splines (See \citet{tsybakov2008}). Nonparametric estimation under \DP{} mainly relies on such reformulations as can be seen, e.g., in Subsection 5.2.2 of \citet{duchi:jordan:wainwright2018}, \citet{cai:chakraborty:vuursteen2025} or Section 6 of \citet{cai:wang:zhang2023}. Here we show that, in our fixed design setting, we can construct optimal pointwise \DP{} estimators of $f$ building on the classical nonparametric Priestley-Chao regression estimator in Definition \ref{def:priestchaoest} that was introduced by \citet{priestley:chao1972}. 

\begin{definition}
    \label{def:priestchaoest}
    Let $x^n, Y^n \in \R^n$ contain design points and responses. Let $K : \R \to \R_\geq$ be a kernel and let $\mu_{1}(K) := \int |s|K(s)ds$. Suppose $K$ is $L_K$-Lipschitz, $\int_\R K(u)du = 1$ and $\mu_1(K), \infnorm{K} < \infty$. Let $b > 0$ be the bandwidth. Then, we define the Priestley-Chao estimator as 
    \begin{align*}
        \hat f_n(\cdot) 
        := \sum_{i=0}^{n-1} \frac{x_{i+1} - x_i}{b} K\Parent{\frac{\cdot-x_i}{b}} Y_i.
    \end{align*}
\end{definition}

Since the Priestley-Chao regression estimator with equidistant $x_i$ \st{} $x_{i+1} - x_i = 1/n$ is an average of $\hat f^n_i(\cdot) := Y_i\frac 1 b K(\frac{\cdot - x_i}{x})$, we can easily privatize it using our Algorithm \ref{alg:winsmean1D}. In order to apply Theorem \ref{thm:rMSEwinsmeanHD} to obtain a finite-sample generalization bound for the privatized estimator, we first need to establish $(\tau, \gamma)_\infty$-concentration of the non-private $\hat f^n_i(x)$ around $f(x)$ for a fixed $x$. We state this intermediate result formally in the following lemma. It is proven in Appendix \ref{app:nonparametricreg}. 

\begin{lemma}
    \label{lem:concpriestchaoest}
    Let $\hat f_n$ be the estimator of Definition \ref{def:priestchaoest} with a kernel that is a 1-sub-Gaussian density with $L_K, \infnorm{K} < \infty$. Assume the model of Definition \ref{def:fixeddesregmod} with constants $L_f$ and $\sigma^2_{\max}$. Then, for all $n \in \N$, bandwidths $b>0$ and $x \in [\zeta,1-\zeta]$ where $\zeta = b\sqrt{2\log(2/b)}$ with probability at least $1-\alpha$, 
    \begin{align*}
        \Abs{\hat f_n(x) - f(x)}
        \leq 
        b \Parent{L_f \mu_{1}(K) + \infnorm{f}} + \frac{L_f \infnorm{K} + \infnorm{f} L_K}{nb} + 
        \sqrt{\frac{2\sigma_{\max}^2 \infnorm{K} \log(2/\alpha)}{nb}}.
    \end{align*}
    Further, $\hat f^n(x) \in \R^n$ with $\hat f^n_i(x) := Y_i \frac{1}{b} K\left(\frac{x_i-x}{b}\right)$ is $(\tau, \gamma)_\infty$-concentrated around $f(x)$ where
    \begin{align*}
        \tau 
        = b \Parent{L_f \mu_{1}(K) + \infnorm{f}} + \frac{1}{b} \Parent{L_f \infnorm{K} + \infnorm{f} L_K + \sqrt{2 \sigma^2_{\max} \infnorm{K} \log(2n/\gamma)}}.
    \end{align*}
\end{lemma}

The proof idea behind Lemma \ref{lem:concpriestchaoest} is to decompose the mean absolute error into bias and mean absolute deviation. The bias is bounded deterministically leveraging the Lipschitzness of $f$, while the mean absolute deviation is controlled using the $L_2$-Lipschitzness of $\hat f_n$ in the errors $\epsilon^n \in \R^n$ and Gaussian-Lipschitz concentration. While such an approach is common when showing in-expectation bounds for kernel estimators, the literature contains only few finite-sample bounds for kernel estimators. The authors are not aware of another one for a Priestley-Chao regression estimator, in particular for correlated Gaussian noise. The closest results we found in the literature are the following: a finite-sample bound for a distance-based kernel regressor for estimating a functions first derivative \citet[Theorem 1]{kpotufe:boularias2012} and generalization bound for a Nadaraya-Watson estimator assuming a uniformly bounded function as we do, but also bounded noise \citep{srivastava:wang:hanasusanto:ho2021, wang:gradi:ho2024}. 

The following lemma shows that besides their concentration, the summands defining $\hat f^n(x)$ are log-Sobolev dependent. A proof is given in Appendix \ref{app:nonparametricreg}. 

\begin{lemma}
    \label{lem:logsobdeppriestchaoest}
    Let $\hat f_n$ be the estimator of Definition \ref{def:priestchaoest} with kernel $K$ on the model of Definition \ref{def:fixeddesregmod}. Then, $\hat f^n(x) \in \R^n$ with $\hat f^n_i(x) = Y_i \frac{1}{b} K\left(\frac{x_i-x}{b}\right)$ fulfills Assumption \ref{ass:logsobdep} with $\rho M^2 \lesssim \sigma^2_{\max}/\sigma^2_{min} \cdot \infnorm{K}$. 
\end{lemma}

Now, by Theorem \ref{thm:rMSEwinsmeanHD} under Assumption \ref{ass:logsobdep} we get the finite-sample generalization bound in Corollary \ref{cor:rMSEpriestchaoest} in whose display we suppress the constants $L_f, L_K, \infnorm{f}, \infnorm{K}$ and logarithmic terms for clarity. Corollary \ref{cor:rMSEpriestchaoest} is proven in Appendix \ref{app:nonparametricreg}. 

\begin{corollary}
    \label{cor:rMSEpriestchaoest}
    Let $\hat f^n(x) \in \R^n$ with $\hat f^n_i(x) = Y_i \frac{1}{b} K\left(\frac{x_i-x}{b}\right)$, and denote by $\mathcal A$ Algorithm \ref{alg:winsmeanHD} taking as input $\hat f^n(x)$ for $x\in [0,1]$. Then, $\mathcal A$ is $(\varepsilon, \delta)$-\DP{} for $\varepsilon, \delta \in (0,1)$. Furthermore, if $\gamma \in (0,1 \wedge\frac n 4)$, $\sigma^2_{\max} \gtrsim 1$ and $x \in [\zeta,1-\zeta]$ where $\zeta = b\sqrt{2\log(2/b)}$ and $b > 0$. Then, if $Y^n$ is as in Definition \ref{def:fixeddesregmod}, with probability at least $1-3\gamma-O(n/\delta \cdot e^{-\kappa n \varepsilon})$ where $\kappa > 0$ is an absolute constant,
    \begin{align*}
        \Abs{\mathcal A(\hat f^n(x)) - f(x)}
        &\lessapprox b + \sigma_{\max} \Parent{\sqrt{\frac{1}{nb}} + \sqrt{\frac{1}{n^2b^2\varepsilon^2}}}. 
    \end{align*}
    For an optimally chosen $b \asymp (\sigma^2_{\max}/n)^{1/3} \vee \sqrt{\sigma_{\max}/n\varepsilon}$ this becomes
    \begin{align*}
        \Abs{\mathcal A(\hat f^n(x)) - f(x)}
        &\lessapprox \Parent{\frac{\sigma^2_{\max}}{n}}^{1/3} \vee \sqrt{\frac{\sigma_{\max}}{n\varepsilon}}
        \lesssim \Parent{\frac{\sigma^2_{\max}}{n}}^{1/3} + \sqrt{\frac{\sigma_{\max}}{n\varepsilon}}. 
    \end{align*}
\end{corollary}

Albeit the result is given in probability instead of in expectation, the error rate in the second statement of Corollary \ref{cor:rMSEpriestchaoest} resembles that of integrated and pointwise in-expectation lower bounds such as Corollary 1, Corollary 2 in \citet{cai:chakraborty:vuursteen2024} or Theorem 6.1 of \citet{cai:wang:zhang2023} in the \iid{} setting, i.e., when $\Sigma^n \asymp I_n$.

\section{User-Level DP Estimators for Dependent Data}
\label{sec:uDP}

Remember that in the user-level DP setting each of the $n$ users contribute $T$ observations that shall be protected as a whole.
While all the theoretical work in user-level DP that we are aware of assumes independent bounded data across both users and time, our results simultaneously cover unbounded observations and enable dependence both across the $n$ users and their $T$ observations.

The user-level DP estimators that we introduce in this section will showcase the versatility of the theoretical analysis of our item-level DP Winsorized mean estimator of the previous section. The construction of the user-level estimators will follow from the same 3 steps. Namely, given a data matrix $X \in \R^{n \times d}$ we will need to show that: 
\begin{enumerate}
    \setlength\itemsep{-0.3em}
    \item $X$ is $(\tau, \gamma)_\infty^\mu$-concentrated, 
    \item $X$ fulfills Assumption \ref{ass:iiddata} or \ref{ass:logsobdep}, 
    \item The sample average $\bar X_n$, fulfills $\twonorm{\bar X_n -\mu} \leq R_{stat}^\mu(n,\alpha)$ with probability at least $1-\alpha$. 
\end{enumerate}
In the longitudinal problems that we will consider below, the data matrix $X^n$ will be a collection of $n$ statistics computed from the $T$ observations of each individual user. In particular, by constructing specific data matrices $X^n$, we will reduce user-level mean estimation, a simple random effects model and longitudinal linear regression to item-level mean estimation. This reduction allows us to obtain finite-sample statistical guarantees for all these models from Theorem \ref{thm:rMSEwinsmeanHD}. 

\subsection{User-Level Mean Estimation}
\label{subsecuserlvlmeanest}

Similar to the item-level case, we control dependence by Assumption \ref{ass:logsobdep} but applied to a data matrix $X^n \in \R^{nT \times d}$ consisting of $nT$ observations as defined next.

\begin{definition}{(User-Level Data).}
\label{def:userlvldatamatrix}
    Collect the observations $X_u \in \R^{T \times d}$ of users $u \in [n]$ over time points $t \in [T]$ in the data matrix $X^n := \bracket{X_1\T,\dots X_n\T}\T \in \R^{nT \times d}$. Define a user-average as $\hat \mu_u := \frac 1 T X_u\T\one_T$ for $u\in[n]$ and collect them in $\hat \mu^n := \bracket{\hat \mu_1,\dots ,\hat \mu_n}\T \in \R^{n \times d}$. We call $X^n$ a user-level data matrix and $\hat \mu^n$ the user-level averages.
\end{definition}

Our user-level mean estimator feeds the matrix $\hat \mu^n \in \R^{n\times d}$ to Algorithm \ref{alg:winsmeanHD}, which then computes a private counterpart to the following empirical mean over the $nT$ observations: 
\begin{align*}
    \bar X_{nT}
    := \frac{1}{nT} \sum_{i=1}^{nT} X_{ij}
    = \frac{1}{n} \sum_{u=1}^n \hat \mu_u
    = \frac 1 n (\hat \mu^n)\T\one_n. 
\end{align*}
Feeding the user-level averages $\hat\mu^n$ to Algorithm \ref{alg:winsmeanHD} ensures privacy for each single $\hat \mu_u$ and thus all observations of a user at once, which is exactly what user-level \DP{} requires. Utility also follows by that of Algorithm \ref{alg:winsmeanHD} with the main difference being that the $T$ observations per user essentially make $\hat \mu_u$ concentrate with a shrinking radius of order $O(1/\sqrt{T})$. The main ingredient needed to make these claims rigorous is Theorem \ref{thm:rMSEwinsmeanHD}. For this we need to show that $\hat \mu^n$ is $(\tau, \gamma)_\infty$-concentrated around the mean $\mu$. This is established in Lemma \ref{lem:concuserlvlmeanest} which additionally gives a non-asymptotic deviation error rate for the non-private $\bar X_{nT}$. The proof of is given in Appendix \ref{app:userlvlmeanest}.

\begin{lemma}
    \label{lem:concuserlvlmeanest}
    Let $X^n \in \R^{nT \times d}$ be a user-level data matrix. Assume $\E[X^n] = \one_{nT}\mu\T$, with mean $\mu\in \R^d$ and let $X^n$ fulfill Assumption \ref{ass:logsobdep} \st{} $\rho M^2 \lesssim 1$. Then, $\hat \mu^n$ is $(\tau, \gamma)_\infty^\mu$-concentrated with $\tau = \sqrt{2\rho\log(2dn/\gamma)/T}$ and with probability at least $1-\alpha$, 
    \begin{align*}
        \Twonorm{\bar X_{nT} - \mu}
        \leq \sqrt{\frac{2d\rho \log(2d/\alpha)}{nT}}
    \end{align*}
\end{lemma}

\begin{remark}
    \label{rem:taugammamaximal}
    The concentration radius in Lemma \ref{lem:concuserlvlmeanest} is actually $\tau = \sqrt{2\rho_\star\log(2dn/\gamma)/T}$ where $\rho_\star = \max_{u \in [n], j \in [d]} \rho_{uj}$ and $\rho_{uj} \leq \rho$ are the marginal log-Sobolev constants of $(X_u)_{\cdot j}$. If $\rho_\star$ is known, this $\tau$ is constant in $n$ up to logarithms if $\rho_{uj} \lesssim 1$ even under strong dependence, i.e., $\rho \asymp n^\omega, \omega \geq 0$. 
\end{remark}

Now, Lemma \ref{lem:concuserlvlmeanest} directly implies an error bound for our Winsorized mean estimator through Theorem \ref{thm:rMSEwinsmeanHD}. We state this guarantee next and prove it in Appendix \ref{app:userlvlmeanest}.

\begin{corollary}
    \label{cor:rMSEuserlvldes}
    Let $\mathcal A$ denote Algorithm \ref{alg:winsmeanHD} taking input $\hat \mu^n$ as in Definition \ref{def:userlvldatamatrix}. Then, $\mathcal A$ is $(\varepsilon, \delta+\varrho)$-\uDP{} for $\delta, \varrho, \varepsilon \in (0,1)$. Further, let $\gamma \in (0, 1 \wedge \frac n 4)$ and assume the user-level data matrix $X^n \in \R^{nT \times d}$ is \st{} $\E[X^n] = \one_{nT}\mu\T$ and fulfills Assumption \ref{ass:logsobdep} \st{} $\rho M^2 \lesssim 1$. Then, with probability at least $1-3d\gamma-O(d^2n/\delta \cdot e^{-\kappa n \varepsilon^\prime})$ for an absolute constant $\kappa > 0$,
    \begin{align*}
        \Twonorm{\mathcal A(\hat \mu^n) - \mu} 
        \lesssim \sqrt{\frac{d\rho \log(2/\gamma)}{nT}} + \sqrt{\frac{d^2\rho\log(2dn/\gamma)\log(1/\varrho)\log(3/(d\gamma))^2}{Tn^2\varepsilon^2}}. 
    \end{align*}
\end{corollary}

The rate in Corollary \ref{cor:rMSEuserlvldes} is additive in the statistical rate of order $d/(nT)$ and a cost of privacy term of order $d^2/(Tn^2\varepsilon^2)$. Both also arise in the user-level in-expectation bound in \citet[Corollary 1]{levy:sun:amin:kale:kulesza:mohri:suresh2021} which was derived for bounded data. Moreover, when limiting our finite-sample result to the \iid{} setting, our rates match those present in the upper and lower bounds of Theorem 1.2 and Theorem 1.4 by \citet{agarwal:kamath:majid:mouzakis:silver:ullman2025} in the limit when $k \to \infty$ directional moments are bounded. 

\begin{remark}
    While the error in Corollary \ref{cor:rMSEuserlvldes} goes to $0$ as $T\to\infty$, the probability of this event cannot be made arbitrarily small unless $n\to\infty$. This is analogous to the impossibility result \citet[Theorem 8]{levy:sun:amin:kale:kulesza:mohri:suresh2021} which shows that learning under user-level \DP{} is impossible in general if only $T \to \infty$ but the number of users $n$ stays constant. 
\end{remark}

\subsubsection{In-Expectation Analysis}

Theorem \ref{thm:inexpub} also immediately implies an MSE upper bound in the user-level setting. We obtain it by running the item-level mean estimator \eqref{eq:mean_data_split} on observations $\hat\mu^n = \bracket{\hat \mu_1\T,\dots ,\hat \mu_n\T}\T \in \R^{n \times d}$ like in Definition \ref{def:userlvldatamatrix}. 
This construction assumes an additional independent copy $\hat \mu^n_Z$ of $\hat \mu^n$ which is used to estimate a crude private midpoint $\hat m := \hat m(\hat \mu_Z^n)$ with Algorithm \ref{alg:privrangestabhist}. We then project the second half $\hat \mu^n$ onto the hypercube $\hat I^d := \hat I_1 \times \dots \times \hat I_d$ where $\hat I_j = [\hat m_j \pm 3\tau]$ to then estimate the mean using $\proj{\hat I^d}{\hat\mu_1},\dots,\proj{\hat I^d}{\hat\mu_n}$.

\begin{corollary}
    \label{cor:consistwinsmeanHDuserlvl}
    Construct $\hat\mu^n$ and $\hat\mu^n_Z$ from user-level data matrices $X^n$ and $Z^n$ as in Definition \ref{def:userlvldatamatrix}. Let $\mathcal A_{\hat m(\hat \mu^n_Z)}$ denote the estimator \eqref{eq:mean_data_split} computed on $\hat\mu^n$, with midpoint $\hat m(\hat\mu^n_Z)$. Then, $\mathcal A_{\hat m(\hat \mu^n_Z)}$ is $(\varepsilon, \delta + \varrho)$-\uDP{} for $\varepsilon, \delta, \varrho \in (0,1)$. Further, let $X^n = [X_1\T,\dotsm, X_n\T]\T \in \R^{nT \times d}$ have rows with mean $\mu \in \R^d$ and satisfy Assumption \ref{ass:iiddata} with $\rho$-sub-Gaussian entries or Assumption \ref{ass:logsobdep} \st{} $\rho M^2 \lesssim 1$. Let $\hat\mu^n_Z$ be an independent copy of $\hat\mu^n$ and choose $\tau^2 \asymp \rho \log(4n)$ in Algorithm \ref{alg:privrangestabhist}. Then,
    \begin{align*}
        \E[\twonorm{\mathcal A_{\hat m(\hat\mu^n_Z)}(\hat\mu^n) - \mu}^2]
        &\lesssim d\rho \log(4n) \Parent{\frac{1}{nT} + \frac{d\log(1/\varrho)}{Tn^2\varepsilon^2} + \frac{dn}{\delta Te^{\kappa n \varepsilon^\prime/2}}} + \frac{d\twonorm{\mu}^2}{\delta e^{\kappa n \varepsilon^\prime}}.
    \end{align*}
\end{corollary}

Corollary \ref{cor:consistwinsmeanHDuserlvl} is proven in Appendix \ref{app:userlvlmeanest}. For reasonable choices of $\delta$ and $\varepsilon$ the estimator is consistent as $n \to \infty$ and if $\twonorm{\mu}^2 = o(\delta e^{\kappa n \varepsilon^\prime})$. As the $\hat \mu_u$ themselves are averages, their variances are of order $\rho/T$ and thus by analogy with $\bar X_{nT}$ we expect an MSE that is smaller by order $1/T$ compared to the item-level case. However, the fourth term in Corollary \ref{cor:consistwinsmeanHDuserlvl} shows that this is not the case. 

\begin{remark}
    \label{rem:imposibilityuDP}
    We see from Corollary \ref{cor:consistwinsmeanHDuserlvl} that for $\delta \in (n\varepsilon^\prime/e^{\kappa n \varepsilon^\prime}, n\inv]$ the rate in $n$ and $\varepsilon$ is the same as in the item-level setting of Theorem \ref{thm:inexpub}. Furthermore, the fourth term does not vanish for a fixed $n$ as $T \to \infty$, so our bound exhibits the same impossibility of learning in the $T \to \infty$ limit that was established in \citet{levy:sun:amin:kale:kulesza:mohri:suresh2021} even though a closer look at \citet[Theorem 8]{levy:sun:amin:kale:kulesza:mohri:suresh2021} reveals that their lower bound is realized for $\delta < (2ne^{n\varepsilon})\inv$ which is excluded by Corollary \ref{cor:consistwinsmeanHDuserlvl}. Still, our rates are comparable to the best existing estimators under \iid{} assumptions. Note also that Theorem \ref{thm:inexplb} is not applicable because the local mean estimators $\hat \mu_u$ are $\frac{\rho}{T}$-sub-Gaussian which makes the condition $\infnorm{\mu}^2 < \frac{\rho}{T}$ that would be required by Theorem \ref{thm:inexplb} excessively stringent.
\end{remark}

\subsection{Random Effects Location Model}

Random effects models are commonly used for the analysis of clustered or correlated data \citep{pinheiro:bates2000,fahrmeir:tutz2001,demidenko2013}. We restrict our presentation here to a simple one-dimensional user-level random effects location model presented in Definition \ref{def:1drandeffects}. We do so to fully focus on the dependence between users introduced by random effects and investigate their impact on the convergence rate of our Winsorized mean estimator. Note, however, that our theoretical guarantees of Section \ref{sec:meanest} are powerful enough to also handle random effects in linear regression settings like the one in Subsection \ref{sec:linreg}. 

\begin{definition}{(Random Effects Location Model).}
    \label{def:1drandeffects}
    Let $\mu \in \R$ and $\sigma_U^2, \sigma^2 < \infty$. Assume independent noise $U_g \sim \mathcal N(0, \sigma_U^2)$ and $\epsilon_u \sim \mathcal N(0, \Sigma_u)$ with $\Sigma_u \in \R^{T \times T}$ and $\Sigma_u \preceq \sigma^2$. For all users $u \in [n]$, groups $g \in [G]$ and times $t \in [T]$, define a random effects model through observations $Y_{gut} := \mu + U_g + \epsilon_{ut}$ collected in $Y^n \in \R^{nT}$. 
\end{definition}

In addition to the random effects that introduce dependence between users of the same group $g$, the model in Definition \ref{def:1drandeffects} also contains noise that is independent between users but correlated across a user's observations over time. Hence, it contains both, inter- and intra-user dependence. Nevertheless, we will see that the analysis of the model in Definition \ref{def:1drandeffects} is simple because it can be viewed as an instance of the user-level mean estimation problem covered in Subsection \ref{subsecuserlvlmeanest}. Hence, we only need to show that the resulting user-level data matrix fulfills log-Sobolev dependence. We do that in the following lemma. 

\begin{lemma}
    \label{lem:randeffectslogsobdes}
    Let $Y^n \in \R^{nT}$ be the design of the random effects model in Definition \ref{def:1drandeffects}. Writing $n_g$ for the group sizes, define $\rho := \max_{g \in [G]} \Parent{\sigma_U^2 n_gT + \sigma^2}$ and $M^2 := (2\pi (\sigma_U^2 + \sigma^2))\inv$. Then, $Y^n$ is a user-level log-Sobolev data matrix and Assumption \ref{ass:logsobdep} holds with constants $(\rho, M)$. 
\end{lemma}

Our proof of Lemma \ref{lem:randeffectslogsobdes} can be found in Appendix \ref{app:randeffects}. It exploits the joint Gaussian distribution of observations $Y_{gut}$ in Definition \ref{def:1drandeffects}. The next result is a direct consequence of Corollary \ref{cor:rMSEuserlvldes}. It gives an upper bound on the estimation error of privately estimating the location parameter $\mu$. 

\begin{corollary}
    \label{cor:rMSErandeffects}

    Construct $\hat \mu^n \in \R^{n}$ from $Y^n\in\mathbb{R}^{nT}$ and denote by $\mathcal A$ the output of Algorithm \ref{alg:winsmeanHD} with input $\hat \mu^n$. Then $\mathcal A$ is $(\varepsilon, \delta)$-\uDP{} for $\delta, \varepsilon \in (0,1)$. Furthermore, let $\gamma \in (0,1 \wedge \frac n 4)$ and $Y^n \in \R^{nT}$ be the design of Definition \ref{def:1drandeffects} with mean $\mu \in \R$ and maximum group size $n_{g^\star}$. Then, if $\rho M^2 \lesssim 1$, with probability at least $1-3\gamma-O(n/\delta \cdot e^{-\kappa n \varepsilon})$ for an absolute constant $\kappa > 0$,
    \begin{align*}
        \Twonorm{\mathcal A(\hat \mu^n) - \mu} 
        \lessapprox \sqrt{\frac{\sigma^2_U \cdot n_{g^\star}T + \sigma^2}{nT}} + \sqrt{\frac{\sigma^2_U \cdot n_{g^\star}T + \sigma^2}{Tn^2\varepsilon^2}}. 
    \end{align*}
\end{corollary}

\begin{remark}
    \label{rem:rMSErandeffects}
    The rate in the error bound only matches the baseline of independent users with uncorrelated noise if $\sigma^2_U n_{g^\star}T + \sigma^2 \asymp 1$. Note that $\sigma^2 \asymp 1$ can be realized, e.g., through sparse correlations, a Toeplitz covariance structure with fast decay or equicorrelated noise with variance $\sigma^2 \asymp 1/T$. However, $\sigma^2_U \asymp 1 / (n_{g^\star} T)$ is stringent and will hold for instance when the maximum group size $n_{g^\star} \asymp 1$ and $\sigma^2_U \asymp 1/T$. Otherwise, the latter example results in slower rates relative to the \iid{} case, but the rates are sharp in this case as shown in the next example.
\end{remark}

\begin{example}{(Slow Rates for the Empirical Mean).}
    \label{exa:failempmean}
    The behavior of the error rate in Corollary \ref{cor:rMSErandeffects} as a function of $n_{g^*}$ is not due to a loose bound when deriving the log-Sobolev dependence of $Y^n$. To see this, assume $\epsilon_u = 0$. Then, in the derivation of Lemma \ref{lem:randeffectslogsobdes} the operator norm of the covariance matrix $\Sigma^n \in \R^{nT \times nT}$ of the joint is exactly
    \begin{align*}
        \opnorm{\Sigma^n}
        = \max_{g \in [G]} \opnorm{\Sigma^U_g}
        = \max_{g \in [G]} \opnorm{\sigma_U^2 \one_{n_gT}\one_{n_gT}\T}
        = \sigma_U^2 \cdot T \cdot \max_{g \in [G]} n_g.
    \end{align*}
    This suggests that already one group with $n_g \asymp n$ kills the MSE-rate for user-level mean estimation. 
    
    Note that this phenomenon is due to the nature of this statistical model and not the user-level setting. To show this, let $X^n \in \R^n$ with $X^n \sim \mathcal N(\mu, \Sigma^n)$ where $\mu \in \R^d$ and 
    \begin{align*}
        \Sigma^n
        := 
        \begin{bmatrix}
             \one_{n_g}\one_{n_g}\T & 0 \\
             0 & 0
        \end{bmatrix}
        + I_n. 
    \end{align*}
    As $\bar X_n$ is unbiased for $g(\mu) = \frac 1 n \one_n\T \mu$, the MSE reduces to the variance that we may compute exactly: 
    \begin{align*}
        \operatorname{MSE}(\bar X_n)
        = \var{\bar X_n}
        = \frac{1}{n^2}\Var{\sum_{i=1}^n X_i}
        = \frac{1}{n^2} \sum_{i,j}^n \cov{X_i, X_j}
        = \frac{1}{n^2} \Parent{\sum_{i,j}^{n_g} 1 + n}
        = \frac{n_g^2}{n^2} + \frac 1 n.
    \end{align*}
    Here, the asymptotic rate of $\bar X_n$ is also constant if $n_g \asymp n$. This is optimal for unbiased estimators by the Cramér-Rao Lower Bound. For any unbiased estimator $T$ of $g(\mu)$, 
    \begin{align*}
         \var{T} 
         \geq \nabla_\mu g(\mu)\T I(\mu)\inv \nabla_\mu g(\mu) 
         = \Parent{\frac 1 n \one_n}\T \Sigma^n \Parent{\frac 1 n \one_n}
         = \frac{n_g^2}{n} + \frac 1 n. 
    \end{align*}
    Note also that assuming $\mu = \tilde \mu \one_n$ with $\tilde \mu \in \R$ as in Corollary \ref{cor:rMSErandeffects} there is potential for improvement through additional information. For instance, an unbiased estimator can do better than $\bar X_n$ if it knows which observations belong to the equi-correlated group. Whether and how such information can be obtained privately to construct better estimators is unclear to the authors at this point. 
\end{example}

\subsection{User-level DP Linear Regression for Longitudinal Data}
\label{sec:linreg}

While the body of literature on item-level differentially private linear regression is growing \citet{wang2018, alabi:mcmillan:sarathy:smith:vadhan2022, liu:jain:kong:oh:suggala2023, avellamedina:bradshaw:loh2023, brown:dvijotham:evans:liu:smith:thakurta2024, bombari:seroussi:mondelli2025}, little work addresses linear regression in user-level \DP{} and nothing has been worked out for dependent longitudinal data. Here, we apply our Winsorized mean estimator to show that it can be used to estimate the regression coefficients $\beta \in \R^p$ in a user-level linear regression model as formalized in Definition \ref{def:linregmodel}. 

\begin{definition}{(User-Level Linear Model).}
    \label{def:linregmodel}
    Let $u\in [n]$ be users and $t \in [T]$ time points. Let $\beta \in \R^p$, $X^n = [X_1\T,\dots ,X_n\T]\T \in \R^{nT \times p}$ and $\epsilon^n \in \R^{nT}$ where $\epsilon^n \sim \mathcal N(0, \Sigma^n)$. Assume $\Sigma^n$ is block-diagonal with blocks $\Sigma_{u} \in \R^{T \times T}$. Define the user-level linear regression model through observations $Y^n := [Y_1\T,\dots ,Y_n\T]\T = X^n\beta + \epsilon^n$. 
\end{definition}

When $T=1$ this user-level linear model reduces to the standard linear model with \iid{} Gaussian errors. When $T > 1$ each user provides several observations that can be correlated due to the block-diagonal structure of the covariance matrix $ \Sigma^n$. When $T$ is reasonably large, a simple idea for privately estimating $\beta$ is to estimate $\beta$ locally for each user and combine these estimates by privately averaging the local non-private estimates over the $n$ users. This strategy amounts to a divide and conquer algorithm that reduces the problem of user-level linear regression under \DP{} to item-level differentially private mean estimation. 
Since the non-private counterpart of the \uDP{} estimator we have in mind is no longer a standard least squares estimator, it is natural to wonder how efficient this estimator is relative to standard and generalized least squares. The next lemma clarifies this point. Its proof can be found in Appendix \ref{app:linreg}.

\begin{lemma}
    \label{lem:userlevelest}
    Let $X^n = [X_1\T,\dots ,X_n\T]\T, Y^n = [Y_1\T,\dots ,Y_n\T]\T, \Sigma^n = {\diagonal{\Sigma_1,\dots , \Sigma_n}}$ be the design, observations and covariance matrix in Definition \ref{def:linregmodel}. Define the following estimators of $\beta$:
    \begin{enumerate}
        \setlength\itemsep{-0.3em}
        \item $\hat\beta := \frac 1 n \sum_{u=1}^n \hat \beta_u$ with $\hat\beta_u = (X_u\T X_u)\inv X_u\T Y_u$,
        \item $\hat\beta_w := \frac 1 n \sum_{u=1}^n \hat \beta_{wu}$ with $\hat\beta_{wu}: = (X_u\T \Sigma_u\inv X_u)\inv X_u\T \Sigma_u\inv Y_u$,
        \item $\hat \beta_{OLS} := (\sum_{u=1}^nX_u\T X_u)\inv \sum_{u=1}^nX_u\T Y_u$,
        \item $\hat \beta_{wOLS} := (\sum_{u=1}^nX_u\T\Sigma_u\inv X_u) \inv \sum_{u=1}^nX_u\T \Sigma_u\inv Y_u$.
    \end{enumerate}
    Then, $\hat\beta, \hat\beta_u, \hat\beta_{w}, \hat\beta_{wu}$ are unbiased for $\beta$ and have the following variance relations:
    \begin{enumerate}[(i)]
        \setlength\itemsep{-0.3em}
        \item $\cov{\hat\beta_{OLS}} \not\preceq \cov{\hat\beta}$, 
        \item $\cov{\hat\beta_{OLS}} \preceq \cov{\hat\beta}$ if $\Sigma_u = \sigma^2 I_T$, \hfill $\forall u \in [n]$
        \item $\cov{\hat\beta_{OLS}} = \cov{\hat\beta}$ if $X_u\T X_u = D$ for some constant $D \succeq 0$, \hfill $\forall u \in [n]$
        \item $\cov{\hat\beta_{wOLS}} \preceq \cov{\hat\beta_w} \preceq \cov{\hat\beta}$. 
    \end{enumerate} 
\end{lemma}

Recall that by the Gauss-Markov Theorem $\hat\beta_{wOLS}$ is the most efficient unbiased linear estimator; more precisely for all unbiased linear estimator $\hat\theta$ we have $\cov{\hat\beta_{wOLS}} \preceq \cov{\hat\theta}$. However, $\hat\beta_{wOLS}$ and its unweighted version $\hat\beta_{OLS}$ are not distributed estimators and cannot be privatized as easily as $\hat\beta$ and $\hat\beta_w$. Lemma \ref{lem:userlevelest} shows that having access to $\Sigma_u$, we should use $\hat\beta_w$ as it is a better divide-and-conquer estimator than $\hat\beta$. Yet, if we only have access to an upper bound on $\Sigma_u$ and therefore cannot reweigh, using $\hat\beta$ is in some cases equally efficient as $\hat \beta_{OLS}$ and generally neither one dominates the other in the general case when $\Sigma_u \neq \sigma^2 I_T$. For a simple counterexample with $\Sigma_u \asymp I_T$ but with heterogeneous variances see the proof of Lemma \ref{lem:userlevelest} in Appendix \ref{app:linreg}. Hence, even in some non-private settings $\hat\beta$ is a reasonable distributed estimator. We work with a private counterpart of this estimator that is straightforward to compute and analyze using our private mean estimator. The following lemma establishes the concentration result of the local per user estimators $\hat\beta_1,\dots,\hat\beta_n$ that we need to get a finite sample guarantee for our private estimator. The proof of Lemma \ref{lem:concuserlvlreg} is given in Appendix \ref{app:linreg}. 

\begin{lemma}
    \label{lem:concuserlvlreg}
    Let $X^n = [X_1\T,\dots ,X_n\T]\T$ and $Y^n = [Y_1\T,\dots ,Y_n\T]\T$ be as in Definition \ref{def:linregmodel}. Define $\hat M_u := \frac 1 T X_u\T X_u$ and assume $\theta I_p \preceq \hat M_u \preceq I_p \vartheta$ and $\Sigma_u \preceq \sigma^2 I_T$. Then, for the estimator $\hat \beta = \frac 1 n \sum_{u=1}^n \hat \beta_u$ with $\hat\beta_u = (X_u\T X_u)\inv X_u\T Y_u$, with probability at least $1-\alpha$,
    \begin{align*}
        \Twonorm{\hat\beta - \beta} 
        \leq \sqrt{\frac{p\vartheta\sigma^2}{nT\theta^2}} + \sqrt{\frac{2\vartheta\sigma^2\log(1/\alpha)}{nT\theta^2}}.
    \end{align*}
    Further, ${\hat\beta}^n := [\hat\beta_1,\dots ,\hat\beta_n]\T \in \R^{n\times p}$ is $(\tau, \gamma)^\beta_\infty$-concentrated with $\tau = \sqrt{2 (\vartheta/\theta^2)\sigma^2 \log\Parent{2dn/\gamma}/T}$. 
\end{lemma}

Lemma \ref{lem:concuserlvlreg} allows us to instantiate Theorem \ref{thm:rMSEwinsmeanHD}, which immediately yields the following result. 

\begin{corollary}
    \label{cor:MSEuserlvlreg}
    Algorithm \ref{alg:winsmeanHD} is $(\varepsilon, \delta+\varrho)$-\uDP{} on input $\hat\beta^n$ for $\delta, \varrho, \varepsilon \in (0,1)$. Let $\gamma \in (0, 1 \wedge \frac n 4)$, and let $X^n \in \R^{nT \times p}$ and $Y^n \in \R^{nT}$ be as in Definition \ref{def:linregmodel}. Construct $\hat\beta^n \in \R^n$ as in Lemma \ref{lem:concuserlvlreg} from $X^n, Y^n$. Then, with probability at least $1-3p\gamma-O(p^2n/\delta \cdot e^{-\kappa \varepsilon^\prime n})$ where $\kappa > 0$,
    \begin{align*}
        \Twonorm{\mathcal A(\hat\beta^n) - \beta} 
        &\lessapprox \sqrt{\frac{p\vartheta\sigma^2}{nT\theta^2}} + \sqrt{\frac{p^2\vartheta\sigma^2}{\theta^2Tn^2\varepsilon^2}}.
    \end{align*}
\end{corollary}

\section{Extensions to Local Differential Privacy}
\label{sec:LDP}

The central model of \DP{} considered up to now requires that items/users trust a central server to aggregate and privatize their data. Whenever such trust is not given, items or users themselves have to privatize their data instead. Such a setting is called the \textit{local} model of \DP{} \citep{kasiviswanathan:lee:nissim:raskhodnikova:smith2011,duchi:jordan:wainwright2018}. 

\begin{definition}{($(\varepsilon, \delta)$-Differential Privacy, Local Model).}
    Let $\varepsilon, \delta > 0$ and let $\mathcal A : \mathcal{X} \to \R^d$ be a randomized algorithm, mapping $x \in \mathcal{X}$ to a $\mathcal B(\R^d)$-measurable random variable $\mathcal A(x)$. Then, $\mathcal A$ is $(\varepsilon, \delta)$-\LDP{} in the local model if for all $x,y \in \mathcal{X}^n$ we have
    \begin{align*}
        \P\Parent{\mathcal A(x_i) \in S} 
        \leq e^\varepsilon \cdot \P\Parent{\mathcal A(y_i) \in S} + \delta,
        \quad \text{for all $S \in \mathcal B(\R^p)$ and $i \in [n]$}. 
    \end{align*}
    Furthermore, we say that 
    \begin{enumerate}
        \setlength\itemsep{-0.3em}
        \item $\mathcal A$ is \textit{item-level}-\LDP{} if $x_i, y_i \in \R^{1 \times d}$ for all $i \in [n]$,
        \item $\mathcal A$ is \textit{user-level}-\LDP{} if $x_i, y_i \in \R^{T \times d}$ for all $i \in [n]$.
    \end{enumerate}
\end{definition}

The randomized response mechanism in Algorithm \ref{alg:randresponse} was the first randomized algorithm used to ensure this local notion of data privacy. It was already introduced by \citet{warner1965} in the context of survey sampling long before local \DP{} was formalized. The randomized response mechanism is well-known to be $(\varepsilon, 0)$-\LDP{}; see \citet[Section 3.2]{dwork:roth2014}. 

\begin{theorem}
    \label{thm:randresponse}
    Algorithm \ref{alg:randresponse} is $(\varepsilon, 0)$-\LDP{}. 
\end{theorem}

\begin{algorithm}
    \caption{RandomizedResponse$(x, \varepsilon)$:}
    \label{alg:randresponse}
    \begin{algorithmic}[1]
        \REQUIRE $x \in \Curly{0,1}$, $\varepsilon > 0$
        \STATE $u \gets U \sim \operatorname{Unif}[0,1]$
        \STATE $\pi \gets e^{\epsilon}/(1+e^{\epsilon})$
        \STATE $\tilde x \gets x \cdot \indicator{U \leq \pi} + (1-x) \cdot \indicator{U >\pi}$
        \RETURN $\tilde x, \pi$
    \end{algorithmic}
\end{algorithm}

In the rest of this section we use the randomized response mechanism in conjunction with the Laplace mechanism of Algorithm \ref{alg:lapmech} to lift the theory on Winsorized mean estimation under dependence developed so far to the local model -- albeit under slightly stronger assumptions. In addition to $(\tau, \gamma)_\infty$-concentration of the data and log-Sobolev dependence or \iid{} rows of the data matrix, we need to assume that $\infnorm{\mu} \leq \mathsf B$. This is because we replace the stable histogram underlying our private midpoints by a histogram estimator based on randomized responses that cannot handle an infinite number of bins. Besides this replacement our approach remains unchanged and therefore proves to be substantially more general than existing techniques as we allow for unbounded dependent data. In particular, in the case of bounded \iid{} observations we recover the upper bound in Theorem 6 by \citet{kent:berrett:yu2024}. Modulo histogram estimation, the proofs of our results in the central and local model all build on general statements that we instantiate for both cases. 

\subsection{Histogram Estimator}

Our local histogram estimator again allows for dependent observations. The estimator in Algorithm \ref{alg:randhist} is a histogram estimator that relies on the randomized response mechanism for privacy also in the local model. The histogram's construction is similar to that used by \citet{kent:berrett:yu2024}. However, we do not require merging multiple neighboring bins and include a debiasing step that ensures that it is consistent. Unlike Algorithm \ref{alg:stablehist} intended for the central model, this histogram can only handle finitely many disjoint bins but it is $(\varepsilon, 0)$-\LDP{}. Remember that the stable histogram is $(\varepsilon,\delta)$-\DP{} with $\delta>0$. 

\begin{algorithm}
    \caption{RandomizedHistogram$(X, (B_k)_{k \in [N_{\mathsf{bins}}]}, \varepsilon, \delta)$:}
    \label{alg:randhist}
    \begin{algorithmic}[1]
        \REQUIRE $X \in \R^n $, $\varepsilon, \delta = 0$
        \FORALL{$k \in [N_{\mathsf{bins}}]$}
            \STATE $c_{ik} \gets \indicator{X_i \in B_k}$ \hfill $\forall i \in [n]$
            \STATE $\tilde c_{ik}, \pi \gets \operatorname{RandomizedResponse}(c_{ik},{\varepsilon}/{2})$ \hfill $\forall i \in [n]$
            \STATE $\tilde p_k \gets \frac 1 n \sum_{i=1}^n \tilde c_{ik}$
            \STATE $\tilde p_k^{\operatorname{debiased}} \gets \frac{\tilde p_k - (1-\pi)}{2\pi-1}$
        \ENDFOR
        \RETURN $\boldsymbol{\tilde p} := (\tilde p_1^{\operatorname{debiased}}, \tilde p_2^{\operatorname{debiased}},\dots,p_{N_{\mathsf{bins}}}^{\operatorname{debiased}} )$
    \end{algorithmic}
\end{algorithm}

In Lemma \ref{lem:randhist} we provide a privacy and utility guarantee for Algorithm \ref{alg:randhist} when estimating the true probability mass $p_1,p_2,\dots,p_{N_{\mathsf{bins}}}$ of bins $(B_k)_{k \in [N_{\mathsf{bins}}]}$ given log-Sobolev dependent or \iid{} data. The proof can be found in Appendix \ref{app:hist_LDP}

\begin{lemma}
    \label{lem:randhist}
    Algorithm \ref{alg:randhist} is $(\epsilon, 0)$-\LDP{} for $\varepsilon > 0$. Let $\eta > 0$, and $X^n \in \R^n$ fulfill Assumption \ref{ass:iiddata} (A1) or Assumption \ref{ass:logsobdep} (A2) with constants $(\rho, M)$. Further, let $( B_k)_{k=1}^{N_{\mathsf{bins}}}$ be a finite collection of disjoint bins covering $[- \mathsf B, \mathsf B] \subset \R$ and define $p_k := \P(X_1 \in B_k)$. Then, for the output $\boldsymbol{\tilde p}$ we have 
    \begin{align*}
        \P\Parent{\max_{k \in N_{\mathsf{bins}}} \abs{\tilde p_k - p_k} \geq \eta} 
        \leq 2N_{\mathsf{bins}} \cdot \exp\Parent{-\frac{n\eta^2 }{4} \tanh^2\Parent{\frac{\varepsilon}{4}}} + \begin{cases}
            2 \exp\Parent{-\frac{n\eta^2}{8}} & \text{(A1)}, \\
            \frac{16}{\eta} \exp\Parent{-\frac{2}{27} \frac{n\eta^3}{4^3 \rho M^2}} & \text{(A2)}.
        \end{cases}. 
    \end{align*}
\end{lemma}

Lemma \ref{lem:randhist} is a local DP counterpart of Lemma \ref{lem:utilhistlogsob} in Section \ref{sec:hist}. The lemma mainly differs in that it assumes a finite number of bins, and provides a $(\varepsilon, 0)$-\LDP{} guarantee instead of the $(\varepsilon, \delta)$-\DP{} guarantee of the stable histogram. Since both lemmas are shown by decomposing the error $\abs{\tilde p_k - p_k}$ into the privacy loss $\abs{\tilde p_k - \hat p_k}$ and the estimation error $\abs{\hat p_k - p_k}$, the second term in the \rhs{} of both statements is identical. Their first terms, obtained by bounding the privacy loss with high probability differs in that the $1/\delta$ factor in front of the exponential is replaced by the number of bins $N_{\mathsf{bins}}$ and the privacy budget parameter $\varepsilon$ shows up as $\tanh^2(\varepsilon)$ instead of just $\varepsilon$. When $\varepsilon \lesssim 1$, we bound this instead by $\varepsilon^2$ which gives minimax optimal rates for the local Winsorized mean estimator as we will see below. 

\subsection{Mean Estimators}

The construction of our local mean estimators are analogous to their central model counterparts in Section \ref{sec:meanest}. Here, the only difference is that we use the randomized histogram when estimating the private midpoint of the projection interval and every item or user privatize their projected observation by adding Laplace noise before sending it to the central server. Algorithm \ref{alg:winsmean1Dlocal} formalizes this protocol. Remember that the randomized histogram can only handle a finite number of bins and thus the algorithm requires $\mathsf B \in (0,\infty)$ \st{} $\infnorm{\mu} \leq \mathsf B$ as input. 

\begin{algorithm}
    \caption{WinsorizedMean1D($X^n, \tau, \varepsilon, \delta, \mathsf B$)}
    \label{alg:winsmean1Dlocal}
    \begin{algorithmic}[1]
        \REQUIRE $X^n \in \R^n$, $\tau, \varepsilon, \delta > 0$
        \STATE $\hat I \gets \operatorname{ProjectionInterval}(X, \tau, \frac \varepsilon 2, \delta, B)$ where $\hat I = \bracket{\hat m \pm 3\tau}$ \hfill \text{Algorithm \ref{alg:privrangerandhist}}
        \RETURN $\frac 1 n \sum_{i=1}^n \proj{\hat I}{X_i} + \xi_i$ with \iid{} $\xi_i \sim \operatorname{Lap}(0, \frac{12\tau}{\varepsilon})$ 
    \end{algorithmic}
\end{algorithm}

In Lemma \ref{lem:winsmeanHDlocal} we show that $\mathcal A(X^n) = \bar X_n + \bar \Xi_n$ with high probability, where $\bar\Xi_n=\frac{1}{n}\sum_{i=1}^n\Xi_i$ with $\Xi_i\sim \operatorname{Lap}\Parent{0, \frac{12\tau}{\varepsilon^\prime}I_d}$ and $\varepsilon'=\varepsilon/\sqrt{8d\log(1/\varrho)}$. We exploit this equivalence to provide the finite-sample error bound in Theorem \ref{thm:rMSEwinsmeanHDlocal}. The proof is given in Appendix \ref{app:theorems_mean_LDP}. 

\begin{theorem}
    \label{thm:rMSEwinsmeanHDlocal}
    Let $\mathcal A$ denote Algorithm \ref{alg:winsmeanHD} calling Algorithm \ref{alg:winsmean1Dlocal} in line 3. $\mathcal A$ is $(\varepsilon,\varrho)$-\LDP{} for $\varepsilon, \varrho \in (0,1)$. Let $\kappa > 0$, $\gamma \in (0,1 \wedge \frac n 4)$ and let $X^n \in \R^{n \times d}$ be $(\tau, \gamma)^\mu_\infty$-concentrated with $\infnorm{\mu} \leq \mathsf B$ and make Assumption \ref{ass:iiddata} or \ref{ass:logsobdep} \st{} $\rho M^2 \lesssim 1$. Then, with probability at least $1-2d\gamma-O\parent{\mathsf B/\tau +1} \cdot de^{-\kappa n (\varepsilon^\prime)^2}$,
    \begin{equation*}
        \Twonorm{\mathcal A(X^n) - \mu} 
        \lesssim \Twonorm{\bar X_n - \mu} + \tau \sqrt{\frac{d^2\log(1/\varrho)\log(3/(d\gamma))^2}{n \varepsilon^2}}.
    \end{equation*}
\end{theorem}

As is typical also for local \DP{}, the error rate in Theorem \ref{thm:rMSEwinsmeanHDlocal} is additive in the statistical rate and a cost of privacy term. Here, classical statistical rates of order $\sqrt{d\rho/n}$ are dominated by the cost of privacy even for fixed $\varepsilon >0$ due to the dependence in $d$ and $\log(1/\varrho)$. Further, the dependence on $\varepsilon$ differs significantly from our result in the central model, because any dependence of the form $\varepsilon \asymp 1/n^\omega$ for $\omega > 0$ immediately affects the error rate in $n$.

In the local model, the high probability event $\mathcal A(X^n) = \bar X_n + \bar \Xi_n$ easily translates into a MSE in-expectation bound. This is, because we can control the error in the complement event using that $\infnorm{\hat m} \leq \mathsf B + \tau$. This inequality holds by design of the local private midpoint estimation in Algorithm \ref{alg:privrangerandhist}. The next result is proven in Appendix \ref{app:theorems_mean_LDP}

\begin{theorem}
    \label{thm:inexpublocal}
    Let $\mathcal A$ denote Algorithm \ref{alg:winsmeanHD} calling Algorithm \ref{alg:winsmean1Dlocal} in Line 3. $\mathcal A$ is $(\varepsilon, \varrho)$-\LDP{} for $\varepsilon, \varrho \in (0,1)$. Further, assume $X^n \in \R^{n \times d}$ is $(\tau, \gamma)_\infty$-concentrated around $\mu \in \R^d$ with $\infnorm{\mu} \leq \mathsf B$. Make Assumption \ref{ass:iiddata} or Assumption \ref{ass:logsobdep} \st{} $\rho M^2 \lesssim 1$. Then, 
    \begin{align*}
        \E[\twonorm{\mathcal A(X^n) - \mu}^2]
        & \lesssim \E\Bracket{\Twonorm{\bar X_n - \mu}^2} + \frac{d^2\tau^2\log(1/\varrho)}{n\varepsilon^2} + 
        \begin{cases}
            d^2 \mathsf B^2 \cdot \Parent{\gamma + \frac{\mathsf B/\tau}{e^{\kappa n (\varepsilon^\prime)^2}}} & \text{if $\mathsf B \geq \tau$}\\
            d^2\tau^2 \cdot \Parent{\gamma + \frac{1}{e^{\kappa n (\varepsilon^\prime)^2}}} & \text{if $\mathsf B \leq \tau$.}
        \end{cases} 
    \end{align*}
\end{theorem}

The interpretation of this result is the same as for the in-probability bound. Statistical rates of order $d\rho/n$ are dominated by the cost of privacy term if we choose $\gamma \asymp 1/(n \varepsilon^2)$. To give a clean interpretation we instantiate the theorem in the user-level setting that subsumes the item-level one. The resulting statement is given in Corollary \ref{cor:inexpubuserlvllocal} below. 

\begin{corollary}
    \label{cor:inexpubuserlvllocal}
    Let $\mathcal A$ denote Algorithm \ref{alg:winsmeanHD} calling Algorithm \ref{alg:winsmean1Dlocal} in Line 3. Then, $\mathcal A$ is $(\varepsilon, \varrho)$-\uLDP{} for $\varepsilon, \varrho \in (0,1)$. Further, let $X^n \in \R^{nT \times d}$ and $\hat \mu^n$ be as in Definition \ref{def:userlvldatamatrix}. Let $X^n$ have rows with mean $\mu \in \R^d$ \st{} $\infnorm{\mu} \leq \mathsf B$ and fulfill Assumption \ref{ass:iiddata} with $\rho$-sub-Gaussian entries or Assumption \ref{ass:logsobdep} \st{} $\rho M^2 \lesssim 1$. Then, choosing $\tau^2 \asymp \rho \log(2dTn^2\varepsilon^2)/T$, 
    \begin{align*}
        \E[\twonorm{\mathcal A(X^n) - \mu}^2]
        & \lesssim \frac{d^2\rho\log(2dTn^2\varepsilon^2)\log(1/\varrho)}{Tn\varepsilon^2} + 
        \begin{cases}
            d^2\mathsf B^2 \cdot \Parent{\frac{1}{Tn\varepsilon^2} + \frac{\mathsf B}{\sqrt \rho}\frac{\sqrt T}{e^{\kappa n (\varepsilon^\prime)^2}}} & \text{if $\mathsf B \geq \tau$}\\
            0 & \text{if $\mathsf B \leq \tau$}. 
        \end{cases}
    \end{align*}
\end{corollary}

Above, we again distinguish the cases of whether or not our apriori bound $\infnorm{\mu} \leq \mathsf B$ localizes the mean more than the concentration radius $\tau$ or not. Whether we are in the item- or user-level setting plays an interesting role in this distinction. In the item-level setting where $T = 1$ the concentration radius $\tau$ grows in $n$ and at some point will surpass $\mathsf B$. Thus, $\mathsf B \leq \tau$ is the more relevant case there and when $d = 1$ our bound attains the $1/(n\varepsilon^2)$-rate that Corollary 1 \citet{duchi:jordan:wainwright2018} predicts for bounded random variables. More interestingly, Corollary \ref{cor:inexpubuserlvllocal} generalizes this result to $\rho$-sub-Gaussian random variables with potentially unbounded domain and bounded first moment. For small $T$, the user-level setting behaves the same as the item-level one. However, as $T$ grows beyond $\rho \log(2dTn^2\varepsilon^2)/\mathsf B^2$ we pay the second term in the upper bound which corresponds to the prize of estimating the projection interval. If we specialize to observations that are bounded in $L_\infty$-norm, for which $\rho = \mathsf B^2$ and we set $\mathsf B = 1$, our result recovers Theorem 6 by \citet{kent:berrett:yu2024} up to logarithmic factors. 

\subsubsection{Extensions beyond Item-Level Mean Estimation}

Instantiating Theorem \ref{thm:rMSEwinsmeanHDlocal} immediately yields finite-sample risk bounds for item-level nonparametric regression, user-level mean estimation, random effects location estimation and linear regression with longitudinal data. The following gives a brief overview on these results, starting with nonparametric regression in Corollary \ref{cor:rMSEpriestchaoestlocal}. This result is a local DP analog of Corollary \ref{cor:rMSEpriestchaoest}.

\begin{corollary}
    \label{cor:rMSEpriestchaoestlocal}
    Let $\mathcal A$ denote Algorithm \ref{alg:winsmeanHD} taking $\hat f^n(x)$ as input with $x\in[0,1]$. $\mathcal A$ is $(\varepsilon, 0)$-\DP{} for $\varepsilon \in (0,1)$. Let $\gamma \in (0,1 \wedge\frac n 4)$ and assume $Y^n$ is as in Definition \ref{def:fixeddesregmod} and $\sigma^2_{\max} \gtrsim 1$. Then, for all $x \in [\zeta,1-\zeta]$ where $\zeta = b\sqrt{2\log(2/b)}$ and $b > 0$, with probability at least $1-3\gamma-O(\infnorm{f}/(L_fb) \cdot e^{-\kappa n \varepsilon^2})$,
    \begin{align*}
        \Abs{\mathcal A(\hat f^n(x)) - f(x)}
        &\lessapprox b + \sigma_{\max} \Parent{\frac{1}{nb} + \sqrt{\frac{1}{nb}} + \sqrt{\frac{1}{nb^2\varepsilon^2}}}. 
    \end{align*}
    For an optimally chosen $b \asymp (\sigma^2_{\max}/n)^{1/3} \vee (\sigma_{\max}/\sqrt n\varepsilon)^{1/2}$ this becomes
    \begin{align*}
        \Abs{\mathcal A(\hat f^n(x)) - f(x)}
        &\lessapprox \Parent{\frac{\sigma^2_{\max}}{n}}^{1/3} \vee \sqrt{\frac{\sigma_{\max}}{\sqrt n\varepsilon}}
        \lesssim \Parent{\frac{\sigma^2_{\max}}{n}}^{1/3} + \sqrt{\frac{\sigma_{\max}}{\sqrt n\varepsilon}}. 
    \end{align*}
\end{corollary}

The pointwise error rate in Corollary \ref{cor:rMSEpriestchaoestlocal} shown in Appendix \ref{app:nonparametricslocal} resembles the almost optimal integrated in-expectation rate in Theorem 3.1 and Theorem 4.1 by \citet{gyorfi:kroll2025} obtained with basis transformation estimators. Our result privatizes the  Priestley-Chao regression estimator of Definition \ref{def:priestchaoest} and thus complements their approach (See also Subsection 2.2.3, \citet{kent:berrett:yu2024}).

The next three results concern user-level estimation for mean estimation, location random effects models and longitudinal regression. We first give a general user-level mean estimation result that is the local counterpart of Corollary \ref{cor:rMSEuserlvldes}.

\begin{corollary}
    \label{cor:rMSEuserlvldeslocal}
    Let $\mathcal A$ denote Algorithm \ref{alg:winsmeanHD} calling Algorithm \ref{alg:winsmean1Dlocal} in Line 3 with input $\hat \mu^n \in \R^{n \times d}$ in Definition \ref{def:userlvldatamatrix}. Then $\mathcal A$ is $(\varepsilon, \varrho)$-\uLDP{} for $\varepsilon, \varrho \in (0,1)$. Further, let $\gamma \in (0, 1 \wedge \frac n 4)$ and let $X^n \in \R^{nT \times d}$ be a user-level data matrix with mean \st{} $\infnorm{\mu} \leq \mathsf B$ that fulfills Assumption \ref{ass:logsobdep} \st{} $\rho M^2 \lesssim 1$. Then, with probability at least $1-3d\gamma-\tilde O(\sqrt{T/\rho} \cdot \mathsf B \cdot d e^{-\kappa n (\varepsilon^\prime)^2})$,
    \begin{align*}
        \Twonorm{\mathcal A(\hat \mu^n) - \mu} 
        \lesssim \sqrt{\frac{d\rho \log(4d/\gamma)}{nT}} + \sqrt{\frac{d^2\rho\log(2dn/\gamma)\log(1/\varrho)\log(3/(d\gamma))^2}{Tn\varepsilon^2}}. 
    \end{align*}
\end{corollary}

Note that the cost of privacy here always dominates the statistical rate of order $\sqrt{d/(nT)}$. While we have an exponentially quickly decaying exponential term in the probability like in the central model, the probability now grows in $T$. Parallel to the user-level mean estimator in \citet{kent:berrett:yu2024}, it is therefore sensible to only consider $T \leq e^{\kappa n (\varepsilon^\prime)^2}$ many observations per user to prevent the probability from blowing up as only $T \to \infty$ while $n$ stays fixed. The same holds for the direct corollary below that applies Corollary \ref{cor:rMSEuserlvldeslocal} in the setting of a user-level random effects location model.

\begin{corollary}
    \label{cor:rMSErandeffectslocal}
    Let $\mathcal A$ denote Algorithm \ref{alg:winsmeanHD} calling Algorithm \ref{alg:winsmean1Dlocal} in Line 3 on input $\hat \mu^n$. $\mathcal A$ is $(\varepsilon, 0)$-\uDP{} for $\varepsilon \in (0,1)$. Let $\gamma \in (0,1 \wedge \frac n 4)$, $n_{g^\star}$ be the maximum group size and let $Y^n \in \R^{nT}$ and $\hat \mu^n$ be as in Definition \ref{def:1drandeffects}. Assuming that $\infnorm{\mu} \leq \mathsf B$, with probability at least $1-3\gamma-\tilde O(\sqrt{T/\rho} \cdot \mathsf B \cdot e^{-\kappa n \varepsilon^2})$,
    \begin{align*}
        \Twonorm{\mathcal A(\hat \mu^n) - \mu} 
        \lessapprox \sqrt{\frac{\sigma^2_U \cdot n_{g^\star}T + \sigma^2}{nT}} + \sqrt{\frac{\sigma^2_U \cdot n_{g^\star}T + \sigma^2}{Tn\varepsilon^2}}. 
    \end{align*}
\end{corollary}

Beware that we suppress log-factors and the cost of privacy again dominates, even when $d = 1$ and $\varepsilon \asymp 1$. Clearly, any privacy budget $\varepsilon \asymp n^{-\omega}$ with $\omega > 0$ decreases the rate below $1/n$. The effect of dependence on estimation is the same as discussed in Remark \ref{rem:rMSErandeffects} covering the central model.

Lastly, we revisit user-level linear regression in Corollary \ref{cor:MSEuserlvlreglocal}.

\begin{corollary}
    \label{cor:MSEuserlvlreglocal}
    Let $\mathcal A$ denote Algorithm \ref{alg:winsmeanHD} calling Algorithm \ref{alg:winsmean1Dlocal} in Line 3 on input $\hat\beta^n$ constructed as in Lemma \ref{lem:concuserlvlreg} from $X^n, Y^n$. Then, $\mathcal A$ is $(\varepsilon, \varrho)$-\uLDP{} for $\varepsilon, \varrho \in (0,1)$. Further, let $\gamma \in (0, 1 \wedge \frac n 4)$. Assume that $X^n, Y^n$ follow the model in Definition \ref{def:linregmodel} with true regression parameter $\beta$ \st{} $\infnorm{\beta} \leq \mathsf B$. Then, with probability at least $1-3p\gamma-\tilde O(\sqrt{T\rho} \cdot \mathsf B \cdot p e^{-\kappa n (\varepsilon^\prime)^2})$,
   \begin{align*}
        \Twonorm{\mathcal A(\hat\beta^n) - \beta} 
        &\lessapprox \sqrt{\frac{p\vartheta\sigma^2}{nT\theta^2}} + \sqrt{\frac{p^2\vartheta\sigma^2}{\theta^2Tn\varepsilon^2}}.
    \end{align*}
\end{corollary}

Note that to keep the probability of the error bound small in Corollary \ref{cor:MSEuserlvlreglocal} we should use only $T \leq e^{\kappa n (\varepsilon^\prime)^2}$ many observations per user. This is again similar to the results in \cite{kent:berrett:yu2024}.
 
\section{Simulations}
\label{sec:sim}

We examine empirically the finite sample guarantees of our noisy Winsorized mean estimator. While our simulations focus on item-level mean estimation in the central model, we also briefly discuss user-level mean estimation and mean estimation under the model of local \DP{}. 

\subsection{Item-Level Mean Estimation}

For our item-level simulations we consider a random sample of dependent univariate Gaussian random variables representing univariate observations of $n$ dependent individuals. More precisely, we have observations of the form $X^n \sim \mathcal N(\mu \one_n,\Sigma^n)$ with $\mu \in \R$ and $\Sigma^n \in \R^{n \times n}$. In this case, $X^n$ is log-Sobolev dependent with constant $\rho=\opnorm{\Sigma^n}$. Hence, we can easily consider both \iid{} observations where $\Sigma^n \asymp I_n$, and weak and strongly dependent $X_i$ where $\opnorm{\Sigma^n} \asymp 1$ and $\opnorm{\Sigma^n} \asymp n^\omega$ for $\omega > 0$. As in our theoretical analysis, we mostly assume the log-Sobolev constant $\rho = \opnorm{\Sigma^n}$ to be known. On top, without loss of generality, we assume that $\Sigma^n_{ii} = 1$ for all $i \in [n]$ to fix a scale. 

\subsubsection{Cost of Privacy in Small Samples}

\begin{figure}[h]
    \centering
    \includegraphics[width=0.9\textwidth]{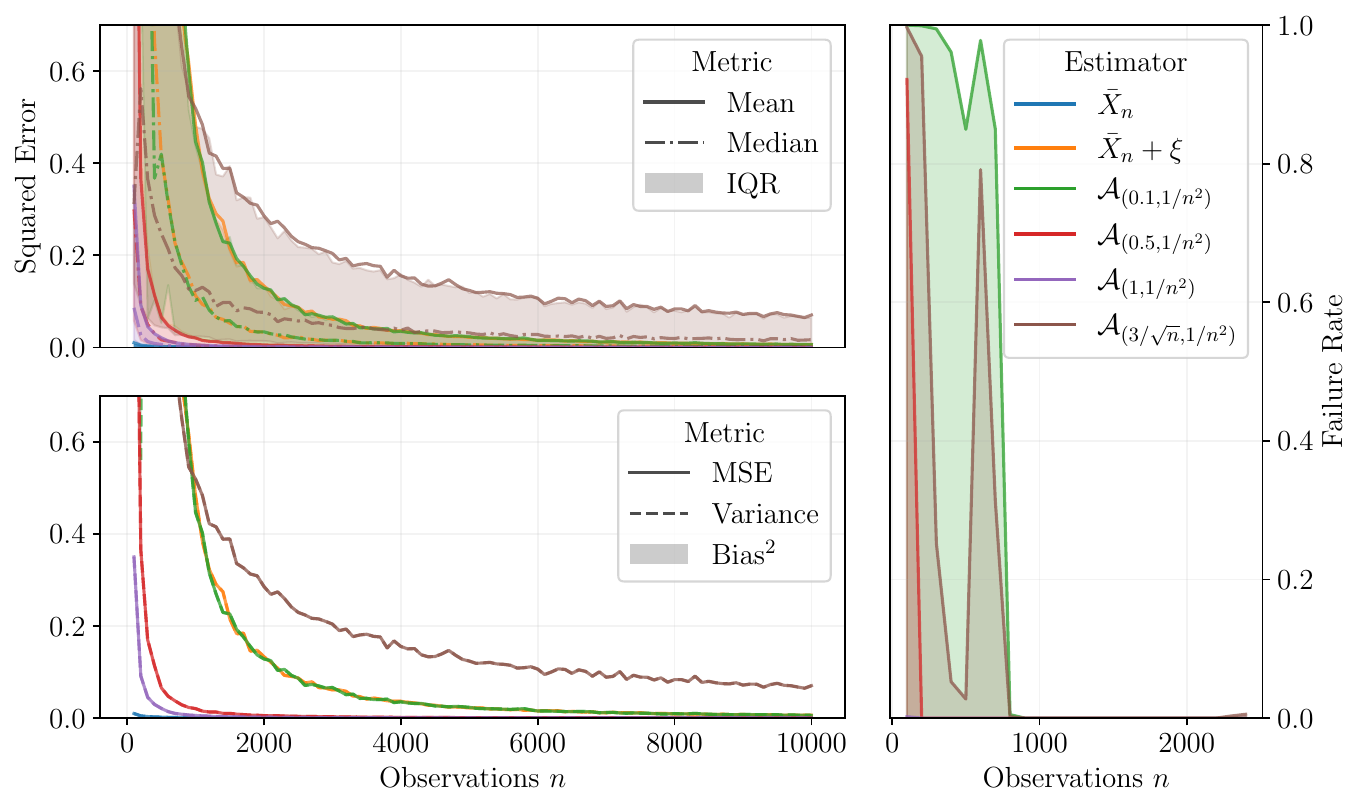}
    \caption{Comparison of the empirical mean $\bar X_n$ with Algorithm \ref{alg:winsmean1D}, denoted $\mathcal A_{(\varepsilon, \delta)}$.} 
    \label{fig:estimators}
\end{figure}

It is well known that for a privacy parameter $\varepsilon \gtrsim 1/\sqrt{n}$ the statistical rate dominates the cost of privacy in mean estimation in the central model. Hence, for $\varepsilon \gtrsim 1/\sqrt{n}$ and in particular for constant $\varepsilon$ we expect the MSE rate of our Winsorized mean estimator to be of order $1/n$. This suggests that privacy can be obtained almost for free in this regime. However, when considering real world applications constants enter the picture and especially on small samples the concrete choice of $\varepsilon$ expresses itself in substantially different MSEs for fixed $n$. To showcase this, Figure \ref{fig:estimators} compares the non-private empirical mean estimator $\bar X_n$ with our private estimator $\mathcal A$ of Algorithm \ref{alg:winsmean1D} parametrized with privacy budget parameters $\varepsilon \in \curly{1/\sqrt{n}, 0.1, 0.5, 1}$ and $\delta = 1/n^2$. The two subfigures on the left depict functionals of the squared errors over $2000$ replications of the empirical measure at increasing sample sizes $n$. Concretely, the upper part depicts the MSE, the median squared error and the interquartile range (IQR) of squared errors. The lower part shows a bias-variance decomposition where the bias squared is given as the pointwise distance between the solid and dashed line for fixed $n$. As expected, increasing $\varepsilon$ lets the private estimators approach the non-private $\bar X_n$. As theory predicts, the estimator with $\varepsilon \asymp 1/\sqrt n$ has a $1/n$ rate throughout. In contrast, when $\varepsilon \in \curly{0.1, 0.5, 1}$ The cost of privacy is of order $1/n^2$ stemming from the added Laplace noise, but for small sample sizes the privacy error can dominate the statistical error of order $1/n$. This interpretation is supported by the non-private $\bar X_n + \xi$ that does not experience truncation but still behaves like $\mathcal A_{(0.1, 1/n^2)}$ and the bias-variance decomposition that shows that the MSE of $\mathcal A$ is almost entirely driven by variance. In fact, the bias is not even visible for $\mathcal A$. The right half of Figure \ref{fig:estimators} presents the empirical failure rate of the stable histogram underlying $\mathcal A$, i.e., the percentage of times Algorithm \ref{alg:privrangestabhist} outputs $\boldsymbol 0$. We observe that this rate rapidly goes to 0 and few non-zero values occur for $n \geq 1000$. 

\subsubsection{Improving Constants Theoretically and Empirically}
\label{sim:improving_constants}

\begin{figure}[h]
    \centering
    \includegraphics[width=0.9\textwidth]{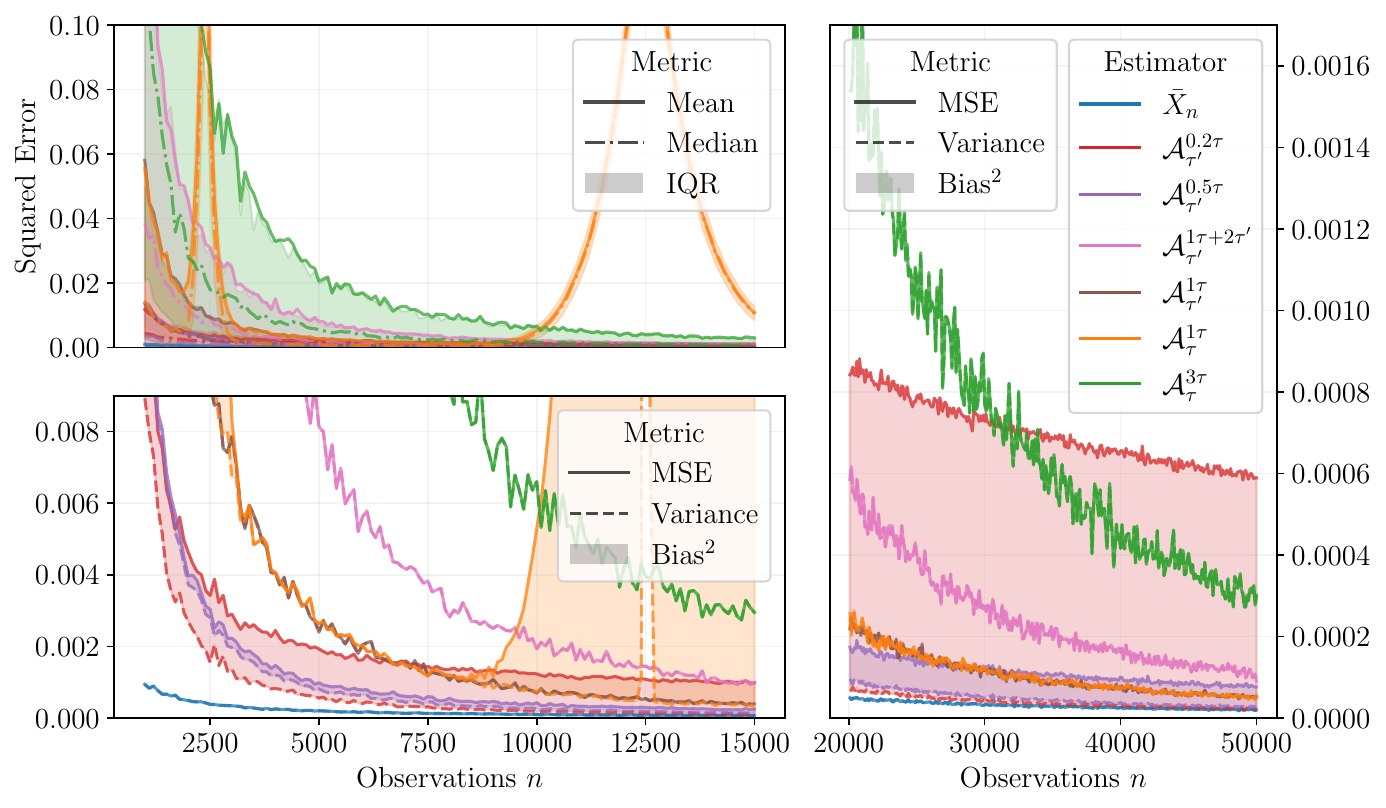}
    \caption{Comparison of $(0.1, 1/n^2)$-\DP{} estimators $\mathcal A_h^\kappa$ where $\kappa$ is the radius of the projection interval and $h$ the length of the histogram bins. Note that here $\tau^\prime \asymp \sqrt{\log(2/\gamma)}$ and $\tau \asymp \sqrt{\log(2n/\gamma)}$.}
    \label{fig:taus}
\end{figure}

We revisit our projection interval construction in order to further reduce the cost of privacy. In Remark \ref{rem:tauprime} we noted that $(\tau, \gamma)_\infty$-concentration of $X^n \in \R^n$ is not necessary for Lemma \ref{lem:privmidpointstabhist} to hold. Instead, for the private midpoint algorithm to output a crude mean estimate not further than $2\tau^\prime$ from $\mu$, it suffices that each $X_i$ is $(\tau^\prime, \gamma)_\infty$-concentrated with $\gamma \in (0,\frac 1 4)$. More concretely, if $X^n$ is log-Sobolev dependent with constant $\rho$, $(\tau^\prime)^2 \asymp \rho \log(2/\gamma)$ is $\rho\log(n)$ smaller than $\tau^2 \asymp \rho \log(2n/\gamma)$. Such a decrease in the size of the projection interval immediately yields privacy noise with smaller variance and therefore smaller constants in the MSE upper bounds and also simulations as can be seen in Figure \ref{fig:taus}. This is obtained through bins of width $2\tau^\prime$ in the histogram. Carrying this change through the analysis leads to $\hat I = [\hat m \pm \tau + 2\tau^\prime]$ instead of $\hat I = [\hat m \pm 3\tau]$ in Algorithm \ref{alg:winsmean1D} to ensure that no $X_i$ is affected by the projection with high probability. On top, we also exploit Remark \ref{rem:taugammamaximal} and use the marginal log-Sobolev constants. 

We further optimize constants empirically. For this, we fix privacy budget parameters $(\varepsilon, \delta) = (0.1, 1/n^2)$ and run Algorithm \ref{alg:winsmean1D} with different projection intervals $\hat I$ and potentially adjusted binwidths for the underlying histogram estimator. Firstly, this lets us empirically assess the impact of choosing the projection interval $\hat I = [\hat m \pm \tau + 2\tau^\prime]$ instead of $\hat I = [\hat m \pm 3\tau]$. Secondly, we experiment with different combinations of binwidths for the stable histogram and even smaller projection intervals. In Figure \ref{fig:taus} we compare Winsorized mean estimators $\mathcal A_h^\kappa$ where $\kappa > 0$ indicates the radius of the projection interval $\hat I$ and $h$ the binwidth used by the histogram. Note that the resulting algorithms are not necessarily grounded in theory, but $\mathcal A^{3\tau}_\tau$ and $\mathcal A^{\tau + 2\tau^\prime}_{\tau^\prime}$ are. Comparing these two, we observe that our theoretical approach to reducing constants already considerably decreases the MSE. Our experiments go beyond our theory with the private Winsorized estimators $\mathcal A_{\tau^\prime}^{\kappa}$ for $\kappa \in \curly{1\tau, 0.5\tau, 0.2\tau}$. The subfigure on the lower left of Figure \ref{fig:taus} suggests that an interval length of $1\tau$ still works empirically and results in even smaller MSE. For an interval length of $0.5\tau$, the same subfigure suggests that introducing reasonable bias through a more aggressive projection decreases the MSE for the $n \leq 50,000$ we consider, but for $0.2\tau$ the bias already dominates for small $n$ and cannot be offset by the decreased variance. The right hand side of the figure suggests that introducing bias by shrinking the length of $\hat I$ is particularly beneficial for small sample sizes. Yet, for bigger $n$ even bias that is at first reasonable ends up pushing the MSE above that of estimators with small or no bias. 

Our empirical results indicate that our bounds are sharp in the sense that shrinking the bin width aggressively beyond our theory eventually worsens the rate.
Indeed, Algorithm $\mathcal A^{1\tau}_\tau$ shows that a less principled approach where we simply shrink the length of the projection interval without adjusting the bin width fails as it introduces excess bias and variance. As the edge between some bins $B_k$ and $B_{k+1}$ approaches the true mean $\mu$ from the left as $n$ grows, the midpoint of $B_{k+1}$ that initially contains $\mu$ moves away from $\mu$ and for symmetric distributions like our Gaussian, up to $50\%$ of the data are eventually projected. This explains that the bias first increases and then decreases again when $B_k$ has more mass and its midpoint approaches $\mu$ from the left. When the edge is almost exactly at $\mu$, the estimator's variance explodes because single observations let the midpoint $\hat m$ jump between the midpoints of $B_k$ and $B_{k+1}$. 

\subsubsection{Cost of Dependence in Small Samples}

\begin{figure}[h]
    \centering
    \includegraphics[width=0.9\textwidth]{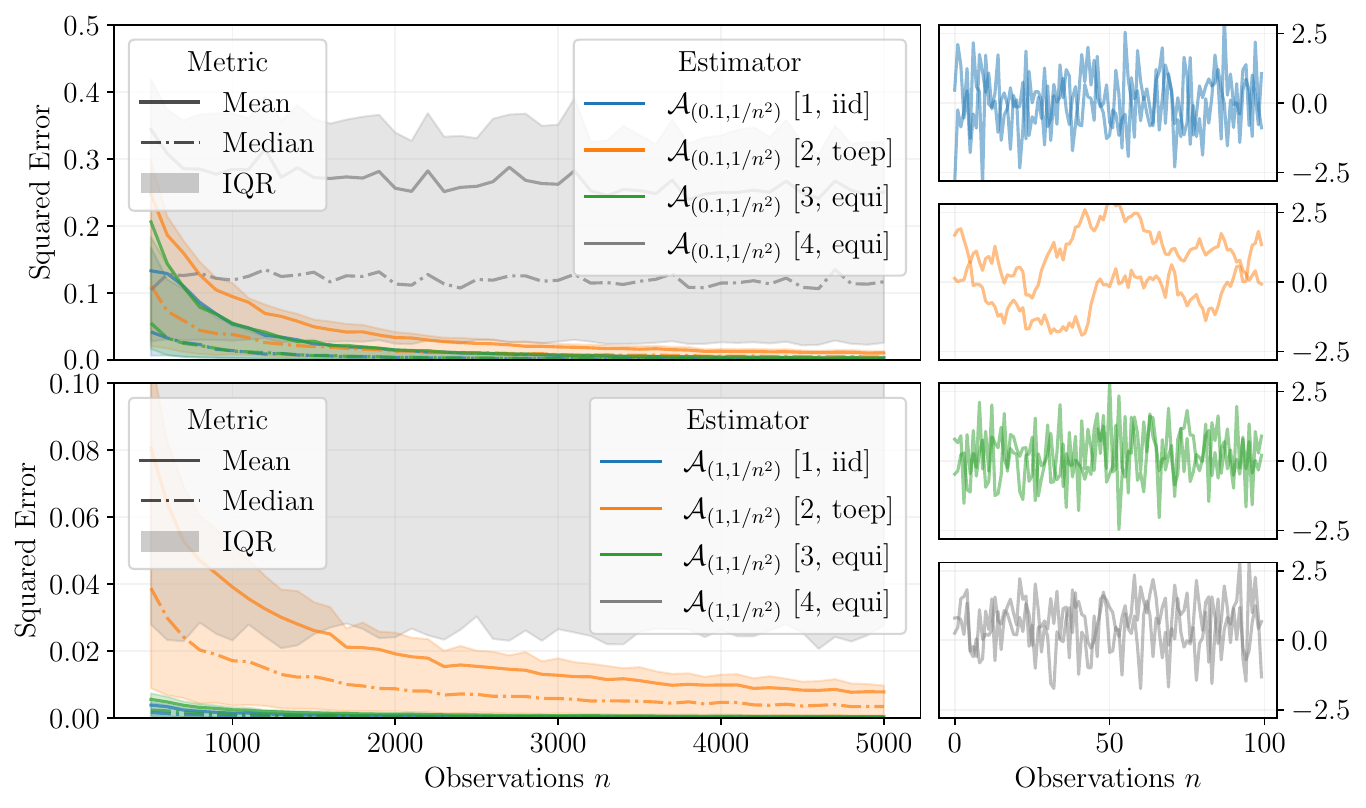}
    \caption{$(\varepsilon, \delta)$-\DP{} estimators $\mathcal A_{(\varepsilon, \delta)}$ on $X^n \in \R^n$ with different covariances $\Sigma^n \in \R^{n\times n}$.}
    \label{fig:dependence}
\end{figure}

Our theory predicts that the main driving factor of the MSE rate under log-Sobolev dependence is the log-Sobolev constant $\rho = \opnorm{\Sigma^n}$. Thus, whenever $\opnorm{\Sigma^n}$ is constant, our estimators should have the same $1/n$ or $1/(n^2\varepsilon^2)$ but MSEs differing in constants. At the same time, when $\opnorm{\Sigma^n} \asymp n$ we expect the MSE to converge to a non-zero value. To verify this, Figure \ref{fig:dependence} depicts the MSE, median squared error and IQR of squared errors for $(\varepsilon, \delta)$-\DP{} estimators $\mathcal A_{(\varepsilon, \delta)}$ on observations $X^n$ with $\Sigma^n = I_n$, $\Sigma^n$ a Toeplitz covariance matrix with decay 0.95, i.e., $\Sigma^n = \parent{0.95^{|j-i|}}_{i,j \in [n]}$ and two equi-covariance matrices $\Sigma^n$ with variance and covariance pairs $(1, 1/(n-1))$ and $(1, 1/4)$, respectively. We depict two samples from each of these models in the four subfigures on the right of Figure \ref{fig:dependence}. On the left, the upper subplot depicts a high privacy regime where $(\varepsilon, \delta) = (0.1, 1/n^2)$. Here, we see that the three weakly dependent settings where $\opnorm{\Sigma^n} \asymp 1$ behave similarly and have comparable squared errors. In turn, in the lower subplot where $(\varepsilon, \delta) = (1, 1/n^2)$, the three weakly dependent settings have considerably different squared errors and the Toeplitz model far exceeds the others. To understand this note that Remark \ref{rem:taugammamaximal} implies that the cost of privacy term only depends on the marginal log-Sobolev constants, which here are the $\Sigma_{ii}^n = 1$. The statistical rate, in turn, always depends on $\opnorm{\Sigma^n}$ which, e.g., for the Toeplitz model is $\approx 38$. When $\varepsilon = 0.1$, the privacy term seems to dominate whereas for $\varepsilon = 1$, it is small enough so that the different constants in the statistical rate are visible. In both subplots, the strongly dependent observations in the second equi-covariance model have MSE converging to $1/4$ with rate $1/n$. 

\subsubsection{Central versus Local Model}
\label{sim:DP_LDP}

\begin{figure}[h]
    \centering
    \includegraphics[width=0.9\textwidth]{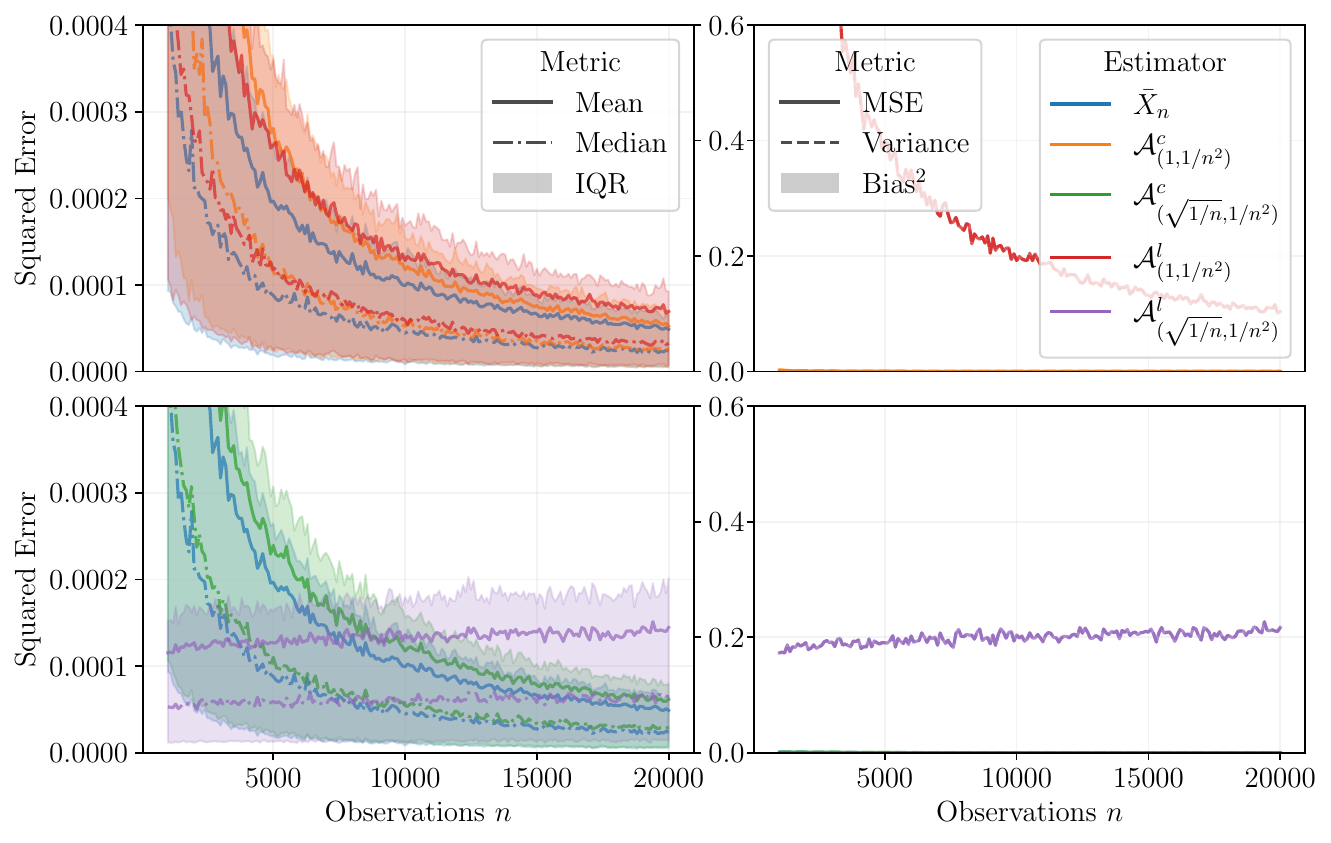}
    \caption{Comparison of $(\varepsilon, \delta)$-\DP{} and \LDP{} estimators $\mathcal A^c_{(\varepsilon, \delta)}$ and $\mathcal A^l_{(\varepsilon, \delta)}$.}
    \label{fig:local}
\end{figure}

In the central and the local model, the privacy budget parameter $\varepsilon$ directly impacts the MSE-rate. Ignoring constants and logarithmic terms our theory predicts that the cost of privacy is $1/(n^2\varepsilon^2)$ and $1/(n\varepsilon^2)$, respectively. Thus, when $\varepsilon$ is constant in $n$, both models of privacy have the same $1/n$-estimation rate. This changes as $\varepsilon \asymp 1/n^\omega$ for $\omega > 0$. Then, our theory predicts that in the central model for $0 \leq \omega \leq 1/2$ the statistical rate dominates the cost of privacy. In the local model, however, for any $\omega > 0$ the cost of privacy should immediately dominate the statistical rate. We investigate this difference in the simulations in Figure \ref{fig:local}. The left hand side of the plot uses two different axes where $\bar X_n$ and the central estimators $\mathcal A^c$ use the left axis and the local estimators $\mathcal A^l$ are plotted on the middle axis that is shared with the right hand side of the figure. There, we plot the estimators on their true scale. This layout allows us to compare the rate and the magnitude of the MSEs. We use $\varepsilon = 1$ in the two subfigures on top. While we observe a rate of $1/n$ in both models of \DP{}, the right hand side shows that constants are substantially larger when privatizing locally. In the bottom, $\varepsilon \asymp \sqrt{1/n}$ and we still observe an MSE-rate of $1/n$ in the central model whereas the MSE in the local model is constant or even slightly increasing due to logarithmic terms. 

\subsubsection{Plug-in Variance Estimation}
\label{subsubsec:pluginvar}

\begin{figure}[h]
    \centering
    \includegraphics[width=0.9\textwidth]{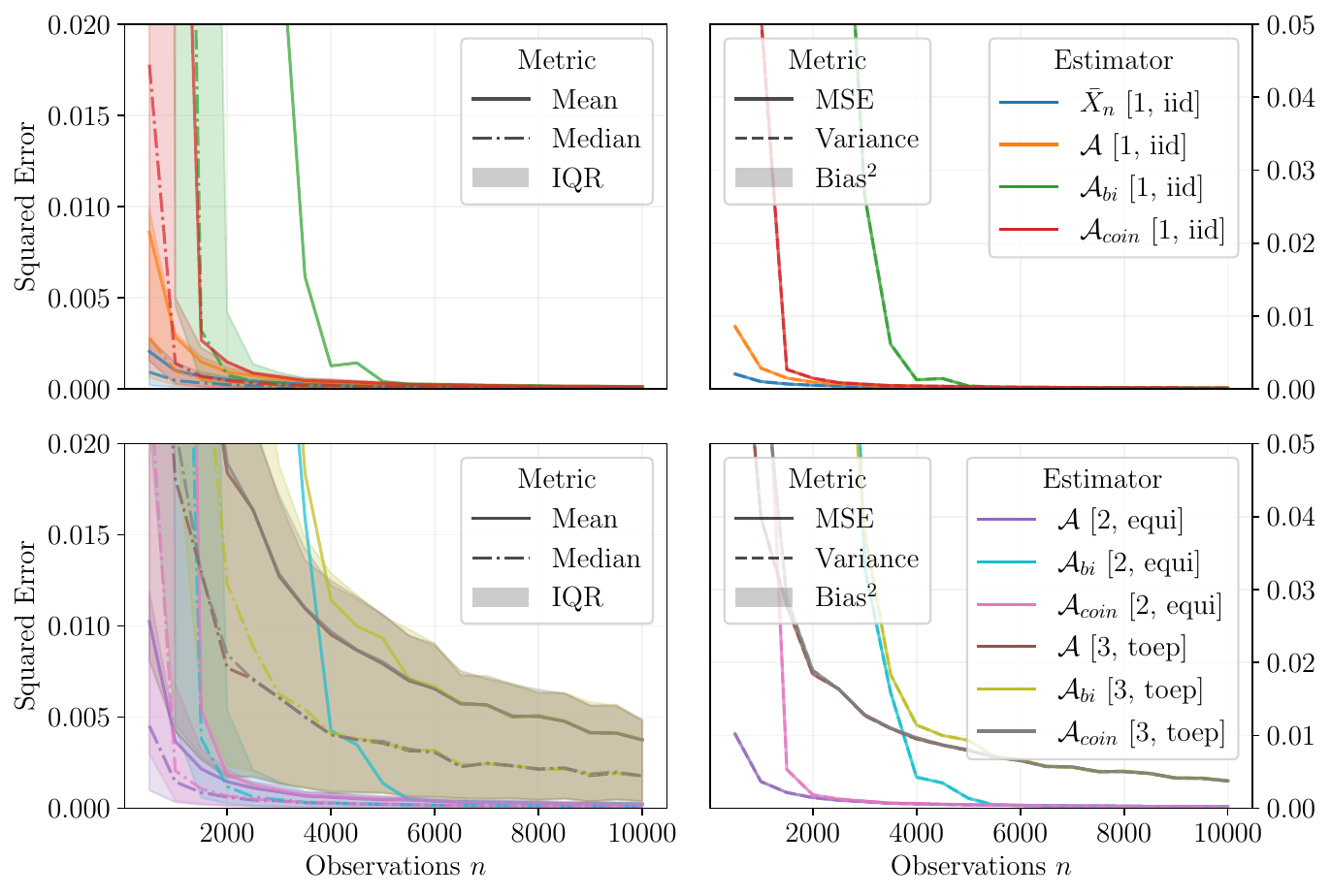}
    \caption{Comparison of the empirical mean $\bar X_n$ with $(1, 1/n^2)$-\DP{} estimator $\mathcal A$ using known variance and $(2, 1/n^2)$-\DP{} estimators $\mathcal A_{bi}$ and $\mathcal A_{coin}$ using plug-in variance estimators.}
    \label{fig:variance}
\end{figure}

So far we assumed that $\rho = \opnorm{\Sigma^n}$ or more precisely the $\Sigma_{ii}$ are known a priori. To make our estimator more applicable, we now present a way to estimate the concentration radius of the data $\tau$ in a private and data-driven way. For this, we rely on Remark \ref{rem:taugammamaximal} which tells us that to compute $\tau$ it is enough to know the marginal sub-Gaussian variance proxies or log-Sobolev constants $\rho_i$ corresponding to the distribution of $X_i \in \R$ for all $i \in [n]$. For Gaussian observations that we investigate here, these constants are the marginal variances. We therefore introduce two private variance estimators that we use as plug-ins for the $\rho_i = \Sigma_{ii}$. Both get a very rough estimate of the first moment, and upper and lower bounds of the variance as input and iteratively and alternatingly refine them. For the variance step, Algorithm \ref{alg:bisection} in Appendix \ref{app:plugin} implements a bisection that starts with the upper bound on the variance and halves it unless the lower bound is reached or a confidence interval around the first moment has too low coverage. The variance step of Algorithm \ref{alg:coinpress} in Appendix \ref{app:plugin} is that of the CoinPress Algorithm due to \citet{biswas:dong:kamath:ullman2020}. We call the resulting Winsorized mean estimators with these plug-in variance estimators $\mathcal A_{bi}$ and $\mathcal A_{coin}$, respectively. Figure \ref{fig:variance} showcases their performance with privacy budget parameters $(\varepsilon, \delta) = (1, 1/n^2)$ each for the mean and the variance estimation. For the true mean and variance we use $\mu = 100$ and $\sigma^2 = 1$ whereas the provided rough first moment estimate is 300 and the upper and lower bounds on the variance are $(0.1, 10000)$. We use three dependence models once with an \iid{}, equi-covariance and Toeplitz covariance matrix: $\Sigma^n = I_n, \Sigma^n = 3/4 \cdot I_n + 1/4 \cdot \one_n\one_n\T$ and $\Sigma^n = \parent{0.95^{|j-i|}}_{i,j \in [n]}$. The small difference between $\bar X_n$ and the $(1, 1/n^2)$-DP estimator $\mathcal A$ using the known variance suggests that we are in a regime with low cost of privacy for mean estimation. Keeping in mind that we allow for an additional privacy budget for the variance estimation, the adjusted CoinPress clearly outperforms the bisection for smaller $1000 \leq n \leq 6000$. For large enough $n \geq 6000$, the performance of the plug-in estimators is almost indistinguishable from that of $\mathcal A$. This suggests that there the cost of privacy paid for variance estimation is low. 

\subsection{User-Level Mean Estimation}

\begin{figure}[h]
    \centering
    \includegraphics[width=0.9\textwidth]{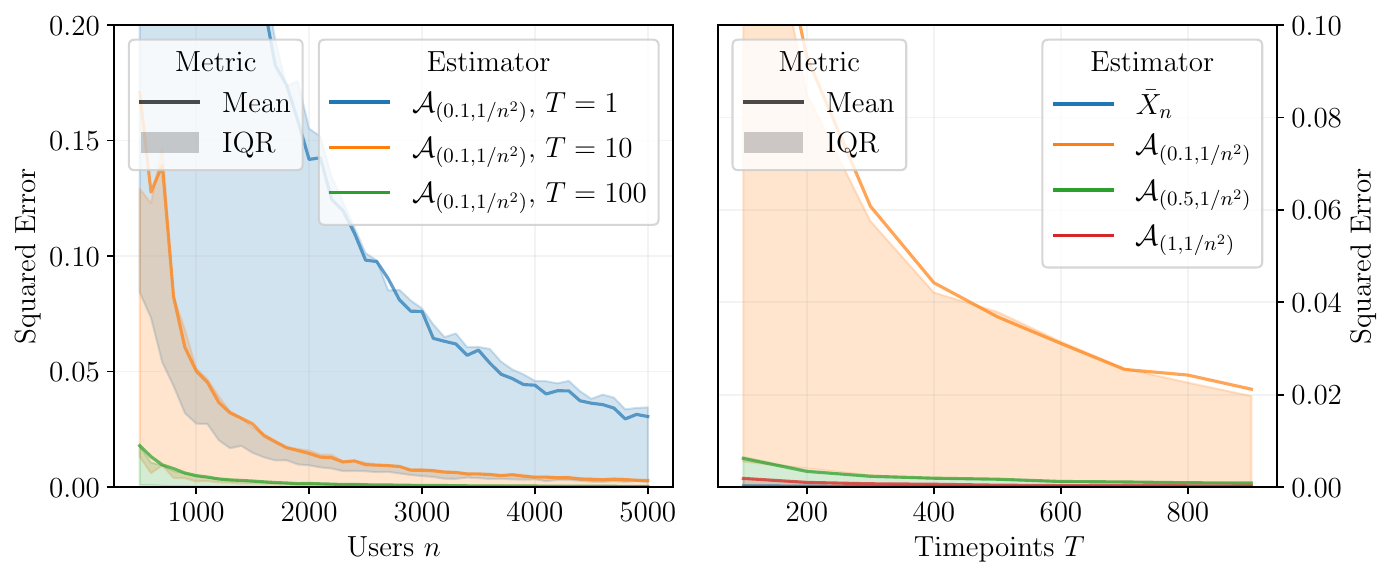}
    \caption{MSE of user-level estimators as a function of users and timepoints.}
    \label{fig:user}
\end{figure}

In our user-level simulations we assume that $n$ users each contribute $T$ observations collected in $X_u \in \R^T$. For the resulting user-level data matrix $X^n := [X_1\T,\dots,X_n\T]\T \in \R^{nT}$ we want to assume log-Sobolev dependence. For our simulations we therefore suppose that $X^n \sim \mathcal N(\mu \one_{nT}, \Sigma^n)$ where $\Sigma^n \in \R^{nT \times nT}$. Throughout, we assume that $\Sigma^n = \diagonal{\Sigma_1,\dotsm, \Sigma_n}$ where $\Sigma_u$ are Toeplitz covariance matrices with decay 0.95. This means that we model users to be independent but their repeated measurements to have decaying covariances over time. The results are depicted in Figure \ref{fig:user}. The plot's left hand side shows the MSEs over $k=1000$ simulations of the $(0.1, 1/n^2)$-DP estimator $\mathcal A_{(0.1, 1/n^2)}$ as a function of $n$ for three different numbers of timepoints $T \in \curly{1, 10, 100}$ per user. The right hand side, in turn, fixes the number of users as $n = 1000$ and plots the MSE as a function of $100 \leq T \leq 1000$. Here, we also consider the estimator $\mathcal A_{(\varepsilon, 1/n^2)}$ for three different privacy budgets $\varepsilon \in \curly{0.1, 0.5, 1}$. As expected, the left hand side in this weakly dependent setting shows that increasing $T$ by one order of magnitude has a similar effect on reducing the MSE as decreasing the variance in an item-level setting by one order of magnitude. In the right, we observe the $1/T$ rate of our Winsorized estimator we expect when $n$ is held constant. 

\section{Final remarks}

We have provided some first differentially private estimation tools for dependent data. The core procedure is a mean estimation algorithm based on noisy Winsorized means that had been introduced by \cite{karwa:vadhan2018} in the context of item-level DP for univariate \iid{} Gaussian data. Our work shows that simple variants of this algorithm cannot only handle dependent data, but can also lead to optimal estimation rates in the context of user-level DP and local DP for various estimation problems such as nonparametric regression, random effects models and longitudinal linear regression. In this sense, our work also provides a simple unifying estimation framework for a wide range of problems in DP that tend to be studied separately. 

From a technical perspective, the key tool allowing for log-Sobolev dependent observations is a DKW-type inequality due to \cite{bobkov:götze2010}. It provides us a way to control the performance of our intermediate histogram estimators which are essential for adaptively finding appropriate projection intervals for our Winsorized mean estimators. Further relaxing the log-Sobolev dependence assumption would require DKW-inequalities under weaker conditions.

Our work opens numerous natural future research directions. Perhaps the most pressing one in practice is to develop inference tools that work well with moderate sample sizes under dependence. For this it will be important to develop good private variance estimators and we believe that ideas based on the CoinPress procedure of \cite{biswas:dong:kamath:ullman2020} are promising.
Finally, some of the many important statistical questions that naturally involve dependent data include multi-armed bandits, online learning, federated learning and in general noisy DP optimization procedures. We hope to address some of these in future work.

\appendix

\section{Auxiliary Results}
\label{app:aux}

\begin{theorem}{(Bakry-Emery, Theorem 21.2, \citet{villani2009}, Corollary 5.7.2, \citet{bakry:gentil:ledoux2014}).}
    \label{thm:bakryemery}
    Let the probability measure $\nu$ on $\R^n$ have density $p(x) \propto \exp(-V(x))$ for all $x \in \R^n$. Suppose there is $\alpha > 0$ \st{} $\nabla^2 V(x) \succeq \alpha I_n$ for all $x \in \R^n$. Then, the following log-Sobolev inequality holds: 
    \begin{align*}
        \ent[\nu]{f^2} 
        \leq \frac 2 \alpha \cdot \E[\twonorm{\nabla f(X)}^2].
    \end{align*}
\end{theorem}

Theorem \ref{thm:bakryemery} states a sufficient condition for a log-Sobolev inequality. It is not a necessary condition as illustrated, e.g., by the Holley-Stroock Perturbation Theorem (\citet[Proposition 2.3.1]{chewi2025}). 

\begin{example}{(Gaussian Log-Sobolev Inequality via Bakry-Emery).}
    \label{exa:gausslogsob}
    Let $\mu \in \R^d$ and $\Sigma \in \R^{d\times d}$ \st{} $\Sigma \preceq \beta I_d$. A multivariate Gaussian measure with mean $\mu$ and covariance matrix $\Sigma$ has the following density $p$ with respect to Lebesgue measure
    \begin{align*}
        p(x) 
        \propto \exp\Parent{-\frac 1 2 (x-\mu)\T \Sigma\inv (x-\mu)}. 
    \end{align*}
    Computing the negative log-Hessian now yields $\nabla_x^2 V(x) = -\nabla^2_x \log(p(x)) = \Sigma\inv \succeq 1/\beta I_d$ due to the covariance bound $\Sigma \preceq \beta I_d$. Hence, by Theorem \ref{thm:bakryemery}, the measure $\mathcal N(\mu, \Sigma)$ fulfills $\operatorname{LSI}(\beta)$. 
\end{example}

The following stability result for log-Sobolev inequalities is useful for our purposes as it implies that marginals of distributions fulfilling a log-Sobolev inequality also fulfill one with the same constant. 

\begin{lemma}{(Lipschitz Contractions for Log-Sobolev Inequality, \citet[p. 1492]{colombo:figalli:jhaveri2017}).}
    \label{lem:lipcontlogsob}
    Let $\nu$ on $\R^d$ fulfill $\LSI{\rho}$ and $T : \R^d \to \R^n$ be L-Lipschitz. Then, for the push-forward measure $\nu = T_\# \mu$ on $\mathcal \R^n$ and $\varphi : \R^n \to \R$ \st{} $\ent[\nu]{\varphi^2} < \infty$ and $\nabla\varphi \in L^2(\nu)$ we have
    \begin{align*}
        \ent[\nu]{\varphi} \leq 2\rho L^2 \cdot \E_\nu[\Twonorm{\nabla_y \varphi(Y)}^2]. 
    \end{align*}
\end{lemma}

\begin{remark}
    Let $T : \R^d \to \R^{k}$ be the map corresponding to marginalization, i.e., dropping $d-k$ coordinates. Let the set of indices of the remaining coordinates be $\mathcal M$. Then, $T$ is 1-Lipschitz: 
    \begin{align*}
        \twonorm{T(x) - T(y)}
        = \sqrt{\sum_{i \in \mathcal M} (x_i - y_i)^2}
        \leq \sqrt{\sum_{i \in [d]} (x_i - y_i)^2}
        = \twonorm{x - y}, 
        \quad \forall x,y \in \R^d.
    \end{align*}
    Hence, any marginal of a measure fulfilling a log-Sobolev inequality has the same $\rho$. 
\end{remark}

The next two results follow from Gaussian and Laplace Lipschitz concentration respectively.

\begin{lemma}
    \label{lem:gaussconciidempmean}
    Let $X_1,\dots ,X_n \in \R^d$ and $X_i \sim \mathcal N(\mu, \Sigma_i)$ independently. Then, with $\bar \Sigma_n := \frac 1 n \sum_{i=1}^n \Sigma_i$, 
    \begin{align*}
        \P\Parent{\Twonorm{\frac 1 n \sum_{i=1}^n X_i - \mu} \geq \sqrt{\frac{\trace(\bar \Sigma_n)}{n}} + \sqrt{\frac{2\opnorm{\bar \Sigma_n} \log(1/\alpha)}{n}}} 
        \leq \alpha. 
    \end{align*}
\end{lemma}

\begin{proof}
    This well-known result immediately follows from the affine transformation property of Gaussians combined with Gaussian Lipschitz concentration \citep[Theorem 5.6]{boucheron:lugosi:massart2013} applied to the $L_2$-norm. 
\end{proof}

\begin{lemma}
    \label{lem:normlapvectors}
    Let $\Xi_1,\dots ,\Xi_n \sim \operatorname{Lap}(0,bI_d)$ independently. Then, it holds that
    \begin{align*}
        \P\Parent{\Twonorm{\frac 1 n \sum_{i=1}^n \Xi_i} \geq b \Parent{\sqrt{\frac{2d}{n}} + \sqrt{\frac{4\log(3/\alpha)^2}{n}}}} \leq \alpha. 
    \end{align*}
\end{lemma}

\begin{proof}
    This proof collects $\Xi := [\Xi_1,\dots ,X_h]\T \sim \operatorname{Lap}(0, b I_{dh})$ and expresses the average in terms of $\Xi$. This allows us to apply Poincar{\'e} Lipschitz concentration \citep[Theorem 2.4.3]{chewi2025}. 

    Concretely, we write $\bar \Xi_n = A\Xi$ where $A = \frac 1 n [I_d,\dots ,I_d] \in \R^{d\times dn}$ and $\opnorm{A} = 1/\sqrt{n}$. We now show that $\twonorm{\bar \Xi_n}$ is Lipschitz. For $\Xi, \Xi^\prime \in \R^{dn}$ and by the reverse triangle inequality, 
    \begin{align*}
        \Abs{\twonorm{\bar \Xi_n} - \twonorm{\bar \Xi_n^\prime}}
        = |\twonorm{A \Xi} - \twonorm{A\Xi^\prime}|
        \leq \twonorm{A(\Xi - \Xi^\prime)}
        \leq \opnorm{A} \twonorm{\Xi - \Xi^\prime}
        \leq \frac{1}{\sqrt{n}} \twonorm{\Xi - \Xi^\prime}.
    \end{align*}

    Further, by concave Jensen's $\E[\twonorm{\bar \Xi_n} \leq \sqrt{\E[\twonorm{\bar \Xi_n}^2]} \leq b \sqrt{2d/n}$ where the last inequality holds since
    \begin{align*}
        \E\Bracket{\twonorm{\bar \Xi_n}^2}
        &= \E\Bracket{\Twonorm{\frac 1 n \sum_{i=1}^n \Xi_i}^2}
        = \E\Bracket{\sum_{j=1}^d \Parent{\frac 1 n \sum_{i=1}^n \Xi_{ij}}^2}
        = \frac{d}{n^2} \cdot \E\Bracket{\Parent{\sum_{i=1}^n \Xi_{i1}}^2} \\
        &= \frac{d}{n^2} \cdot \sum_{i=1}^n \sum_{k=1}^n \E\Bracket{\Xi_{i1} \Xi_{k1}}
        = \frac{d}{n^2} \cdot \sum_{i=1}^n \E\Bracket{\Xi_{i1}^2} 
        = \frac{d}{n} \cdot \var{\Xi_{i1}}
        = \frac{d}{n} \cdot 2b^2.
    \end{align*}

    Note that the distribution of $\Xi$ is a product distribution with Poincar{\'e} constant $4b^2$ \citep[Exercise 3.22 + Theorem 3.1]{boucheron:lugosi:massart2013}. Hence, for $t\geq 0$ Poincar{\'e} Lipschitz concentration yields
    \begin{align*}
        \P\Parent{\twonorm{\bar \Xi_h} \geq b\sqrt{\frac{2d}{h}} + t}
        \leq \P\Parent{\twonorm{\bar \Xi_h} \geq \E[\twonorm{\bar \Xi_h}] + t}
        \leq 3\exp\Parent{-\frac{\sqrt{h}t}{2b}}. 
    \end{align*}

    A choice of $t = \frac{2b}{\sqrt n} \log(3/\alpha)$ then recovers the statement. 
\end{proof}

\section{Proofs of Section \ref{sec:prelim}}
\label{app:relim}

\begin{proof}[Proof of Corollary \ref{cor:varlogsobdes}]\hfill 

    The $\LSI{\rho}$ condition on $X^n_{\cdot j}$'s distribution implies a Poincar{\'e} constant $\rho$. Thus, for $\varphi : \R^n \to \R$, 
    \begin{align*}
        \var{\varphi(X^n_{\cdot j}})
        \leq \rho \cdot \E[\twonorm{\nabla \varphi(X^n_{\cdot j})}^2]. 
    \end{align*}

    For the first statement we use the test function $\phi(X^n_{\cdot j}) = \theta\T X^n_{\cdot j}$ with $\theta \in \R^n$. Hence, for all $\theta \in \R^n$:
    \begin{align*}
        \theta\T \cov{X^n_{\cdot j}}\theta \leq 
        \rho \cdot \E[\theta\T \theta]
        = \theta\T (\rho I_n) \theta. 
    \end{align*}

    The Loewner ordering follows by definition. To show the second statement we evaluate with another test function. For each $j \in [d]$ define $f(X^n_{\cdot j}) := (\bar X_n)_j = \frac 1 n \sum_{i=1}^n X_{ij}$. For $x,y \in \R^n$, 
    \begin{align*}
        | f(x) - f(y) |
        \leq \frac 1 n \Abs{\sum_{i=1}^n x_i - y_i}
        \leq \frac 1 n \sum_{i=1}^n \abs{x_i - y_i}
        \leq \frac{1}{\sqrt n} \twonorm{x - y}. 
    \end{align*}

    Hence, $f$ is $1/\sqrt{n}$-Lipschitz. By definition of the variance of a random vector, we have
    \begin{align*}
        \E\bracket{\twonorm{\bar X_n - \E[\bar X_n]}^2}
        = \sum_{j=1}^d \E[\abs{(\bar X_n)_j - \E[(\bar X_n)_j]}^2]
        = \sum_{j=1}^d \var{(\bar X_n)_j}. 
    \end{align*}
    
    Evaluating the Poincar{\'e} inequality with the test function $f$ yields
    \begin{align*}
        \var{(\bar X_n)_j}
        = \var{f(X^n_{\cdot j}})
        \leq \rho \cdot \E[\twonorm{\nabla f(X^n_{\cdot j})}^2]
        \leq \frac{\rho}{n}. 
    \end{align*}
\end{proof}

\begin{proof}[Proof of Lemma \ref{lem:logsobconcmean}]\hfill 

    The proof relies on log-Sobolev Lipschitz concentration applied to a dimension-wise estimator $(\bar X_n)_l = \frac 1 n \sum_{j=1}^n X_{jl} = \frac 1 n \one \T X^n_{\cdot l} =: f(X^n_{\cdot l})$ of $\mu$ and union bounding over dimensions.

    To apply Lipschitz concentration, we first compute the mean of $f$ that is $\E[f(X^n_{\cdot l})] = \frac 1 n \sum_{i=1}^n \E[X_{il}] = \mu_l$. Second, $f$ is $1/\sqrt n$-Lipschitz as shown in the proof of Corollary 
    \ref{cor:varlogsobdes}. 
    
    By Assumption \ref{ass:logsobdep} for all $l\in [d]$, the vector $X^n_{\cdot l}$ is distributed according to a distribution that is $\LSI{\rho}$. By log-Sobolev Lipschitz concentration of Theorem \ref{thm:logsoblipconc} we have: 
    \begin{align*}
        \P\Parent{\left|(\bar X_n)_l - \mu_l\right| \geq t}
        = \P\Parent{|f(X_{\cdot l}) - \E[f(X_{\cdot l})]| \geq t}
        \leq 2\exp\Parent{-\frac{nt^2}{2\rho}}. 
    \end{align*}

    Through the choice $t := \sqrt{2 \rho \log(2d/\alpha)/n}$ and a union bound over dimensions we obtain: 
    \begin{align*}
        \P\Parent{\Infnorm{\bar X_n - \mu} \geq \sqrt{\frac{2\rho \log(2d/\alpha)}{n}}}
        = \P\Parent{\exists l \in [d] : |(\bar X_n)_l - \mu_l| \geq \sqrt{\frac{2\rho \log(2d/\alpha)}{m}}} 
        \leq \alpha. 
    \end{align*}

    Using that $\frac{1}{\sqrt d}\twonorm{x} \leq \infnorm{x}$ for all $x \in \R^d$ on the complement event yields the statement. 
\end{proof}

\section{Proofs of Section \ref{sec:hist}}
\label{app:hist}

\subsection{Proofs for Stable Histogram}

We briefly describe the main idea behind the proof of the privacy guarantee stated in Lemma \ref{lem:privhistlearner}. The stability mechanism of the histogram estimator releases all $\boldsymbol{\tilde p} = (\dots, \tilde p_{-1}, \tilde p_0, \tilde p_1,\tilde p_2,\dots )$ which by construction contains at most $n$ non-zero elements. This is not trivial since if we only add noise to the bins $B_k$ with $\hat p_k >0$, even though we protect the counts, we release information about where these bins are. We cannot fix this by adding unbounded noise to all bins $(B_k)_{k \in \Z}$, as this blows up the estimation error. Therefore, the algorithm zeroes out the unstable bins whose noisy count is comparable to one plus the $(1-\delta/2)$-quantile of a standard Laplace. This stability mechanism prevents leaking the location of non-zero bins at the cost of a $\delta>0$ in the privacy guarantee. 

\begin{lemma}
{(Adjusted from Theorem 7.3.5, \citet{vadhan2017}).}
    \label{lem:privhistlearner}
    Algorithm \ref{alg:stablehist} is $(\varepsilon, \delta)$-DP for $\varepsilon, \delta > 0$. 
\end{lemma}

\begin{proof}

    The proof exploits that for two neighboring datasets only two empirical frequencies $\hat p_k$ can be affected. If the corresponding bins both contain at least one data point, a Laplace mechanism is used contributing the $\varepsilon$ and otherwise thresholding yields the $\delta$. 

    Let $\mathcal A$ denote the histogram learner of Algorithm \ref{alg:stablehist}. For two non-random $x, x^\prime \in \R^n$ with $d_H(x,x^\prime) \leq 1$ we want to show that $\mathcal A$ is $(\varepsilon, \delta)$-DP, i.e., for all measurable $S$,
    \begin{equation}
        \label{eq:diffpriv}
        \P\Parent{\mathcal A(x) \in S}
        \leq e^\varepsilon \P\Parent{\mathcal A(x^\prime) \in S} + \delta. 
    \end{equation}
    Remember $\hat p_k=\hat p_k(x) = \frac 1 n \sum_{i=1}^n \indicator{x_i \in B_k}$ and $\tilde p_k$ is the corresponding noisy estimate for all $k \in \Z$. We write also $\hat p_k^\prime := \hat p_k(x^\prime)$ and $\tilde p_k^\prime$ for the noisy estimate. Let $M := \{k \in \Z : \hat p_k \neq \hat p_k^\prime\}$ be the set of bins affected by the change of the single entry between $x$ and $x^\prime$. As only one pair $(x_j, x_j^\prime)$ differs, we have $|M| \leq 2$. For $k,l \in M$ three cases occur: 
    \begin{enumerate}
        \setlength\itemsep{-0.3em}

        \item[] \textbf{Case 1}: $x_j, x_j^\prime$ are in the same bin $B_k$.

        \item[] \textbf{Case 2}: $x_j, x_j^\prime$ are in different bins $B_k, B_l$ where $\hat p_k = \frac 1 n, \hat p_k^\prime = 0$ and $\hat p_l = 0, \hat p_l^\prime = \frac 1 n$. 

        \item[] \textbf{Case 3}: $x_j, x_j^\prime$ are in different bins $B_k, B_l$ where $\hat p_k, \hat p_k^\prime, \hat p_l, \hat p_l^\prime > 0$. 
    \end{enumerate} 

    For all other $k \in M^c$, $\tilde p_k$ and $\tilde p_k^\prime$ are equal in distribution implying constants $(0,0)$ in Equation \eqref{eq:diffpriv} for this part of the output of $\mathcal A$. It hence suffices to show \DP{} for the remaining $k$ in the three cases. 
    
    For \textbf{Case 1} the output of $\mathcal A$ is unchanged and we have $\P\Parent{\mathcal A(x) \in S} = \P\Parent{\mathcal A(x^\prime) \in S}$ for all measurable sets $S$. This yields Equation \eqref{eq:diffpriv} with constants $(0,0)$. 

    For \textbf{Case 2} when $\hat p_l = 0$, by definition of $\mathcal A$, $\tilde p_l = 0$. With $t = \frac{2}{\varepsilon n} \log(\frac 2 \delta) + \frac 1 n$ we then have
    \begin{align*}
        \P\Parent{\tilde p_l \neq \tilde p_l^\prime}
        = \P\Parent{\tilde p_l^\prime \neq 0}
        &= \P\Parent{\Curly{\tilde p_l^\prime \neq 0} \cap \Curly{\hat p_l^\prime + \xi_k < t}} + \P\Parent{\Curly{\tilde p_l^\prime \neq 0} \cap \Curly{\hat p_l^\prime + \xi_k \geq t}} \\
        &= 0 + \P\Parent{\hat p_l^\prime + \xi_k \geq \frac{2}{\varepsilon n} \log\Parent{\frac 2 \delta} + \frac 1 n} \\
        &= \P\Parent{\xi_k \geq \frac{2}{\varepsilon n} \log\Parent{\frac 2 \delta}} 
        \leq \frac{\delta}{2}. 
    \end{align*}

    By a symmetry argument we also have $\P\Parent{\tilde p_k \neq \tilde p_k^\prime} \leq \frac \delta 2$. A union bound yields $\P(\mathcal A(x) \neq \mathcal A(x^\prime)) = \P(\Curly{\tilde p_l \neq \tilde p_l^\prime} \cup \Curly{\tilde p_k \neq \tilde p_k^\prime}) \leq \delta$. Now, we can show Equation \eqref{eq:diffpriv} with constants $(0,\delta)$ as follows: 
    \begin{align*}
        \P\Parent{\mathcal A(x) \in S}
        &= \P\Parent{\Curly{\mathcal A(x) \in S} \cap \Curly{\mathcal A(x) = \mathcal A(x^\prime)}} + \P\Parent{\Curly{\mathcal A(x) \in S} \cap \Curly{\mathcal A(x) \neq \mathcal A(x^\prime)}} \\
        &\leq \P\Parent{\mathcal A(x^\prime) \in S} + \delta.
    \end{align*}
    
    For $\textbf{Case 3}$ we first note that the sensitivity of $\hat p_k(\cdot)$ is $1/n$. Indeed, for all $k \in \Z$ and $d_H(x, x^\prime) \leq 1$: 
    \begin{align*}
        \Delta 
        = \sup_{x, x^\prime} |\hat p_k(x) - \hat p_k(x^\prime)| 
        = \sup_{x, x^\prime} \left|\frac 1 n \Parent{\indicator{x_j \in B_k} - \indicator{x_j^\prime \in B_k}}\right|
        \leq \frac 1 n. 
    \end{align*}

    The noisy statistic $\check p_k = \hat p_k + \xi_k$ ensures that Equation \eqref{eq:diffpriv} holds for $\check p_k$ with constants $(\varepsilon, 0)$ by the Laplace mechanism, because $\xi_k \sim \operatorname{Lap}(0, b)$ and $b = \frac{2\Delta}{\varepsilon }=\frac{2}{\varepsilon n}$. Therefore, 
    \begin{align*}
        \tilde p_k =\check p_k \cdot \indicator{\check p_k\geq \frac{2}{\varepsilon n}\log(\frac 2 \delta)+\frac{1}{n}}
    \end{align*}
    also fulfills Equation \eqref{eq:diffpriv} with constants $(\varepsilon,0)$ by post-processing.
\end{proof}

\begin{proof}[Proof of Lemma \ref{lem:utilhistlogsob}]\hfill 

    Algorithm \ref{alg:stablehist} is $(\varepsilon, \delta)$-DP by Lemma \ref{lem:privhistlearner}. 

    For utility, we adapt the proof of Lemma 2.3 by \citet{karwa:vadhan2018}. The triangle inequality and the law of total probability yield the upper bound
    \begin{align}
        \nonumber \P\Parent{\max_{k \in \Z} |\tilde p_k - p_k| > \eta}
        &\leq \P\Parent{\max_{k \in \Z} |\tilde p_k - \hat p_k| > \frac \eta 2} + \P\Parent{\max_{k \in \Z} |\hat p_k - p_k| > \frac \eta 2} \\
        &= \E\Bracket{\P\Parent{\max_{k \in \Z} |\tilde p_k - \hat p_k| > \frac \eta 2 \Big|X^n}} + \P\Parent{\max_{k \in \Z} |\hat p_k - p_k| > \frac \eta 2}. \label{eq:proof_stable_hist}
    \end{align}

    By conditioning on $X^n$ in the first term, we isolate the estimation error due to privacy of the stability-based histogram given a specific realization. Hence, by Lemma \ref{lem:privlossstabhist} we have 
    \begin{align*}
        \E\Bracket{\P\Parent{\max_{k \in \Z} \abs{\tilde p_k - \hat p_k} \geq \frac \eta 2 \Big | X^n}}
        \leq n \Parent{1+\frac{e^\varepsilon}{\delta}} \exp\Parent{-\frac{\varepsilon n\eta}{4}}. 
    \end{align*}

    For the estimation error of $\hat p_k$ we then invoke Lemma \ref{lem:estlossemphist} under Assumption \ref{ass:logsobdep}. This yields
    \begin{align*}
        \P\Parent{\max_{k \in \Z} |\hat p_k - p_k| \geq \frac \eta 2}
        \leq \frac{16}{\eta} \exp\Parent{-\frac{2}{27} \frac{n\eta^3}{4^3 \rho M^2}}. 
    \end{align*}

    Plugging both bounds into the inequality \eqref{eq:proof_stable_hist} shows the desired result. 
\end{proof}

\begin{proof}[Proof of Lemma \ref{lem:utilhistiid}]\hfill 

    Analogous to the proof of Lemma \ref{lem:utilhistlogsob}. Uses Lemma \ref{lem:estlossemphist} under the \iid{} assumption on the entries of $X^n \in \R^n$. Thus, it uses the Dvoretzky-Kiefer-Wolfowitz inequality of Corollary 1 in \citet{massart1990} instead of Theorem 1.2 by \citet{bobkov:götze2010}. 
\end{proof}

\paragraph{Auxiliary Results for Stable Histogram}

\begin{theorem}{(Theorem 1.2, \citet{bobkov:götze2010})}
    \label{thm:concavgmargCDF}
    Let $X \sim \nu$ on $\R^n$ where $\nu$ is a probability measure fulfilling $\LSI{\rho}$. Assume that the average marginal cumulative distribution function $\bar F(x) = \frac{1}{n} \sum_{i=1}^n \P(X_i \leq x)$ is $M$-Lipschitz. Then, for any $r > 0$
    \begin{align*}
        \P\Parent{\sup_{x\in\R} |\bar F(x) - \hat F_n(x)| \geq r} 
        \leq \frac 4 r \exp\Parent{-\frac{2}{27}\frac{nr^3}{\rho M^2}}. 
    \end{align*}
\end{theorem}

\begin{lemma}
    \label{lem:estlossemphist}
    Let $\eta > 0$ and $(B_k)_{k\in\Z}$ be disjoint bins covering $[-\mathsf B, \mathsf B] \subseteq \R$. Assume $X \in \R^n$ has \iid{} entries or fulfills Assumption \ref{ass:logsobdep} with constants $(\rho, M)$. Let $\hat p_k := \frac 1 n \sum_{i=1}^n \indicator{X_i \in B_k}$ be the empirical mass of bin $B_k$. Then, when estimating the $p_k = \E[\hat p_k]$ uniformly over $k\in\mathbb{Z}$, the estimation error is
    \begin{align*}
        \P\Parent{\max_{k \in \Z} |\hat p_k - p_k| \geq \eta}
        \leq \begin{cases}
            2 \exp\Parent{-\frac{n\eta^2}{2}} & \text{if $X_1,\dots ,X_n$ are \iid{}}, \\
            \frac{8}{\eta} \exp\Parent{-\frac{2}{27} \frac{n\eta^3}{2^3 \rho M^2}} & \text{if $X$ fulfills Assumption \ref{ass:logsobdep}}.
        \end{cases}
    \end{align*}
\end{lemma}

\begin{proof}

    Let $\hat F_n(x) = \frac 1 n \sum_{i=1}^n \indicator{X_i \leq x}$ be the empirical cdf and $\bar F(x) = \E[\hat F_n(x)]$ the average marginal cdf. As $X^n$ fulfills Assumption \ref{ass:logsobdep}, Theorem \ref{thm:concavgmargCDF} yields: 
    \begin{align*}
        \P\Parent{\sup_{x\in\R} |\bar F(x) - \hat F_n(x)| \geq r} 
        \leq \frac 4 r \exp\Parent{-\frac{2}{27}\frac{nr^3}{\rho M^2}}. 
    \end{align*}

    Let $-\infty = x_0 < x_1 < \dots < x_\infty = \infty$ be the endpoints of the bins $(B_k)_{k \in \Z}$. Then, 
    \begin{align*}
        p_k = \bar F(x_k) - \bar F(x_{k-1}) \quad \text{ and } \quad \hat p_k = \hat F_n(x_k) - \hat F_n(x_{k-1}). 
    \end{align*}

    Using these relations and a triangle inequality, for all $k \in \Z$: 
    \begin{align*}
        |\hat p_k - p_k|
        &= |\bar F(x_k) - \bar F(x_{k-1}) - (\hat F_n(x_k) - \hat F_n(x_{k-1}))| \\
        &\leq |\bar F(x_k) - \hat F_n(x_k)| + |\bar F(x_{k-1}) - \hat F_n(x_{k-1}))|. 
    \end{align*}

    Therefore, Theorem \ref{thm:concavgmargCDF} shows that 
    \begin{align*}
        \P\Parent{\max_{k \in \Z} |\hat p_k - p_k| \geq \eta}
        &\leq \P\Parent{\exists k \in \Z : |\bar F(x_k) - \hat F_n(x_k)| + |\bar F(x_{k-1}) - \hat F_n(x_{k-1}))| \geq \eta} \\
        &\leq \P\Parent{\sup_{x\in\R} |\bar F(x) - \hat F_n(x)| \geq \frac \eta 2} 
        \leq \frac{8}{\eta} \exp\Parent{-\frac{2}{27} \frac{n\eta^3}{2^3 \rho M^2}}. 
    \end{align*}

    The proof for $X^n \in \R^n$ with \iid{} entries is analogous but uses the Dvoretzky-Kiefer-Wolfowitz inequality of Corollary 1 \citet{massart1990} instead of Theorem \ref{thm:concavgmargCDF} by \citet{bobkov:götze2010}. 
\end{proof}

\begin{lemma}
    \label{lem:privlossstabhist}
    Let $\eta, \varepsilon > 0$ and let $ (B_k)_{k \in \Z}$ be a disjoint set of bins covering $\R$. Then, for input vector $x^n \in \R^n$, Algorithm \ref{alg:stablehist} is $(\epsilon, \delta)$-DP and the histogram output $\tilde{\boldsymbol{p}}$ satisfies:
    \begin{align*}
        \P\Parent{\max_{k \in \Z} \abs{\tilde p_k - \hat p_k} \geq \eta}
        \leq n \Parent{1+\frac{e^{\varepsilon/2}}{\delta}} \exp\Parent{-\frac{\varepsilon n\eta}{2}}. 
    \end{align*}
\end{lemma}

\begin{proof}

    Algorithm \ref{alg:stablehist} is $(\varepsilon, \delta)$-DP by Lemma \ref{lem:privhistlearner}.

    For utility, we follow the proof of Lemma 2.3 in \citet{karwa:vadhan2018} to bound the privacy loss of the stability-based histogram. We perform a case distinction on whether or not $\hat p_k = 0$. 
    \begin{enumerate}

        \item[] \textbf{Case 1}: We have $\tilde p_k = 0$ for all $k \in \mathbb{A} := \curly{k\in\mathbb{Z} : \hat p_k = 0}$ by design of Algorithm \ref{alg:stablehist}. This yields the trivial bound $\P\Parent{\max_{k \in \mathbb{A}} | \tilde p_k - \hat p_k| \geq \eta} = 0$. 
    
        \item[] \textbf{Case 2}: For the indices $k \in \mathbb{A}^c$ where $\hat p_k > 0$ we have: 
        \begin{align*}
            \P&\Parent{|\tilde p_k - \hat p_k| \geq \eta} \\
            &= \P\Parent{\{|\tilde p_k - \hat p_k| \geq \eta \} \cap \{\hat p_k + \xi_k \geq t\}} + \P\Parent{\{|\tilde p_k - \hat p_k| \geq \eta \} \cap \{\hat p_k + \xi_k < t\}} \\
            &= \P\Parent{\{|\xi_k| \geq \eta \} \cap \{\hat p_k + \xi_k \geq t\}} + \P\Parent{\{|\hat p_k| \geq \eta \} \cap \{\hat p_k + \xi_k < t\}} \\
            &\leq \P\Parent{|\xi_k| \geq \eta } + \P\Parent{\xi_k < t - \eta} \\
            &= \P\Parent{|\xi_k| \geq \eta } + \P\Parent{\xi_k > \eta - t}. 
        \end{align*}
        
        Above, the first equality holds as we split the event into two disjoint sets. By definition of Algorithm \ref{alg:stablehist} we have that $\tilde p_k = \hat p_k + \xi_k$ and $\tilde p_k = 0$ if $\tilde p_k < t$, which justifies the second equality. The inequality holds by inclusion. The last equality holds as the distribution of $\xi_k$ is symmetric around zero. Using that $\xi_k \sim \operatorname{Lap}(0, \frac{2}{\varepsilon n})$ we see that
        \begin{equation*}
        \P\Parent{|\xi_k| \geq l} 
        = \exp\Parent{-\frac{\varepsilon n l }{2}}, 
        \end{equation*}
        
        Furthermore, choosing $t = \frac{2}{\varepsilon n}\log\Parent{\frac 2 \delta} + \frac 1 n$ we see that
        \begin{align*}
            \P\Parent{\xi_k > l - t}
            &= \frac 1 2 \exp\Parent{-\frac{\varepsilon n (l-t)}{2}} \\
            &= \frac 1 2 \exp\Parent{-\frac{\varepsilon nl}{2}}\exp\Parent{\frac{\varepsilon n}{2} \Parent{\frac{2}{\varepsilon n}\log\Parent{\frac 2 \delta} + \frac 1 n}} \\
            &= \frac{e^{\varepsilon/2}}{\delta}\exp\Parent{-\frac{\varepsilon nl}{2}}
        \end{align*}

        Plugging both into the bound, we have
        \begin{align*}
            \P\Parent{|\tilde p_k - \hat p_k| \geq \eta}
            &\leq \Parent{1+\frac{e^{\varepsilon/2}}{\delta}} \exp\Parent{-\frac{\varepsilon n\eta}{2}}. 
        \end{align*}
    \end{enumerate}

    By a union bound over all $k \in \Z$ and using that only $n$ of the probabilities can be non-zero,
    \begin{align*}
        \P\Parent{\max_{k \in \Z} |\tilde p_k - \hat p_k| > \eta}
        \leq \sum_{k \in \Z} \P\Parent{|\tilde p_k - \hat p_k| > \eta}
        \leq n \Parent{1+\frac{e^{\varepsilon/2}}{\delta}} \exp\Parent{-\frac{\varepsilon n\eta}{2}}. 
    \end{align*}
\end{proof}

\subsection{Proofs for Private Projection Interval}
\label{app:privmidpoint}

\begin{proof}[Proof of Lemma \ref{lem:privmidpointstabhist}]\hfill 

    The privacy of Algorithm \ref{alg:privrangestabhist} follows from that of Algorithm \ref{alg:stablehist}, shown in Lemma \ref{lem:privhistlearner}. 

    For utility, by Lemma \ref{lem:utilhistlogsob} with $\eta = 1/16$ we have that
    \begin{align*}
        \P\Parent{\max_{k \in \Z} |\tilde p_k - p_k| \leq \frac{1}{16}} 
        &\geq 1 - \Parent{1+\frac{e^\varepsilon}{\delta}} n \exp\Parent{-\frac{\varepsilon n}{4\cdot 16}} - \frac{16}{\eta} \exp\Parent{-\frac{1}{864} \frac{n}{16^3\rho M^2}} \\
        &= 1 - O\Parent{\frac{n}{\delta} \cdot e^{-\kappa n \varepsilon} \vee e^{-\kappa^\prime n}}. 
    \end{align*}

    Here, we suppress the constant $\rho M^2$ in the big-O notation, use that $\delta \leq e^\varepsilon$ for all $\varepsilon > 0$. Combining the guarantee above with the $(\tau, \gamma)_\infty$-concentration of $X^n$ allows us to call Lemma \ref{lem:midpoint} to get 
    \begin{align*}
        \P\Parent{\hat m \in \Big[x_0 \pm 2\tau \Big]}
        \geq 1 - O\Parent{\frac{n}{\delta} \cdot e^{-\kappa n \varepsilon} \vee e^{-\kappa^\prime n}}.
    \end{align*}

    The proof under Assumption \ref{ass:iiddata} is analogous but uses Lemma \ref{lem:utilhistiid} instead of Lemma \ref{lem:utilhistlogsob}. 
\end{proof}

\begin{proof}[Proof of Corollary \ref{cor:privmedianiid}]\hfill

    As $X_i$ are sub-Gaussian with variance proxy $\sigma^2$
    \begin{align*}
        \P\Parent{|X_i - \mu| > t} 
        \leq 2\exp\Parent{-\frac{t^2}{2\sigma^2}}
        =: \frac \gamma n. 
    \end{align*}

    Hence, $X^n$ is $(\tau, \gamma)^\mu_\infty$-concentrated with $\tau = \sqrt{2\sigma^2\log(2n/\gamma)}$. Under Assumption \ref{ass:iiddata} by Lemma \ref{lem:privmidpointstabhist}
    \begin{align*}
        \P\Parent{\hat m \in \Big[\mu \pm 2\sqrt{2\sigma^2\log(2n/\gamma)} \Big]}
        \geq 1 - O\Parent{\frac{n}{\delta} \cdot e^{-\kappa n \varepsilon} \vee e^{-\kappa^\prime n}}.
    \end{align*}

    This holds under the condition that $\gamma \in (0,1 \wedge \frac n 4)$. 
\end{proof}

\paragraph{Auxiliary Results for Private Projection Intervals}

\begin{lemma}
    \label{lem:midpoint}
    Let $\gamma \in (0,1 \wedge \frac n 4)$ and let $X^n \in \R^n$ be $(\tau, \gamma)^{x_0}_\infty$-concentrated. Assume there is a histogram estimator $\mathcal A$ that outputs $(\dots,\tilde p_{-2}, \tilde p_{-1},\tilde{p}_0,\tilde p_1, \tilde p_2,\dots )$ \st{} w.p.\ at least $1-\alpha$ we have $\max_{k \in \Z}\abs{\tilde p_k - p_k} \leq 1/16$. Then, with probability at least $1-\alpha$ the midpoint $\hat m$ in Algorithm \ref{alg:privrangestabhist} fulfills 
    \begin{align*}
        \hat m \in \Big[x_0 \pm 2\tau \Big]. 
    \end{align*}
\end{lemma}

\begin{proof}

    By $(\tau, \gamma)_\infty$-concentration, for all $i \in [n]$ with probability at least $1-\gamma/n =: 1-\alpha$ we have $X_i \in I := [x_0 \pm \tau]$. We construct a set $S$ of bin-indices $k \in \Z$ whose bins cover $I$: 
    \begin{align*}
        S := \{k \in \Z : I \cap B_k \neq \emptyset\}.
    \end{align*}

    Note that given the binwidth $2\tau$ we have $|S|\leq 2$. Since bins in $S$ cover $I$, by a union bound
    \begin{align}
      \nonumber \sum_{k \in S} p_k
        &=
         \frac 1 n \sum_{i=1}^n\sum_{k \in S} \P(X_i \in B_k)
        \geq \frac 1 n \sum_{i=1}^n \P(\bigcup_{k \in S} \Curly{X_i \in B_k}) \\
        &= \frac 1 n \sum_{i=1}^n \P(X_i \in \bigcup_{k \in S} B_k)
        \geq \frac 1 n \sum_{i=1}^n \P(X_i \in I)
        = \frac 1 n \sum_{i=1}^n \P(|X_i - x_0| \leq \tau)
        \geq 1 - \alpha.
        \label{eq:lb_S}
    \end{align}

    By the pigeon hole principle, i.e., a lower bound by the uniform distribution over the $|S|$ bins, 
    \begin{equation}
        \exists k^\star \in S\quad \mbox{ such that } \quad p_{k^\star} \geq \frac{1-\alpha}{|S|}\geq \frac{1-\alpha}{2}.
        \label{eq:pigeon_hole}
    \end{equation}

    At the same time for all $k \in S^c$
    \begin{equation}
        p_k 
        = \frac 1 n \sum_{i=1}^n \P(X_i \in B_k)
        \leq \frac 1 n \sum_{i=1}^n \P(X_i \in \R\setminus I) \leq \frac 1 n \sum_{i=1}^n \P(|X_i - x_0| \geq \tau) 
        \leq \alpha.
        \label{eq:ub_pk}
    \end{equation}

    We now translate the bounds on $p_k$ above into bounds on $\tilde p_k$. If there is an $\eta$ such that
    \begin{align*}
        \P\Parent{\forall k \in \Z : \tilde p_k \in \Bracket{p_k \pm \eta}} 
        = \P\Parent{\max_{k \in \Z} \abs{\tilde p_k - p_k} \leq \eta}
        &\geq 1-\alpha,
    \end{align*} 

    then using \eqref{eq:pigeon_hole} and \eqref{eq:ub_pk} we have $\tilde p_{k^\star} \geq p_{k^\star} - \eta \geq \frac{1-\alpha}{2} - \eta$ for $k^*\in S$ and $\tilde p_k \leq p_k + \eta \leq \alpha + \eta$ for all $k \in S^c$. To ensure that any such $k^\star$ is selected in the $\arg \max$, we need $\tilde p_{k^\star} > \tilde p_k$ for all $k \in S^c$. This holds if: 
    \begin{align*}
        \frac{1-\alpha}{2} - \eta > \alpha + \eta
        \quad \Leftrightarrow \quad \alpha < \frac{1-4\eta}{3}. 
    \end{align*}

    Since by assumption $\eta=1/16$, the last condition becomes $\alpha < 1/4$ and it holds as $\alpha = \gamma/n$ and we assume that $\gamma < n/4$. Hence, with probability at least $1-\alpha$ we select $\hat k \in S$. The center of $B_k$ for any $k \in S$ can at most be $\tau$ away from $I$. The distance from $I$'s endpoints to $x_0$ is also at most $\tau$. A triangle inequality recovers $|\hat m - x_0| \leq 2\tau$.
\end{proof}

\section{Proofs of Section \ref{sec:meanest}}

\begin{proof}[Proof of Lemma \ref{lem:winsmean1D}]\hfill 

    The private midpoint estimator of Algorithm \ref{alg:privrangestabhist} is $(\frac \varepsilon 2, \delta)$-DP by Lemma \ref{lem:privmidpointstabhist} as we call it with $\frac \varepsilon 2$ in Algorithm \ref{alg:winsmean1D}. Further, the Laplace mechanism therein is $(\frac{\varepsilon}{2},0)$-DP by Theorem \ref{thm:lapmech}. Therefore, algorithm $\mathcal A$ is $(\varepsilon,\delta)$-DP through basic composition in Theorem \ref{thm:compthms}. 

    For utility, by Lemma \ref{lem:privmidpointstabhist} under Assumption \ref{ass:iiddata} or \ref{ass:logsobdep} we have 
    \begin{align*}
        \P\Parent{\hat m \in [x_0 \pm 2\tau]}
        \geq 1-O({n}/{\delta} \cdot e^{-\kappa n \varepsilon} \vee e^{-\kappa^\prime n}).
    \end{align*}
    
    Hence, by an application of Lemma \ref{lem:winsmean1Dgeneral} with $C=2$, the event $\mathcal E := \curly{\forall i \in [n] : \proj{[\hat m \pm 3\tau]}{X_i} = X_i}$ has probability at least $1 - \gamma - O({n}/{\delta} \cdot e^{-\kappa n \varepsilon} \vee e^{-\kappa^\prime n})$ and 
    \begin{align*}
        \mathcal A(X) \cdot \indicator{\mathcal E} 
        = \Parent{\bar X_n + \xi} \cdot \indicator{\mathcal E}, 
        \quad \text{almost surely}. 
    \end{align*}
\end{proof}

\begin{lemma}
    \label{lem:winsmeanHD}
    Algorithm \ref{alg:winsmeanHD} denoted by $\mathcal A$ is $(\varepsilon,\delta+\varrho)$-DP for $\delta, \varrho, \varepsilon \in (0,1)$. Let $\gamma \in (0, 1 \wedge \frac n 4)$ and $X^n \in \R^{n \times d}$ be $(\tau, \gamma)_\infty^{x_0}$-concentrated. Make Assumption \ref{ass:iiddata} or \ref{ass:logsobdep} \st{} $\rho M ^2 \lesssim 1$. Then, the event $\mathcal E := \curly{\forall i \in [n], j \in [d] : \proj{R_j}{X_{ij}} = X_{ij}}$ has probability at least $1 - d\gamma - O(d^2n/\delta \cdot e^{-\kappa n \varepsilon^\prime})$ and a.s.
    \begin{align*}
        \mathcal A(X) \cdot \indicator{\mathcal E}
        = \Parent{\bar X_n + \Xi} \cdot \indicator{\mathcal E}, \quad \text{where} \quad \Xi \sim \operatorname{Lap}\Parent{0, \frac{12\tau}{n \varepsilon^\prime}I_d}
        \quad \text{and} \quad \varepsilon'=\varepsilon/\sqrt{8d\log(1/\varrho)}.
    \end{align*}
\end{lemma}

\begin{proof}
    Since the $d$ one-dimensional mean estimators fulfill $(\varepsilon, \delta)$-DP, the coordinate-wise multi-dimensional estimator is private by advanced composition in Theorem \ref{thm:compthms}. Thus, we use the privacy parameters $\varepsilon^\prime, \delta^\prime, \varrho$ defined in Algorithm \ref{lem:winsmeanHD}. Note that this is where $\varrho$ occurs.

    $X^n$ is $(\tau,\gamma)_\infty^{x_0}$-concentrated. Applying Algorithm \ref{alg:winsmean1D} per dimension with $(X^n_{\cdot j}, \tau, \varepsilon^\prime, \delta^\prime)$, by Lemma \ref{lem:winsmean1D} $\mathcal E_j := \curly{\forall i \in [n] : \proj{R_j}{X_{ij}} = X_{ij}}$ has probability at least $1 - \gamma - O(n/\delta \cdot e^{- \kappa n \varepsilon^\prime} \vee e^{-\kappa^\prime n})$ and 
    \begin{align*}
        \mathcal A_1(X^n_{\cdot j}) \cdot \indicator{\mathcal E_j} = \Parent{\frac 1 n \sum_{i=1}^n X_{ij} + \xi_j} \cdot \indicator{\mathcal E_j}, \quad \text{almost surely,}
    \end{align*}

    where we denote the output of Algorithm \ref{alg:winsmean1D} by $\mathcal A_1$. By Lemma \ref{lem:winsmeanHDgeneral} with $h=1$ and $b = \frac{12\tau}{n\varepsilon^\prime}$, ${\mathcal E} := \curly{\forall j \in [d], i \in [n] : \proj{R_j}{X_{ij}} = X_{ij}}$ has probability at least $1-d\gamma-O(dn/\delta^\prime \cdot e^{- \kappa n \varepsilon^\prime})$ and a.s.\
    \begin{align*}
        {\mathcal A}(X^n) \cdot \indicator{{\mathcal E}}
        = \Parent{\frac 1 n \sum_{i=1}^n X_{i\cdot}\T + \Xi} \cdot \indicator{{\mathcal E}}, 
        \quad \text{with $\Xi := (\xi_{1},..., \xi_{d})\T$}.
    \end{align*}

    Because the $\xi_j$ are \iid{} Laplace random variables, $\Xi \sim \operatorname{Lap}(0, \frac{12\tau}{n\varepsilon} I_d)$. We now recover the statement using that $\bar X_n = \frac 1 n \sum_{i=1}^n X_{i\cdot}\T$ and $\delta^\prime = \delta/d$.
\end{proof}

\subsubsection*{Auxiliary Results for Mean Estimators}
\label{app:meanest}

The statements of the following Lemma \ref{lem:winsmean1Dgeneral} and Lemma \ref{lem:winsmeanHDgeneral} are general enough to incorporate both the central and local model, with $h = 1$ and $h = n$, respectively. We provide these general versions to highlight that both models of \DP{} can be treated in a unified way. 

\begin{lemma}
    \label{lem:winsmean1Dgeneral}
    Let $X^n \in \R^n$ be $(\tau, \gamma)_\infty$-concentrated around $x_0 \in \R$. Let algorithm $\tilde{\mathcal A}$ output a midpoint $\hat m$ \st{} $\hat m \in [x_0 \pm C\tau^\prime]$ with probability at least $1-\alpha$. Denote Algorithm \ref{alg:winsmean1D} calling $\tilde{\mathcal A}$ in Line 1 of $\mathcal A$. Then, the event $\mathcal E := \curly{\forall i \in [n] : \proj{[\hat m \pm \tau+C\tau^\prime]}{X_i} = X_i}$ has probability at least $1 - \gamma - \alpha$ and 
    \begin{align*}
        \mathcal A(X^n) \cdot \indicator{\mathcal E} 
        = \Parent{\bar X_n + \bar \xi_h} \cdot \indicator{\mathcal E}, 
        \quad \text{almost surely}. 
    \end{align*}
\end{lemma}

\begin{proof}

    To prove the statement, we note that $X_i$ is only projected if it is more than $\tau+C\tau^\prime$ away from the midpoint. The triangle inequality gives
    \begin{align*}
        |X_i - \hat m| 
        = |X_i - x_0 + x_0 - \hat m| 
        \leq |X_i - x_0| + |x_0 - \hat m|.
    \end{align*}

    Note that $(\tau, \gamma)^{x_0}_\infty$-concentration of $X^n$ and a union bound imply that with probability at least $1-\gamma$ we have that $|X_i - x_0|\leq \tau$ for all $i\in [n]$. Furthermore, by our assumption $\hat m \in [x_0 \pm C\tau^\prime]$ with probability at least $1-\alpha$. Therefore, a union bound shows that with probability at least $1 - \gamma - \alpha$,
    \begin{align*}
        |X_i - \hat m| 
                \leq |X_i - x_0| + |x_0 - \hat m|
        \leq \tau+C\tau^\prime, \quad \forall i \in [n].
    \end{align*}

    In other words, $\P(\mathcal E) \geq 1 - \gamma - \alpha$. The almost sure equality on $\mathcal E$ holds since there
    \begin{align*}
      \mathcal A(X^n) = \frac 1 n \sum_{i=1}^n \proj{\hat I}{X_i} + \bar \xi_h = \bar X_n + \bar \xi_h.
    \end{align*}
\end{proof}

\begin{lemma}
    \label{lem:winsmeanHDgeneral}
    Let $\mathcal A_1 : \R^n \to \R$ be an estimator of $x_0 \in \R$ with arguments $(X^n_{\cdot j}, \tau, \varepsilon, \delta)$ for any $j \in [d]$ where $\tau, \varepsilon, \delta > 0$ and $X^n \in \R^{n \times d}$. Assume that for $h \in \N$, on an event $\mathcal E_j$ that has probability at least $1-\alpha$, we almost surely have 
    \begin{align*}
        \mathcal A_1 (X^n_{\cdot j}) \cdot \indicator{\mathcal E_j} 
        = \Parent{\frac 1 n \sum_{i=1}^n X_{ij} + \frac 1 h \sum_{l=1}^h \xi_{lj}} \cdot \indicator{\mathcal E_j}, 
        \quad \text{with \iid{} $\xi_{lj} \sim \operatorname{Lap}(0,b)$}. 
    \end{align*}

    Let ${\mathcal A}$ be Algorithm \ref{alg:winsmeanHD} calling $\mathcal A_1$ in Line 3. Then, ${\mathcal E} = \bigcap_{j=1}^d \mathcal E_j$ has probability at least $1-d\alpha$ and 
    \begin{align*}
        {\mathcal A}(X^n) \cdot \indicator{{\mathcal E}}
        = \Parent{\frac 1 n \sum_{i=1}^n X_{i\cdot}\T + \frac 1 h \sum_{l=1}^h \Xi_l} \cdot \indicator{{\mathcal E}}, 
        \quad \text{with} \quad \Xi_l = (\xi_{l1},...\xi_{ld})\T.
    \end{align*}
\end{lemma}

\begin{proof}

    By assumption the one-dimensional mean estimator applied to $X^n_{\cdot j}$ is a noised version of the empirical mean on $\mathcal E_j$. We apply the mean estimator $\mathcal A_1$ per dimension with arguments $(X^n_{\cdot j}, \tau, \varepsilon, \delta)$ to form the multivariate estimator ${\mathcal A}$. We then combine the guarantees per coordinate through a union bound over $j \in [d]$. Concretely, on the event $\mathcal E_j$ with $\P(\mathcal E_j) \geq 1-\alpha$: 
    \begin{align*}
        \mathcal A(X^n_{\cdot j}) = \frac 1 n \sum_{i = 1}^n X_{ij} + \frac 1 h \sum_{l=1}^h \xi_{lj}.
    \end{align*}

    Letting ${\mathcal E} = \bigcap_{j=1}^n \mathcal E_j$ we see that an union bound over $\mathcal E_1^c,\dots ,\mathcal E_d^c$ gives $\P({\mathcal E}) \geq 1-d\alpha$. Due to the coordinate-wise construction of ${\mathcal A}$ this recovers the statement. 
\end{proof}

\begin{remark}
   Lemma \ref{lem:winsmeanHDgeneral} is essentially analogous to \citet[Theorem 2]{levy:sun:amin:kale:kulesza:mohri:suresh2021} by Lemma \ref{lem:betaclose} which implies that $\mathcal A(X^n) \sim_\beta \bar X_n + \Xi$, where $X\sim_\beta Y$ means that the total variation distance between the distributions of $X$ and $Y$ is smaller or equal to $\beta$.
\end{remark}

\begin{lemma}
    \label{lem:betaclose}
    Let $X,Y$ be two random variables on $(\Omega, \Sigma, \mu)$ and $(\Omega, \Sigma, \nu)$, respectively. Let there be an event $E$ on which $X = Y$ and $\P(E) \geq 1-\beta$. Then, $X \sim_\beta Y$, i.e., $X$ and $Y$ are $\beta$-close.
\end{lemma}

\begin{proof}
    To show that $X \sim_\beta Y$ we need to show the inequality below: 
    \begin{align*}
        d_{TV}(\mu,\nu)
        = \sup_{A \in \Sigma} |\mu(A) - \nu(A)|
        = \sup_{A \in \Sigma} |\P(X \in A) - \P(Y \in A)| 
        \leq \beta. 
    \end{align*}

    We know that $X(\omega) = Y(\omega)$ for all $\omega \in E \subset \Omega$. Hence, for all $A \in \Sigma$ we have $\P(\curly{X \in A} \cap E) = \P(\curly{Y \in A} \cap E)$. To use this equality, we decompose the probabilities in the TV-distance as follows: 
    \begin{align*}
        \P(X \in A) &= \P(\curly{X \in A} \cap E) + \P(\curly{X \in A} \cap E^c), \\
        \P(Y \in A) &= \P(\curly{Y \in A} \cap E) + \P(\curly{Y \in A} \cap E^c). 
    \end{align*}

    Plugging the decompositions into $d_{TV}(\mu, \nu)$ and we get
    \begin{align*}
        d_{TV}(\mu,\nu)
        &= \sup_{A \in \Sigma} |\P(\curly{X\in A} \cap E) + \P(\curly{X\in A} \cap E^c) - \P(\curly{Y\in A} \cap E) - \P(\curly{Y\in A} \cap E^c)| \\
        &= \sup_{A \in \Sigma} |\P(\curly{X\in A} \cap E^c) - \P(\curly{Y\in A} \cap E^c)| \\
        &\leq \sup_{A \in \Sigma} \max \Curly{\P(\curly{X\in A} \cap E^c), \P(\curly{Y\in A} \cap E^c)} \\
        &\leq \P(E^c)
        \leq \beta. 
    \end{align*}
\end{proof}

\subsection{Proofs of Theoretical Guarantees}

\subsubsection{Proof of Finite Sample Bound}
\label{app:finitesample}

\begin{proof}[Proof of Theorem \ref{thm:rMSEwinsmeanHD}]\hfill 

   The privacy guarantee follows from Lemma \ref{lem:winsmeanHD}. This proof directly follows from Lemma \ref{lem:concwinsmeanhd}. To apply it, we use that by Lemma \ref{lem:winsmeanHD} there is an event $\mathcal E$ that has probability at least $1 - d\gamma - O(d^2n/\delta \cdot e^{-\kappa n \varepsilon^\prime} \vee e^{-\kappa^\prime n})$ \st{} almost surely, 
    \begin{align*}
        \mathcal A(X^n) \cdot \indicator{\mathcal E}
        = \Parent{\bar X_n + \Xi} \cdot \indicator{\mathcal E}, \quad \text{where $\Xi \sim \operatorname{Lap}\Parent{0, \frac{12\tau}{n\varepsilon^\prime}I_d}$}.
    \end{align*}

    Applying Lemma \ref{lem:concwinsmeanhd} with $h=1$, with probability at least $1 - \alpha - d\gamma - O(d^2n/\delta \cdot e^{-\kappa n \varepsilon^\prime} \vee e^{-\kappa^\prime n})$, 
    \begin{align*}
        \Twonorm{\mathcal A(X^n) - \mu}
        &\leq \Twonorm{\bar X_n - x_0} + \frac{12\tau}{n\varepsilon^\prime} \Parent{\sqrt{2d} + \sqrt{4\log(3/\alpha)^2}} \\
        &=\Twonorm{\bar X_n - x_0} + 12\tau \Parent{\sqrt{\frac{16d^2\log(1/\varrho)}{n^2\varepsilon^2}} + \sqrt{\frac{32d\log(1/\varrho)\log(3/\alpha)^2}{n^2\varepsilon^2}}}\\
        &\lesssim \Twonorm{\bar X_n - x_0} + \tau \sqrt{\frac{d^2\log(1/\varrho)\log(3/\alpha)^2}{n^2\varepsilon^2}}.
    \end{align*}

    The statement is obtained using that $e^{-\kappa^\prime n} \leq e^{-\kappa^{\prime} n \varepsilon^\prime}$ as $\varepsilon < 1$ and setting $\alpha = d\gamma$. 
\end{proof}

Lemma \ref{lem:concwinsmeanhd} below is general enough to handle both the central and local model of \DP{} with $h = 1$ and $h = n$, respectively. 

\begin{lemma}
    \label{lem:concwinsmeanhd}
    Let $\mathcal A : \R^{n\times d} \to \R^d$ be an estimator of $x_0 \in \R^d$, $X^n := [X_1\T,\dots,X_n\T]\T \in \R^{n\times d}$ and \iid{} $\Xi_j \sim \operatorname{Lap}(0, b I_d)$ for all $j \in [h]$. Assume that on an event $\mathcal E$ having probability at least $1-\upsilon$, 
    \begin{align*}
        \mathcal A(X^n) \cdot \indicator{\mathcal E} = (\bar X_n + \bar \Xi_h) \cdot \indicator{\mathcal E}, 
        \quad \text{almost surely}. 
    \end{align*}
    
    Then, with probability at least $1-\alpha-\upsilon$
    \begin{align*}
        \Twonorm{\mathcal A(X^n) - x_0} 
        \leq \Twonorm{\bar X_n - x_0} + b \Parent{\sqrt{\frac{2d}{h}} + \sqrt{\frac{4\log(3/\alpha)^2}{h}}}. 
    \end{align*}
\end{lemma}

\begin{proof}

    With $t = b(\sqrt{2d/h} + \sqrt{4\log(1/\alpha)^2/h})$, define the events
    \begin{align*}
        A
        := \curly{\twonorm{\mathcal A(X^n) - x_0} \leq \twonorm{\bar X_n - x_0} + t})
        \quad \text{and} \quad 
        L 
        := \curly{\twonorm{\bar \Xi_h} \leq t}. 
    \end{align*}
    
    Then, on $\mathcal E \cap L$, by a triangle inequality
    \begin{align*}
        \twonorm{\mathcal A(X^n) - x_0}
        = \twonorm{\bar X_n + \bar \Xi_h - x_0}
        \leq \twonorm{\bar X_n - x_0} + \twonorm{\bar \Xi_h}
        \leq \twonorm{\bar X_n - x_0} + t. 
    \end{align*}

    Thus, $\mathcal E \cap L \subseteq A$ and by a union bound and Lemma \ref{lem:normlapvectors} it holds that
    \begin{align*}
        \P(A) 
        \geq \P(\mathcal E \cap L)
        = 1-\P(\mathcal E^c \cup L^c)
        \geq 1 - \P(\mathcal E^c) - \P(L^c)
        \geq 1- \upsilon - \alpha. 
    \end{align*}
\end{proof}

\subsubsection{Proof of In-Expectation Bound}
\label{app:asymptotics}

As $\mathcal A_{\hat m(Z^n)}$ is constructed per coordinate, the proof reduces to the one-dimensional case of Lemma \ref{lem:consistwinsmean1D} where $X^n,Z^n \in \R^n$. Therein, by iterated expectation and a bias-variance decomposition conditional on $\hat I$ that collects $Z^n$ and the randomness of Algorithm \ref{alg:privrangestabhist}:
\begin{align*}
    \E[\abs{\mathcal A_{\hat m(Z^n)}(X^n) - \mu}^2] 
    \leq \E[\var{A_{\hat m(Z^n)}(X^n)| \hat I} + \operatorname{Bias}(A_{\hat m(Z^n)}(X^n)| \hat I)^2]. 
\end{align*}

The conditional variance is bounded easily, because $\hat I$, $X^n$ and $\Xi$ are independent and the projection is a contraction. Our main technical innovation is a bound on the conditional bias. This is straightforward for $X_i \in [-B,B]$, since then similar to Appendix B.1 in \citet{duchi:jordan:wainwright2018},
\begin{align*}
    \E[\operatorname{Bias}(A_{\hat m(Z^n)}(X^n)|\hat I)^2]
    &\lesssim \E[\E\bracket{\abs{X_{i} - \hat m}^2| \hat I} \cdot \P\parent{X_{i} \notin \hat I| \hat I}]
    \leq B^2 \cdot \P\parent{X_{i} \notin \hat I}.
\end{align*}

Our contribution is to avoid the boundedness assumption in the argument above. In particular, this involves handling the failure event of the private midpoint estimator in Algorithm \ref{alg:privrangestabhist} caused by the stable histogram outputting $\boldsymbol{\tilde p} = \boldsymbol{0}$. On this event with exponentially small probability in $n\varepsilon$ as shown in Lemma \ref{lem:histfailure} in Appendix \ref{app:asymptotics}, the MSE is bounded by $\twonorm{\mu}^2$. When the private midpoint algorithm does not fail, we exploit the design of the stable histogram estimator in Algorithm \ref{alg:stablehist} to bound the bias. Concretely, we use that the midpoint $\hat m(Z^n)$ can at most be of order $\tau$ away from one of the $X_i$ and can thus be controlled using maximal inequalities. 

\begin{proof}[Proof of Theorem \ref{thm:inexpub}]\hfill 

    The estimator $\mathcal A_{\hat m(Z^n)}$ used in this theorem differs from the one in Algorithm \ref{alg:winsmeanHD} only by computing the midpoint $\hat m := \hat m(Z^n) \in \R^d$ using an additional independent dataset $Z^n$ instead of $X^n$.

    The estimator $\mathcal A_{\hat m(Z^n)}$ is private by reasoning analogous to Theorem \ref{lem:winsmeanHD}.

    By definition of the MSE of an estimator, we have the first line in the following: 
    \begin{align*}
        \operatorname{MSE}(\mathcal A_{\hat m(Z^n)})
        &:= \E[\twonorm{\mathcal A_{\hat m(Z^n)}(X^n) - \mu}^2]
        = \sum_{j=1}^d \E[\abs{\mathcal A_{1, \hat m(Z^n_{\cdot j})}(X^n_{\cdot j}) - \mu_j}^2] \\
        &\lesssim \sum_{j=1}^d \left(\rho \log(4n) \Parent{\frac{1}{n} + \frac{1}{n^2(\varepsilon^\prime)^2} + \frac{n}{\delta^\prime e^{c n \varepsilon^\prime/(2\cdot 4^3)}}} + \frac{\mu_j^2}{\delta^\prime e^{c n \varepsilon^\prime}} \right)\\
        &\lesssim d\rho \log(4n) \Parent{\frac{1}{n} + \frac{d\log(1/\varrho)}{n^2\varepsilon^2} + \frac{dn}{\delta e^{c n \varepsilon^\prime/(2\cdot 4^3)}}} + \frac{d\twonorm{\mu}^2}{\delta e^{c n \varepsilon^\prime}} \\
        &\lesssim d\rho \log(4n) \Parent{\frac{1}{n} + \frac{d\log(1/\varrho)}{n^2\varepsilon^2} + \frac{dn}{\delta e^{c n \varepsilon^\prime/(2\cdot 4^3)}}} + \frac{d\twonorm{\mu}^2}{\delta e^{c n \varepsilon^\prime/4^3}}. 
    \end{align*}

    The first inequality holds by Lemma \ref{lem:consistwinsmean1D} that is applicable as $A_{\hat m(Z^n)}$ is constructed from coordinate-wise estimators $\mathcal A_{1, \hat m(Z^n_{\cdot j})}$ specified in Algorithm \ref{alg:winsmean1D} with the midpoints $\hat m_j$ computed on $Z^n_{\cdot j}$ and the Winsorized means computed on $X^n_{\cdot j}$. The privacy parameters 
    \begin{align*}
        \varepsilon^\prime = \varepsilon / \sqrt{8d \log(1/\varrho)} < 1
        \quad \text{and} \quad 
        \delta^\prime = \delta / d
    \end{align*}
    are adjusted according to advanced composition in Theorem \ref{thm:compthms}, which yields the second to last inequality. Finally, the last inequality holds, because $c \geq c/4^3$. In the statement $\kappa := c/4^3$. 
\end{proof}

\begin{corollary}
    \label{cor:inexpminimax}
    Let $\mathcal A_{\hat m(Z^n)}$ denote Algorithm \ref{alg:winsmeanHD} with midpoint $\hat m(Z^n)$. $\mathcal A_{\hat m(Z^n)}$ is $(\varepsilon, \delta)$-DP for $\varepsilon, \delta \in (0,1)$. Let $X^n,Z^n \in \R^{n \times d}$ have rows with mean $\mu \in \R^d$ \st{} $\infnorm{\mu}^2 \lesssim \rho n e^{\kappa n \varepsilon^\prime/2}$. Make Assumption \ref{ass:iiddata} with $\rho$-sub-Gaussian entries or Assumption \ref{ass:logsobdep} \st{} $\rho M^2 \lesssim 1$. For $\delta \in (n\varepsilon^\prime/e^{\kappa n \varepsilon^\prime}, 1/n]$,
    \begin{align*}
        \E[\twonorm{\mathcal A_{\hat m(Z^n)}(X^n) - \mu}^2]
        &\lesssim d\rho \log(4n) \Parent{\frac{1}{n} + \frac{d\log(2/\delta)}{n^2\varepsilon^2}}.
    \end{align*}

\end{corollary}

\begin{proof}

    By Theorem \ref{thm:inexpub} with $\varrho = \delta$ and using that $\twonorm{\mu}^2 \leq d \infnorm{\mu}^2 \lesssim d\rho ne^{\kappa n \varepsilon^\prime/2}$ we have 
    \begin{align*}
        \E[\twonorm{\mathcal A_{\hat m(Z^n)}(X^n) - \mu}^2]
        &\lesssim d\rho \log(4n) \Parent{\frac{1}{n} + \frac{d\log(1/\varrho)}{n^2\varepsilon^2} + \frac{dn}{\delta e^{\kappa n \varepsilon^\prime/2}}} + \frac{d\twonorm{\mu}^2}{\delta e^{\kappa n \varepsilon^\prime}} \\
        &\lesssim d\rho \log(4n) \Parent{\frac{1}{n} + \frac{d\log(1/\varrho)}{n^2\varepsilon^2} + \frac{dn}{\delta e^{\kappa n \varepsilon^\prime/2}}} + \frac{d^2 \rho n}{\delta e^{\kappa n \varepsilon^\prime/2}} \\
        &\lesssim d\rho \log(4n) \Parent{\frac{1}{n} + \frac{d\log(1/\varrho)}{n^2\varepsilon^2} + \frac{d}{\delta^2 e^{\kappa n \varepsilon^\prime/2}}}.
    \end{align*}

    Here, for the last inequality we use that $\delta \leq 1/n$ and $1 \leq \log(4n)$. We now choose $\delta$ \st{} the second dominates the third term. Note that $\varepsilon^\prime \asymp \varepsilon / \sqrt{d\log(1/\varrho)}$ and thus we require
    \begin{align*}
        \frac{d}{\delta^2 e^{\kappa n \varepsilon^\prime/2}}
        \lesssim \frac{d}{(n\varepsilon^\prime)^2}
        \quad \Leftrightarrow \quad
        \delta
        \gtrsim \frac{n\varepsilon^\prime}{e^{\kappa n \varepsilon^\prime/4}}.
    \end{align*}

    The statement then follows by setting $\varrho = \delta$ and $\delta$ is replaced by $\delta / 2$ to obtain $(\varepsilon, \delta)$-\DP{}. 
\end{proof}

\begin{lemma}
    \label{lem:consistwinsmean1D}
    Let $\mathcal A_{\hat m(Z^n)}$ denote Algorithm \ref{alg:winsmean1D} with midpoint $\hat m$ computed on $Z^n \in \R^n$. $\mathcal A_{\hat m(Z)}$ is $(\varepsilon, \delta)$-DP for $\varepsilon, \delta \in (0,1)$. Let $X^n \in \R^n$ fulfill Assumption \ref{ass:iiddata} and have $\rho$-sub-Gaussian entries or Assumption \ref{ass:logsobdep} \st{} $\rho M^2 \lesssim 1$. If $Z^n$ is an independent copy of $X^n$ and choosing $\tau^2 \asymp \rho\log(4n)$,
    \begin{align*}
        \E[\abs{\mathcal A_{\hat m(Z^n)}(X^n) - \mu}^2]
        \lesssim \rho \log(4n) \Parent{\frac{1}{n} + \frac{1}{n^2\varepsilon^2} + \frac{n}{\delta e^{c n \varepsilon/(2\cdot 4^3)}}} + \frac{\mu^2}{\delta e^{c n \varepsilon}}.
    \end{align*}
\end{lemma}

\begin{proof}
    Algorithm \ref{alg:winsmeanHD} is private by reasoning analogous to Lemma \ref{lem:winsmean1D}. 

    Let the the output of Algorithm \ref{alg:privrangestabhist} on $Z^n$ be the projection interval $\hat I := \hat I(Z^n)$. For brevity, define the estimator $T$ and its noise-free version $T_0$ as
    \begin{align*}
        T 
        := \mathcal A_{\hat m(Z^n)}(X^n)
        = \frac 1 n \sum_{i=1}^n \proj{\hat I}{X_i} + \xi
        \quad \text{and} \quad 
        T_0 
        := \frac 1 n \sum_{i=1}^n \proj{\hat I}{X_i}. 
    \end{align*}
    This construction of $T$ ensures that $X^n$ and $\hat I$ are independent and simplifies the analysis. By definition of $\operatorname{MSE}(T)$, iterated expectation and a bias-variance decomposition applied to the conditional mean squared error $\E[\abs{T - \mu}^2| \hat I]$ we obtain 
    \begin{align*}
        \operatorname{MSE}(T)
        &:= \E[\abs{T - \mu}^2]
        = \E[\E[\abs{T - \mu}^2| \hat I]]
        = \E[\E[\abs{T - \E[T| \hat I]}^2| \hat I] + \abs{\E[T| \hat I] - \mu}^2] \\
        &= \E[\var{T| \hat I} + \operatorname{Bias}(T| \hat I)^2]. 
    \end{align*}
    We now bound the conditional variance and bias of $T$ separately. For the conditional variance, by the mutual independence of $\xi, X^n$ and $\hat I$ and the variance of a Laplace random variable
    \begin{align*}
        \var{T| \hat I}
        &= \var{T_0 + \xi| \hat I}
        = \var{T_0 | \hat I} + \var{\xi} \\
        &= \var{T_0 | \hat I} + 2b^2
        \leq \frac{\rho}{n} + 2b^2
        \lesssim \frac{\rho}{n} + \frac{\tau^2}{n^2\varepsilon^2}.
    \end{align*}

    Above, the inequality holds by Corollary \ref{cor:varL2projiid} or \ref{cor:varL2projlogsob} which are both applicable, because we condition on $\hat I$ that is independent of $X^n$. Further, we use that $b \asymp \tau/(n\varepsilon)$.
    
    It remains to upper bound the conditional bias of $T$. For the following we use that $\E[\xi] = 0$, $\xi, \hat I$ are independent and Jensen's inequality: 
    \begin{align*}
        \operatorname{Bias}(T| \hat I)^2
        &:= \abs{\E[T| \hat I] - \mu}^2
        =  \abs{\E[T_0| \hat I] + \E[\Xi] - \mu}^2 \\
        &= \abs{\E[T_0 - \bar X_n| \hat I]}^2
        = \Abs{\E\Bracket{\frac 1 n \sum_{i=1}^n( \proj{\hat I}{X_i} - X_i) | \hat I}}^2
        \leq \frac 1 n \sum_{i=1}^n \E\bracket{\abs{\proj{\hat I}{X_i} - X_i} | \hat I}^2. 
    \end{align*}

    Note that the distance between $X_i$ and $\proj{\hat I}{X_i}$ is upper bounded as
    \begin{align*}
        |\proj{\hat I}{X_{i}} - X_{i}|
        \leq \begin{cases}
                0 & \text{if $X_{i} \in \hat I$} \\
                X_{i} - (\hat m + 3\tau) & \text{if $X_{i} \notin \hat I$ and $X_{i} > \hat m$} \\
                \hat m - 3\tau - X_{i} & \text{if $X_{i} \notin \hat I$ and $X_{i} < \hat m$}. 
            \end{cases}
    \end{align*}
    
    These three cases can be combined into the following bound
    \begin{align*}
        \abs{\proj{\hat I}{X_{i}} - X_{i}}
        &\leq (\abs{X_{i} - \hat m} - 3\tau) \cdot \indicator{X_{i} \notin \hat I}
        \leq \abs{X_{i} - \hat m} \cdot \indicator{X_{i} \notin \hat I}. 
    \end{align*}

    We introduce the event $\mathcal E_{zero} := \curly{\forall k \in \Z : \tilde p_k = 0}$ on which the histogram fails. On $\mathcal E_{zero}$ we have that $\hat I := [\pm 3\tau]$ and otherwise $\hat I:= [\hat m \pm 3\tau]$. Distinguishing both cases, for each index $i \in [n]$, 
    \begin{align*}
    \E\bracket{\abs{\proj{\hat I}{X_{i}} - X_{i}}| \hat I}^2
        &\leq \E\bracket{\abs{X_{i} - \hat m} \cdot \indicator{X_{i} \notin \hat I}| \hat I}^2 \\
        &= \E\bracket{\abs{X_i} \cdot \indicator{X_i \notin \hat I} \cdot \indicator{\mathcal E_{zero}} + \abs{X_{i} - \hat m} \cdot \indicator{X_{i} \notin \hat I} \cdot \indicator{\mathcal E_{zero}^c}| \hat I}^2 \\
        &\leq \E\bracket{\abs{X_i} \cdot \indicator{\mathcal E_{zero}} + \abs{X_{i} - \hat m} \cdot \indicator{\mathcal E_{zero}^c} \cdot \indicator{X_{i} \notin \hat I}| \hat I}^2 \\
        &\lesssim \underbrace{\E\bracket{\abs{X_i} \cdot \indicator{\mathcal E_{zero}}| \hat I}^2}_{T_1} + \underbrace{\E\bracket{\abs{X_{i} - \hat m} \cdot \indicator{\mathcal E_{zero}^c} \cdot \indicator{X_{i} \notin \hat I}| \hat I}^2}_{T_2}. 
    \end{align*}

    Using the Cauchy-Schwarz inequality, the independence of $X_i, \hat I$ and $\var{X_i} \leq \rho$ which holds for $\rho$-sub-Gaussian random variables \citep[Exercise 2.40]{vershynin2025}, we obtain the bound
    \begin{align*}
        T_1
        = \E\bracket{\abs{X_i} \cdot \indicator{\mathcal E_{zero}}| \hat I}^2
        \leq \E[\abs{X_i}^2] \cdot \E[\indicator{\mathcal E_{zero}}^2| \hat I]
        \leq (\rho + \mu^2) \cdot \P\parent{\mathcal E_{zero}| \hat I}. 
    \end{align*}

    Taking expectation, using Lemma \ref{lem:histfailure} and that $\delta, \varepsilon < 1$ with $c > 0$ that remains the same throughout the whole proof, we have 
    \begin{align*}
        \E[T_1]
        \leq (\rho + \mu^2) \cdot \E[\P\parent{\mathcal E_{zero}| \hat I}]
        = (\rho + \mu^2) \cdot \P\Parent{\mathcal E_{zero}}
        \lesssim (\rho + \mu^2) \cdot 1/\delta \cdot e^{-c n\varepsilon}.
    \end{align*}

    Similarly, for $T_2$ the Cauchy-Schwarz inequality yields
    \begin{align*}
        T_2
        = \E\bracket{\abs{X_{i} - \hat m} \cdot \indicator{\mathcal E_{zero}^c} \cdot \indicator{X_{i} \notin \hat I}| \hat I}^2
        \leq \underbrace{\E\bracket{\parent{\abs{X_{i} - \hat m} \cdot \indicator{\mathcal E_{zero}^c}}^2| \hat I}}_{T_{3}} \cdot \underbrace{\E\bracket{\indicator{X_{i} \notin \hat I}^2| \hat I}}_{T_{4}}.
    \end{align*}

    We bound $T_{3}$ via a triangle inequality. By independence of $X_{i}, \hat I$ and because $\rho \lesssim \tau^2$ we have $\var{X_{i}| \hat I} = \var{X_{i}} \leq \rho \leq 2\tau^2$. Therefore,
    \begin{align*}
        T_3 
        := \E\bracket{\abs{X_{i} - \hat m}^2 \cdot \indicator{\mathcal E_{zero}^c}| \hat I}
        &\leq \E\bracket{\parent{\abs{X_{i} - \mu} + \abs{\mu - \hat m}}^2| \hat I} \cdot \indicator{\mathcal E_{zero}^c}\\
        &\lesssim (\E\bracket{\abs{X_i - \mu}^2| \hat I} + \E\bracket{\abs{\mu - \hat m}^2| \hat I}) \cdot \indicator{\mathcal E_{zero}^c} \\
        &\leq \var{X_i| \hat I} + \E\bracket{\abs{\mu - \hat m}^2| \hat I} \cdot \indicator{\mathcal E_{zero}^c} \\
        &\leq \rho + \abs{\mu - \hat m}^2 \cdot \indicator{\mathcal E_{zero}^c} \\
        &\lesssim \tau^2 + \max_{i \in [n]} \abs{\mu - Z_{i}}^2.
    \end{align*}

    Above, the last inequality holds for the following reason. By design of Algorithm \ref{alg:privrangestabhist} and Algorithm \ref{alg:stablehist} that compute $\hat m$, on $\mathcal E_{zero}^c$ any bin $B_k$ with $\tilde p_k > 0$ must have at least one $Z_{i}$ inside. Bins have width $2\tau$. Hence, the midpoint $m_k$ must be less than $\tau$ away from any $Z_{i} \in B_k$. The same holds for $\hat m$, the center of $B_k$ with the biggest $\tilde p_k$. This reasoning deterministically yields the last inequality: 
    \begin{align*}
        \min_{i\in[n]} Z_{i} - \tau
        \leq \hat m
        \leq \max_{i\in[n]} Z_{i} + \tau, 
        \quad &\Rightarrow \quad
        \abs{\mu - \hat m}
        \leq \max_{i \in [n]} \abs{\mu - Z_{i}} + \tau, \\
        \quad &\Rightarrow \quad
        \abs{\mu - \hat m}^2
        \leq 2\max_{i \in [n]} \abs{\mu - Z_{i}}^2 + 2\tau^2.
    \end{align*}

    To bound $T_4$ we again use that $X_i$ and $\hat I$ are independent. Observe that under Assumption \ref{ass:logsobdep}, by Theorem \ref{thm:logsoblipconc} applied with $f(x) := x_i$ that is $1$-Lipschitz, the $X_i$ are sub-Gaussian. Hence, we have $\P(\abs{X_i - \mu} \geq \tau) \leq \gamma/n$ for $\tau^2 \asymp \rho \log(2n/\gamma)$. Then, by a triangle inequality
    \begin{align*}
        T_4
        = \E\bracket{\indicator{X_i \notin \hat I}| \hat I}
        &= \P\Parent{\abs{X_i - \hat m} > 3\tau| \hat I} \\
        &\leq \P\Parent{|X_{i} - \mu| \geq \tau| \hat I} + \P\Parent{|\mu - \hat m| > 2\tau| \hat I} \\
        &\leq \P\Parent{|X_{i} - \mu| \geq \tau} + \indicator{\hat m \notin [\mu \pm 2\tau]} \\
        &\leq \gamma / n + \indicator{\hat m \notin [\mu \pm 2\tau]}.
    \end{align*}

    For convenience, define $\mathcal E_{bad} := \curly{\hat m \notin [\mu\pm 2\tau]}$ and $M_n^p := \max_{i\in[n]}\abs{Z_i - \mu}^p$ for $p\geq 1$. As the maximum is monotone, $(M_n^1)^p = M_n^p$ for such $p$. To control $\E[\operatorname{Bias}(T| \hat I)^2]$ we now bound $T_3T_4$ in expectation using the Cauchy-Schwarz inequality and the fact that $\indicator{\cdot} = \indicator{\cdot}^2$: 
    \begin{align*}
        \E[T_3T_4]
        &\lesssim \E\Bracket{\parent{\tau^2 + M_n^2}\cdot \Parent{\gamma/n + \indicator{\mathcal E_{bad}}}} \\
        &= \gamma/n \cdot \Parent{\tau^2 + \E\bracket{M_n^2}} + \tau^2 \cdot \P\Parent{\mathcal E_{bad}} + \E\bracket{M_n^2 \cdot \indicator{\mathcal E_{bad}}} \\
        &\leq \tau^2 \cdot (\gamma/n + \P\Parent{\mathcal E_{bad}}) + \gamma/n \cdot \E\bracket{M_n^2} + \sqrt{\E\bracket{M_n^4} \cdot \P\parent{\mathcal E_{bad}}}. 
    \end{align*}

    Again, using that the $X_i$ are sub-Gaussian, by Corollary \ref{cor:subgaussmaxineqs}, Lemma \ref{lem:privmidpointstabhist} and since $\varepsilon < 1$ we have
    \begin{align*}
        \E[M_n^2] \lesssim \rho \log(2n), 
        \quad 
        \E[M_n^4] \lesssim \rho^2 \log(2n)^2
        \quad \text{and} \quad 
        \P(\mathcal E_{bad}) 
        \lesssim \frac{n}{\delta e^{c n \varepsilon/4^3}}. 
    \end{align*}

    Plugging the above bounds into $\E[T_3T_4]$ and using that $1 \leq n/\delta$ and $e^{-\kappa n \varepsilon}\leq e^{-\kappa n \varepsilon/2}$ yields
    \begin{align*}
        \E[T_3T_4]
        &\lesssim \tau^2 \cdot \Parent{\frac \gamma n + \frac{n}{\delta e^{c n \varepsilon/4^3}}} + \frac \gamma n \cdot \rho \log(2n) + \rho \log(2n) \cdot \sqrt{\frac{n}{\delta e^{c n \varepsilon/4^3}}} \\
        &\lesssim \tau^2 \cdot \Parent{\frac \gamma n + \frac{n}{\delta e^{c n \varepsilon/(2\cdot 4^3)}}}.
    \end{align*}

    Collecting terms we have thus obtained the conditional bias bound to follow: 
    \begin{align*}
        \E[\operatorname{Bias}(T| \hat I)^2]
        &\leq \frac 1 n \sum_{i=1}^n \E[\E\bracket{\abs{\proj{\hat I}{X_i} - X_i} | \hat I}^2]
        \leq \frac 1 n \sum_{i=1}^n \E[T_1] + \E[T_3 T_4] \\
        &\lesssim \frac{\mu^2}{\delta e^{c n \varepsilon}} + \tau^2 \Parent{\frac{\gamma}{n} + \frac{n}{\delta e^{c n \varepsilon/(2\cdot 4^3)}}}. 
    \end{align*}

    The bounds on the conditional variance and bias then imply that 
    \begin{align*}
        \operatorname{MSE}&(T)
        = \E[\var{T| \hat I} + \operatorname{Bias}(T| \hat I)^2] \\
        &\lesssim \Parent{\frac{\rho}{n} + \frac{\tau^2}{n^2\varepsilon^2}} + \tau^2 \Parent{\frac{\gamma}{n} + \frac{n}{\delta e^{c n \varepsilon/(2\cdot 4^3)}}} + \frac{\mu^2}{\delta e^{c n \varepsilon}} \\
        &\lesssim \tau^2 \Parent{\frac{\gamma+1}{n} + \frac{1}{n^2\varepsilon^2} + \frac{n}{\delta e^{c n \varepsilon/(2\cdot 4^3)}}} + \frac{\mu^2}{\delta e^{c n \varepsilon}}.
    \end{align*}

    Plugging in $\tau^2 \asymp \rho \log(2n/\gamma)$ and choosing $\gamma = 1/2$ recovers the statement. 
\end{proof}

\paragraph{Proofs for Failure of Stable Histogram}

\begin{lemma}
    \label{lem:binthresholding}
    Fix observations $x \in \R^n$ and bins $(B_k)_{k\in\Z}$. Define $\hat p_k := \frac 1 n \sum_{i=1}^n \indicator{x_i \in B_k}$ and assume that $\hat p_k \geq \alpha$ with $\alpha > 0$. Then, the output $\tilde p_k$ of Algorithm \ref{alg:stablehist} fulfills 
    \begin{align*}
        \P\Parent{\curly{\tilde p_k = 0} \cap \curly{\hat p_k \geq \alpha}}
        \leq \frac{e^{\varepsilon/2}}{\delta} \cdot \exp\Parent{-\frac{\alpha n \varepsilon}{2}}.
    \end{align*}
\end{lemma}

\begin{proof}

    By design of Algorithm \ref{alg:stablehist}, when $\tilde p_k = 0$, straightforward calculations yield 
    \begin{align*}
        \P\Parent{\curly{\tilde p_k = 0} \cap \curly{\hat p_k \geq \alpha}}
        &= \P\Parent{\Curly{\xi_k < \frac{2}{\varepsilon n}\log\Parent{\frac 1 \delta} + \frac 1 n - \hat p_k} \cap \curly{\hat p_k \geq \alpha}} \\
        &\leq \P\Parent{\xi_k < \frac{2}{\varepsilon n}\log\Parent{\frac 1 \delta} + \frac 1 n - \alpha} \\
        &= \P\Parent{\xi_k > \alpha - \frac 1 n - \frac{2}{\varepsilon n} \log\Parent{\frac 1 \delta}} \\
        &\leq \exp\Parent{-\frac{\varepsilon n}{2}\Curly{\alpha - \frac 1 n - \frac{2}{\varepsilon n} \log\Parent{\frac 1 \delta}}} \\
        &= \frac{e^{\varepsilon / 2}}{\delta} \cdot \exp\Parent{- \frac{\alpha n \varepsilon}{2}}. 
    \end{align*}

    Here, we use that $-\xi_k = \xi_k$ in distribution and we apply a standard Laplace tail bound. 
\end{proof}

\begin{lemma}
    \label{lem:histfailure}
    Let $\gamma \in (0, 1\wedge \frac n 4)$ and let $X^n \in \R^n$ be $(\tau, \gamma)^{x_0}_\infty$-concentrated. Then, for Algorithm \ref{alg:stablehist}
    \begin{align*}
        \P\Parent{\forall k\in \Z : \tilde p_k = 0}
        \leq \frac{e^{\varepsilon/2}}{\delta} \cdot \exp\Parent{-\frac{n\varepsilon}{8}} +
        \begin{cases}
            2 \exp\Parent{-\frac{n}{128}} & \text{if $X^n$ fulfills Assumption \ref{ass:iiddata}}, \\
            64 \exp\Parent{-\frac{2n}{48^3 \rho M^2}} & \text{if $X^n$ fulfills Assumption \ref{ass:logsobdep}}.
        \end{cases}
    \end{align*}
\end{lemma}

\begin{proof}

    We start by reducing the problem of bounding the probability that all bins are zero to bounding the probability that a bin with high empirical mass is thresholded.

    Define $\mathcal E_{zero} := \curly{\tilde p_k = 0, \,\forall k\in \Z}$ and $\mathcal E_{zero}^k := \curly{\tilde p_k = 0}$. Note that 
    \begin{align*}
        \P\Parent{\mathcal E_{zero}}
        = \P\Parent{\forall k \in \Z : \tilde p_k = 0}
        \leq \P\Parent{\tilde p_l = 0}
        = \P\Parent{\mathcal E_{zero}^l},
        \quad \text{for all \,$l \in \Z$}.
    \end{align*}

    As we want to apply Lemma \ref{lem:binthresholding}, we want to use this for a bin with high mass and accordingly also high empirical mass. By $(\tau, \gamma)_\infty$-concentration, for all $i \in [n]$ with probability at least $1-\gamma/n =: 1-\alpha$ we have $X_i \in I := [x_0 \pm \tau]$. We construct a set $S$ of bin-indices $k \in \Z$ that cover $I$: 
    \begin{align*}
        S := \{k \in \Z : I \cap B_k \neq \emptyset\}.
    \end{align*}

    Note that $|S|\leq 2$, \eqref{eq:lb_S} and \eqref{eq:pigeon_hole} hold, and take $k^\star$ as in \eqref{eq:pigeon_hole}.

    We will now bound $\P(\mathcal E_{zero}^{k^\star})$ assuming that $X^n$ fulfills the \iid{} Assumption \eqref{ass:iiddata}. We split it on the event where $\hat p_{k^\star}$ estimates $p_{k^\star}$ well and where it does not: 
    \begin{align*}
        \P\Parent{\mathcal E_{zero}^{k^\star}}
        &=\P\Parent{\mathcal E_{zero}^{k^\star} \cap \Curly{\max_{k \in \Z} \abs{\hat p_k - p_k} < \eta}} + \P\Parent{\mathcal E_{zero}^{k^\star} \cap \Curly{\max_{k \in \Z} \abs{\hat p_k - p_k} \geq \eta}} \\
        &\leq \P\Parent{\mathcal E_{zero}^{k^\star} \cap \Curly{\max_{k \in \Z} \abs{\hat p_k - p_k} <\eta}} + \P\Parent{\max_{k \in \Z} \abs{\hat p_k - p_k} \geq \eta} \\
        &\leq \P\Parent{\mathcal E_{zero}^{k^\star} \cap \Curly{\abs{\hat p_{k^\star} - p_{k^\star}} < \eta}} + 2\exp\Parent{-\frac{n\eta^2}{2}}.
    \end{align*}

    Here, the last inequality holds by Lemma \ref{lem:estlossemphist}. On the event $\mathcal E_{zero}^{k^\star} \cap \curly{\abs{\hat p_{k^\star} - p_{k^\star}} < \eta}$ we know that there is $k^\star \in \Z$ \st{} $p_{k^\star} \geq \frac{1-\alpha}{2}$ and therefore $\hat p_{k^\star}> p_{k^\star} - \eta \geq \frac{1-\alpha-2\eta}{2}$. When $\frac{1-\alpha-2\eta}{2} > 0$ we know that $\hat p_{k^\star} > 0$ and hence $\tilde p_{k^\star}$ can only be 0 through thresholding. Formally, by Lemma \ref{lem:binthresholding}
    \begin{align*}
        \P\Parent{\mathcal E_{zero}^{k^\star} \cap \Curly{\abs{\hat p_{k^\star} - p_{k^\star}} \leq \eta}}
        &\leq \mathbb{E}\left[\P\Parent{\curly{\tilde p_{k^\star} = 0} \cap \Curly{\hat p_{k^\star} \geq \frac{1-\alpha-2\eta}{2}} \Big |X^n}\right] \\
        &\leq \frac{e^{\varepsilon/2}}{\delta} \cdot \exp\Parent{-\frac{(1-\alpha-2\eta) n \varepsilon}{4}} \\
        &\leq \frac{e^{\varepsilon/2}}{\delta} \cdot \exp\Parent{-\frac{(3/4-2\eta) n \varepsilon}{4}}.
    \end{align*}

    For the last inequality we use that $\alpha < 1/4$ by assumption. We set $\eta = 1/8$ and collect terms: 
    \begin{align*}
        \P\Parent{\mathcal E_{zero}}
        \leq \P\Parent{\mathcal E_{zero}^{k^\star}}
        \leq \frac{e^{\varepsilon/2}}{\delta} \cdot \exp\Parent{-\frac{n\varepsilon}{4\cdot 2}} + 2 \exp\Parent{-\frac{n}{2\cdot 8^2}}. 
    \end{align*}

    The proof under Assumption \ref{ass:logsobdep} is analogous. 
\end{proof}

\paragraph{Auxiliary Results for Asymptotic Analysis}

\begin{corollary}
    \label{cor:varL2projlogsob}
    Let $X^n \in \R^n$ fulfill Assumption \ref{ass:logsobdep} with constants $(\rho, \infty)$. For a fixed closed convex set $C \subseteq \R$ and the $L_2$-projection $\proj{C}{x} := \arg \min_{y \in C} \abs{x - y}^2$ we have 
    \begin{align*}
        \Var{\frac 1 n \sum_{i=1}^n \proj{C}{X_i}}
        \leq \frac \rho n. 
    \end{align*}
\end{corollary}

\begin{proof}   
    Since $X^n$ fulfills Assumption \ref{ass:logsobdep} with constants $(\rho, \infty)$, the distribution of $X^n$ fulfills $\LSI{\rho}$. The non-expansiveness of the $L_2$-projection shows that for all $x,y \in \R^n$ we have
    \begin{align*}
        \Abs{\frac 1 n \sum_{i=1}^n \proj{C}{x_i} - \frac 1 n \sum_{i=1}^n \proj{C}{y_i}}
        \leq \frac 1 n \sum_{i=1}^n \abs{\proj{C}{x_i} - \proj{C}{y_i}}
        \leq \frac 1 n \sum_{i=1}^n \abs{x_i - y_i}
        \leq \frac{1}{\sqrt n} \twonorm{x-y}. 
    \end{align*}

    Thus, $f(x) := \frac 1 n \sum_{i=1}^n \proj{C}{x_i}$ is $1/\sqrt n$-Lipschitz. By a Poincar{\'e} inequality
    \begin{align*}
        \Var{\frac 1 n \sum_{i=1}^n \proj{C}{X_i}}
        = \var{f(X)}
        \leq \rho \cdot \E\bracket{\twonorm{\nabla f(X)}^2}
        \leq \frac \rho n. 
    \end{align*}
\end{proof}

\begin{corollary}
    \label{cor:varL2projiid}
    Let $X^n \in \R^n$ fulfill Assumption \ref{ass:iiddata} and have $\sigma^2$-sub-Gaussian entries. For a fixed closed convex set $C \subseteq \R$ and the $L_2$-projection $\proj{C}{x} := \arg \min_{y \in C} \abs{x - y}^2$ we have 
    \begin{align*}
        \Var{\frac 1 n \sum_{i=1}^n \proj{C}{X_i}}
        \leq \frac{\sigma^2}{n}. 
    \end{align*}
\end{corollary}

\begin{proof}

    By independence of $X_1,\dots ,X_n$ we have the equality in the following: 
    \begin{align*}
        \Var{\frac 1 n \sum_{i=1}^n \proj{C}{X_i}}
        = \frac{1}{n^2} \sum_{i=1}^n \var{\proj{C}{X_i}}
        \leq \frac{1}{n^2} \sum_{i=1}^n \var{X_i}
        \leq \frac{\sigma^2}{n}.
    \end{align*}

    The last inequality holds as $\var{X_i} \leq \sigma^2$ for sub-Gaussian $X_i$ \citep[Exercise 2.40]{vershynin2025}.
\end{proof}

\begin{corollary}
    \label{cor:subgaussmaxineqs}
    Let $Z_1,\dots ,Z_n$ have mean $\mu$ and be $\sigma^2$-sub-Gaussian. Then, 
    \begin{align*}
        \E\bracket{\max_{i\in [n]} \abs{Z_{i} - \mu}^4}
        \lesssim \sigma^4 \log(2n)^2 
        \quad \text{and} \quad 
        \E\bracket{\max_{i\in [n]} \abs{Z_{i} - \mu}^2}
        \lesssim \sigma^2 \log(2n).
    \end{align*}
\end{corollary}

\begin{proof}

    First note that a union bound leads to the following tail-bound: 
    \begin{align*}
        \P\Parent{\max_{i\in [n]} \abs{Z_{i} - \mu} \geq t}
        = \P\Parent{\bigcup_{i\in[n]} \curly{\abs{Z_{i} - \mu} \geq t}}
        \leq \sum_{i\in[n]} \P\Parent{\abs{Z_{i} - \mu} \geq t}
        \leq 2n \exp\Parent{-\frac{t^2}{2\sigma^2}}. 
    \end{align*}

    Taking the 4-th power, the change of variables $a = t^4 \Leftrightarrow t^2 = \sqrt a$ immediately give us
    \begin{align*}
        \P\Parent{\max_{i\in [n]} \abs{Z_{i} - \mu_j}^4 \geq a}
        \leq 2n \exp\Parent{-\frac{\sqrt{a}}{2\sigma^2}}. 
    \end{align*}

    The desired in-expectation results follows from the integral identity and the tails bounds we have established. Indeed,
    \begin{align*}
        \E\bracket{\max_{i\in [n]} \abs{Z_{i} - \mu}^4}
        = \int_0^\infty \P\Parent{\max_{i\in [n]} \abs{Z_{i} - \mu}^4 \geq a} da
        \leq 2n \int_0^\infty \exp\Parent{-\frac{\sqrt{a}}{2\sigma^2}} da
        = 16n\sigma^4
    \end{align*}

    The order of this bound can be improved as follows: 
    \begin{align*}
        \E\bracket{\max_{i\in [n]} \abs{Z_{i} - \mu}^4}
        &= \int_0^\infty \P\Parent{\max_{i\in [n]} \abs{Z_{i} - \mu}^4 \geq a} da \\
        &\leq \int_0^b 1 da + \int_b^\infty 2n \exp\Parent{-\frac{\sqrt{a}}{2\sigma^2}} da \\
        &\leq b + 2n \Parent{8\sigma^4 + 4\sigma^2 \sqrt b} e^{-\frac{\sqrt b}{2\sigma^2}}
    \end{align*}

    Minimizing over $b$ then yields $b = 4\sigma^4\log(2n)^2$ and the maximal inequality becomes
    \begin{align*}
        \E\bracket{\max_{i\in [n]} \abs{Z_{i} - \mu}^4}
        &\leq b + 2n \Parent{8\sigma^4 + 4\sigma^2 \sqrt b} e^{-\frac{\sqrt b}{2\sigma^2}}
        = 4\sigma^4 \log(2n)^2 + 8\sigma^4 + \frac{8\sigma^4\log(2n)}{2n}. 
    \end{align*}

    The first statement follows as the above is bounded in order by $\sigma^4 \log(2n)^2$. The second statement follows because higher moment bounds imply lower moment bounds by Jensen's inequality. 
\end{proof}

\subsection{Proofs for Nonparametric Regression}
\label{app:nonparametricreg}

\begin{proof}[Proof of Lemma \ref{lem:concpriestchaoest}]\hfill

    Throughout, we extend $f$ by 0 outside of its support $[0,1]$. The proof starts with a triangle inequality that lets us decompose the mean absolute error of $f, \hat f_n$ into bias and mean absolute deviation, 
    \begin{align*}
        | f(x) - \hat f_n(x)| 
        &\leq \underbrace{|f(x) - \E[\hat f_n(x)]|}_{\operatorname{Bias}(\hat f_n)} + \underbrace{|\E[\hat f_n(x)] - \hat f_n(x)|}_{\operatorname{Mad}(\hat f_n)}. 
    \end{align*}

    Introducing the integral $I := \int_\R f(t) \frac 1 bK(\frac{x-t}{b})dt$ , the bias can be further decomposed by adding and subtracting $I$ and using a triangle inequality: 
    \begin{align*}
        \operatorname{Bias}(\hat f_n)
        &= |f(x) - \E[\hat f_n(x)]|
        \leq \abs{f(x) - I} + \abs{I - \E[\hat f_n(x)]} \\
        &\overset{(\star)}\leq b \Parent{L_f\mu_{1}(K) + \infnorm{f}} + \frac{L_f \infnorm{K} + \infnorm{f} L_K}{nb}.
    \end{align*}

    Here, it remains to establish $(\star)$. We handle both terms separately, starting as follows: 
    \begin{align*}
        \Abs{f(x) - I}
        &=\Abs{\int_\R (f(x) - f(t))\frac{1}{b}K\Parent{\frac{t-x}{b}} dt} \\
        &\leq \int_0^1 \abs{f(x) - f(t)}\frac{1}{b}K\Parent{\frac{t-x}{b}} dt + \int_{\R \setminus [0,1]} \abs{f(x)} \frac{1}{b}K\Parent{\frac{t-x}{b}}dt 
        =: T + T_{tail}.
    \end{align*}

    By Lipschitzness of $f$, we obtain the following bound on $T$:
    \begin{align*}
        T
        = \int_0^1 \abs{f(x) - f(t)}\frac{1}{b}K\Parent{\frac{t-x}{b}} dt
        \leq L_f \int_{0}^{1} \Abs{x-t}\frac{1}{b}K\Parent{\frac{t-x}{b}} dt
        \leq b L_f \mu_1(K). 
    \end{align*}

    Given the assumption that $x \in [\zeta, 1-\zeta]$ and the 1-sub-Gaussian tails, 
    \begin{align*}
        T_{tail}
        \leq \infnorm{f} \int_{\R \setminus [-\frac{x}{b}, \frac{1-x}{b}]} K(u) du
        \leq \infnorm{f} \cdot \P\Parent{\abs{X} \geq \zeta/b}
        \leq 2e^{\zeta^2/b^2} \infnorm{f} 
        \leq b\infnorm{f}. 
    \end{align*}

    Above, the last inequality holds assuming $\zeta = b\sqrt{2\log(2/b)}$. For the second term we need to bound to obtain $(\star)$, our choice of the Kernel and the zero mean of $\epsilon_i$ yield
    \begin{align*}
        T_2 
        &= \Abs{I - \E[\hat f_n(x)]}
        = \Abs{\frac 1 b \int_0^1 f(t) K\Parent{\frac{t-x}{b}}dt - \frac{1}{nb} \sum_{i=1}^n \E[Y_i] K\Parent{\frac{x_i-x}{b}}} \\
        &= \frac{1}{b} \Abs{\int_0^1 f(t) K\Parent{\frac{t-x}{b}}dt - \frac{1}{n} \sum_{i=1}^n f(x_i) K\Parent{\frac{x_i-x}{b}}}
        \leq \frac{L_f \infnorm{K}}{nb} + \frac{\infnorm{f} L_K}{nb}. 
    \end{align*}

    Here, the last inequality is by Lemma \ref{lem:riemannsumapprox}. Unlike the deterministic bias bound, bounding the mean absolute deviation requires a probabilistic argument: 
    \begin{align*}
        \operatorname{Mad}(\hat f_n)
        = \abs{\E[\hat f_n(x)] - \hat f_n(x)}
        = \Abs{\frac{1}{n} \sum_{i=1}^n (f(x_i) - Y_i) \frac 1 bK\left(\frac{x_i-x}{b}\right)}
        = \frac{1}{n} \Abs{\sum_{i=1}^n \epsilon_i \frac 1 bK\left(\frac{x_i-x}{b}\right)}. 
    \end{align*}

    We now show that the $\operatorname{Mad}(\hat f_n)$ is an $\ell$-Lipschitz function of $\epsilon^n := (\epsilon_1,\dots , \epsilon_n)$ with $\ell = O((nb)^{-1/2})$. By the triangle and Cauchy-Schwarz inequality, for all $\epsilon, \epsilon' \in \R^n$ we have 
    \begin{align*}
        \frac{1}{n} \Abs{\Abs{\sum_{i=1}^n \epsilon_i' \frac 1 bK\left(\frac{x_i-x}{b}\right)} - \Abs{\sum_{i=1}^n \epsilon_i \frac 1 bK\left(\frac{x_i-x}{b}\right)}}
        &\leq \frac{1}{n} \Abs{\sum_{i=1}^n (\epsilon_i'-\epsilon_i) \frac 1 bK\left(\frac{x_i-x}{b}\right)} \\
        &
        \leq \frac{1}{n} \Parent{\sum_{i=1}^n \abs{\epsilon'_i-\epsilon_i}^2}^{1/2} \Parent{\sum_{i=1}^n \frac 1 bK\left(\frac{x_i-x}{b}\right)}^{1/2} \\
        &\leq \frac{1}{n} \twonorm{\epsilon' - \epsilon} \frac{\sqrt{n\infnorm{K}}}{\sqrt{b}}
        = \sqrt{\frac{\infnorm{K}}{nb}} \twonorm{\epsilon'-\epsilon}
        .
    \end{align*}

    By log-Sobolev Lipschitz concentration of \ref{thm:logsoblipconc} with constants $\ell=\sqrt{{\infnorm{K}}/{nb}}$ and $\sigma^2_{\max}$, 
    \begin{align*}
        \P\Parent{\operatorname{Mad}(\hat f_n) \geq \sqrt{\frac{2\sigma_{\max}^2 \infnorm{K} \log(2/\alpha)}{nb}}}
        = \P\Parent{\operatorname{Mad}(\hat f_n) \geq t}
        = \exp\Parent{-\frac{t^2}{2\ell^2\sigma^2_{\max}}}
        = \alpha. 
    \end{align*}

    On the complement and therefore with probability $1-\alpha$, combining this with our deterministic bound on the bias, we obtain the following bound
    \begin{align*}
        \abs{f(x) - \hat f_n(x)}
        \leq 
        b \Parent{L_f\mu_{1}(K) + \infnorm{f}} + \frac{L_f \infnorm{K} + \infnorm{f} L_K}{nb} + 
        \sqrt{\frac{2\sigma_{\max}^2 \infnorm{K} \log(2/\alpha)}{nb}}.
    \end{align*}

    We now obtain $(\tau, \gamma)_\infty$-concentration from the general statement. As it holds for $n=1$ and since $\hat f_i(x) = Y_i\frac{1}{b}K\left(\frac{x_i-x}{b}\right)$, for any single $(x_i, Y_i)$ with probability at least $1-\alpha$
    \begin{align*}
        \abs{f(x) - \hat f_1(x)} 
        \leq b \Parent{L_f\mu_{1}(K) + \infnorm{f}} + \frac{L_f \infnorm{K} + \infnorm{f} L_K}{b} + \sqrt{\frac{2\sigma_{\max}^2 \infnorm{K} \log(2/\alpha)}{b}}. 
    \end{align*}

    Since $b \leq 1 \Leftrightarrow 1 \leq 1/b$, the statement above yields $(\tau, \gamma)_\infty$-concentration of $\hat f^n(x)$. Thereby, 
    \begin{align*}
        \tau 
        = b \Parent{L_f\mu_{1}(K) + \infnorm{f}} + \frac{1}{b} \Parent{L_f \infnorm{K} + \infnorm{f} L_K + \sqrt{2 \sigma^2_{\max} \infnorm{K} \log(2n/\gamma)}}. 
    \end{align*}
Finally, we note that since $K$ is a $1$-sub-Gaussian density we have $\mu_1(K) \leq 1$.
\end{proof}

\begin{proof}[Proof of Lemma \ref{lem:logsobdeppriestchaoest}]\hfill 

    Define $f^n := (f(x_1),\dots ,f(x_n))\T$ and $D := \frac 1 b \diagonal{K_b(x_1),\dots ,K_b(x_n)}$, with $K_b(t) := K(\frac{x-t}{b})$. Note that $D_{ii} = \frac 1 b K_b(x_i) \leq \frac{\infnorm{K}}{b}$. By assumption $\epsilon^n \sim \mathcal N(0,\Sigma^n)$ in $\operatorname{LSI}(\sigma^2_{\max})$. As the $x_i$ are fixed, 
    \begin{equation*}    
        Z_\epsilon
        := D Y
        = D (f^n + \epsilon^n)
        = D f^n + D \epsilon^n, \quad \mathbb{E}[ Z_\epsilon]=D f^n  \quad\mbox{ and }\quad \cov{Z_\epsilon}  = D \Sigma^n D\T.
    \end{equation*}

    By Gaussian invariance under affine transformations $Z_\epsilon$ is Gaussian and more concretely $Z_\epsilon \sim \mathcal N(Df^n, D\Sigma^n D\T)$. Now, by Example \ref{exa:gausslogsob} the log-Sobolev constant of this distribution is given by the operator norm of the covariance matrix which we bound as follows
    \begin{align*}
        \opnorm{\cov{Z_\epsilon}}
        &\leq \opnorm{D\Sigma^n D}
        \leq \opnorm{D}^2 \opnorm{\Sigma^n}
        \leq \frac{\sigma^2_{\max} \infnorm{K}}{b^2}
        =: \rho. 
    \end{align*}

    Any marginal of $Z_\epsilon$ is Gaussian with variance $\sigma_i^2 := \var{(Z_\epsilon)_i} = {\Sigma_{ii}}/{b^2} \geq \sigma_{\min}^2/b^2$. Thus, 
    \begin{align*}
        d\mathcal N(x; \mu_i, \sigma_i^2) \leq (2\pi\sigma_i^2)^{-1/2} \leq (2\pi\sigma_{\min}^2/b^2)^{-1/2} =: M.
    \end{align*}

     Accordingly, $\rho M^2 \lesssim {\sigma^2_{\max}}/{\sigma^2_{\min}} \cdot \infnorm{K}$. 
\end{proof}

\begin{proof}[Proof of Corollary \ref{cor:rMSEpriestchaoest}]\hfill 

    In Lemma \ref{lem:concpriestchaoest} we derived $(\tau, \gamma)_\infty$-concentration of $\hat f^n(x)$ around $f(x)$ with
    \begin{align*}
        \tau 
        &\lesssim b + \frac{1}{b} \Parent{1 + \sqrt{\sigma^2_{\max} \infnorm{K} \log(2n/\gamma)}}
        \lesssim \frac{1}{b} \sqrt{\sigma^2_{\max} \infnorm{K} \log(2n/\gamma)}
        =: \frac{C}{b}.
    \end{align*}
    
    There, we also derived a finite sample error bound showing that with probability at least $1-\alpha,$ 
    \begin{align*}
        \Abs{\hat f_n(x) - f(x)}
        &\lesssim 
        b + \frac{1}{nb} + 
        \sqrt{\frac{\sigma_{\max}^2 \infnorm{K} \log(2/\alpha)}{nb}}. 
    \end{align*}

    Moreover, we showed in Lemma \ref{lem:logsobdeppriestchaoest} that $\hat f^n(x)$ fulfills Assumption \ref{ass:logsobdep}. This makes Theorem \ref{thm:rMSEwinsmeanHD} with $d=1$ applicable under Assumption \ref{ass:logsobdep}. Thus, via a union bound and setting $\alpha = \gamma$, with probability at least $1-3\gamma-O(n/\delta \cdot e^{-\kappa n \varepsilon})$, 
    \begin{align*}
        \Abs{\mathcal A(\hat f^n(x)) - f(x)} 
        &\lesssim \Abs{\hat f_n(x) - f(x)} + \tau \sqrt{\frac{\log(1/\varrho)\log(3/\gamma)^2}{n^2\varepsilon^2}} \\
        &\lesssim \Abs{\hat f_n(x) - f(x)} + \sqrt{\frac{C^2\log(1/\varrho)\log(3/\gamma)^2}{b^2n^2\varepsilon^2}} \\
        &\lesssim b + \frac{1}{nb} + \sqrt{\frac{\sigma_{\max}^2 \infnorm{K} \log(2/\gamma)}{nb}} + \sqrt{\frac{C^2\log(1/\varrho)\log(3/\gamma)^2}{n^2b^2\varepsilon^2}}.
    \end{align*}

    Modulo logarithms, assuming $\sigma^2_{\max} \gtrsim 1$ and as $1 \leq 1 / \varepsilon$ matching the second with the fourth term, 
    \begin{align*}
        \Abs{\mathcal A(\hat f^n(x)) - f(x)} 
        &\lessapprox b + \frac{1}{nb} + \sigma_{\max} \Parent{\sqrt{\frac{1}{nb}} + \sqrt{\frac{1}{n^2b^2\varepsilon^2}}}
        \lesssim b + \sigma_{\max} \Parent{\sqrt{\frac{1}{nb}} + \sqrt{\frac{1}{n^2b^2\varepsilon^2}}}.
    \end{align*}

    We call the three terms in the \rhs{} $T_1, T_2$ and $T_3$. To choose the bandwidth $b$, we balance $T_1, T_2$ and $T_1, T_3$ separately. For this, we perform a case distinction on whether $T_2$ or $T_3$ dominate: 
    \begin{enumerate}
        \item[] \textbf{Case 1}: If $T_2 \gtrsim T_3$, i.e., when $\varepsilon \gtrsim 1/\sqrt{nb}$ we balance $T_1$ and $T_2$ as follows: 
        \begin{align*}
            T_1 \asymp T_2 
            \quad \Leftrightarrow \quad 
            b \asymp \frac{\sigma_{\max}}{\sqrt{nb}}
            \quad \Leftrightarrow \quad 
            b \asymp \Parent{\frac{\sigma^2_{\max}}{n}}^{1/3}
            \quad \Rightarrow \quad 
            \varepsilon \gtrsim \Parent{\frac{1}{n \sigma_{\max}}}^{1/3}.
        \end{align*}

        \item[] \textbf{Case 2}: If $T_3 \gtrsim T_2$, i.e., when $\varepsilon \lesssim 1/\sqrt{nb}$ we balance $T_1$ and $T_3$ as follows: 
        \begin{align*}
            T_1 \asymp T_3
            \quad \Leftrightarrow \quad
            b \asymp \frac{\sigma_{\max}}{\sqrt{n^2b^2\varepsilon^2}}
            \quad \Leftrightarrow \quad 
            b \asymp \sqrt{\frac{\sigma_{\max}}{n\varepsilon}}
            \quad \Rightarrow \quad
            \varepsilon \lesssim \Parent{\frac{1}{n\sigma_{\max}}}^{1/3}. 
        \end{align*}
    \end{enumerate}

    Hence, $T_2$ dominates when $\varepsilon \lesssim (n\sigma_{\max})^{-1/3}$ and otherwise $T_3$. To incorporate both cases, we choose $b \asymp (\sigma^2_{\max}/n)^{1/3} \vee (\sigma_{\max}/n\varepsilon)^{1/2}$ and plug this into the bound on the rMSE. Doing the case distinction, we obtain the statement by bounding the maximum by the sum of both terms. 
\end{proof}

\paragraph{Auxiliary Results for Nonparametric Regression}

\begin{lemma}{(Adjusted from Lemma A.1, \citet{gugushvili:klaassen2012}).}
    \label{lem:riemannsumapprox}
    Let $f : [0,1] \to \R$ be $L_f$-Lipschitz. Further, assume that the kernel $K : \R \to \R_\geq$ is $L_K$-Lipschitz. Then, with the mesh $x_i := i/n$ for $0 \leq i \leq n$, for all $x \in [0,1]$, 
    \begin{align*}
        \frac{1}{b} \Abs{\int_0^1 f(t) K\Parent{\frac{x-t}{b}} dt - \frac 1 n \sum_{i=0}^{n-1} f(x_i) K\Parent{\frac{x-x_i}{b}}}
        \leq \frac{L_f\infnorm{K}}{nb} + \frac{L_K\infnorm{f}}{nb}.
    \end{align*}

    If instead of Lipschitzness $K(u) := \frac 1 2 \indicator{|u| \leq 1}$ then the same holds with $L_K = 1$. 
\end{lemma}

\begin{proof}

    For brevity of notation denote the \lhs{} of the statement by $I$ and $K_b(u) := K\Parent{\frac{x-u}{b}}$. To bound $I$ we use that the $x_i$ are a grid on $[0,1]$ and $|x_{i+1} - x_i| = 1/n$. Concretely, we first split the integral into intervals $[x_i, x_{i+1}]$. Second, we add and subtract $f(t) K(x_i)$ to use a triangle inequality: 
    \begin{align*}
        I 
        &= \frac{1}{b}\Abs{\sum_{i=0}^{n-1} \int_{x_i}^{x_{i+1}} f(t) K_b(t) dt - \frac 1 n \sum_{i=0}^{n-1} f(x_i) K_b(x_i)}
        \leq \frac{1}{b} \sum_{i=0}^{n-1} \int_{x_i}^{x_{i+1}} \Abs{f(t) K_b(t) - f(x_i) K_b(x_i)} dt \\
        &\leq \frac{1}{b} \sum_{i=0}^{n-1} \int_{x_i}^{x_{i+1}} \Abs{f(t) K_b(t) - f(t) K_b(x_i)} dt + \frac{1}{b} \sum_{i=0}^{n-1} \int_{x_i}^{x_{i+1}} \Abs{f(t) K_b(x_i) - f(x_i)K_b(x_i)} dt \\
        &\leq \frac{1}{b} \sum_{i=0}^{n-1} \int_{x_i}^{x_{i+1}} \Abs{f(t)} \Abs{K_b(t) - K_b(x_i)}dt + \frac{1}{b} \sum_{i=0}^{n-1} \int_{x_i}^{x_{i+1}} \Abs{K_b(x_i)} \Abs{f(t) - f(x_i)}dt \\
        &\leq \frac{L_K\infnorm{f}}{b} \sum_{i=0}^{n-1} \int_{x_i}^{x_{i+1}} \Abs{t-x_i}dt + \frac{L_f\infnorm{K}}{b} \sum_{i=0}^{n-1} \int_{x_i}^{x_{i+1}} \Abs{t-x_i}dt\\
        &
        \leq \frac{L_K\infnorm{f}}{nb} + \frac{L_f\infnorm{K}}{nb}.
    \end{align*}

    Here, the last inequality holds because $\abs{t-x_i} \leq 1/n$ and $\abs{x_{i+1} - x_i} = 1/n$, and thus 
    \begin{align*}
        \sum_{i=0}^{n-1} \int_{x_i}^{x_{i+1}} \abs{t-x_i} dt 
        \leq \frac 1 n \sum_{i=0}^{n-1} [t]_{x_i}^{x_{i+1}} 
        = \frac 1 n \sum_{i=0}^{n-1} x_{i+1} - x_i
        \leq \frac 1 n \sum_{i=0}^{n-1} \abs{x_{i+1} - x_i} 
        = \frac 1 n.
    \end{align*}

    For the special case when $K(u) = \frac 1 2 \indicator{|u| \leq 1}$ we have 
    \begin{align*}
        I
        &\leq \frac{1}{b} \sum_{i=0}^{n-1} \int_{x_i}^{x_{i+1}} \Abs{f(t)} \Abs{K_b(t) - K_b(x_i)}dt + \frac{1}{b} \sum_{i=0}^{n-1} \int_{x_i}^{x_{i+1}} \Abs{K_b(x_i)} \Abs{f(t) - f(x_i)}dt \\
        &\leq \frac{\infnorm{f}}{b} \sum_{i=0}^{n-1} \int_{x_i}^{x_{i+1}} \Abs{K_b(t) - K_b(x_i)} dt + \frac{L_f\infnorm{K}}{nb}. 
    \end{align*}

    Since $K$ is neither Lipschitz nor continuous we have to make a more direct argument. Note that
    \begin{align*}
        \one(t,x_i)
        := \Abs{K_b(t) - K_b(x_i)}
        &= \frac 1 2 \Abs{\indicator{x-b \leq t \leq x+b} - \indicator{x-b \leq x_i \leq x+b}} \\
        &= \frac 1 2 \indicator{\Curly{x-b \leq t \leq x+b} \operatorname{ or} \Curly{x-b \leq x_i \leq x+b}}. 
    \end{align*}

    This can only evaluate to $1$ at the boundary $x-b$ or $x+b$ and hence $\one(t,x_i) \neq 0$ only for two $x_l$ and $x_u$ such that $x-b \in [x_l, x_{l+1}]$ and $x+b \in [x_u, x_{u+1}]$. Thus, 
    \begin{align*}
        \sum_{i=0}^{n-1} \int_{x_i}^{x_{i+1}} \Abs{K_b(t) - K_b(x_i)} dt
        &\leq \Parent{\int_{x_l}^{x_{l+1}} K(t, x_l)dt + \int_{x_u}^{x_{u+1}} \one(t, x_u)dt} \\
        &\leq \frac 1 2 \Parent{\int_{x_l}^{x_{l+1}}dt + \int_{x_u}^{x_{u+1}}dt}
        = \frac 1 n. 
    \end{align*}

    Plugging into our bound on $I$ we recover the statement $I \leq \frac{\infnorm{f}}{nb} + \frac{L_f\infnorm{K}}{nb}$. 
\end{proof}

\section{Proofs of Section \ref{sec:uDP}}

\subsection{Proofs for User-Level Mean Estimation}
\label{app:userlvlmeanest}

\begin{proof}[Proof of Lemma \ref{lem:concuserlvlmeanest}]\hfill 

    First, we show that $\hat \mu^n$ is $(\tau, \gamma)_\infty$-concentrated. To justify Remark \ref{rem:taugammamaximal}, this proof is more elaborate than necessary, involving marginal log-Sobolev constants. Second, we show the bound on $\twonorm{\bar X_{nT} -\mu}$. 

    By Assumption \ref{ass:logsobdep} the distribution of $X^n_{\cdot j}$ is $\operatorname{LSI}(\rho)$ for all $j \in [d]$. For all $u \in [n]$ and $j \in [d]$, the $(X_u)_{\cdot j}$ are sub-matrices of $X^n_{\cdot j}$ and thus have distributions $\mathcal D_{uj}$ that are marginals of the distributions of $X^n_{\cdot j}$. By Lemma \ref{lem:lipcontlogsob} $\mathcal D_{uj}$ fulfills a log-Sobolev inequality with constant $\rho$ too. 
    
    However, because log-Sobolev constants are defined as the smallest constant \st{} a log-Sobolev inequality holds, the log-Sobolev constants $\rho_{uj}$ of $\mathcal D_{uj}$ can be smaller than $\rho$. Using these and noting that $(\hat \mu_u)_j = \frac 1 T (X_u)_{\cdot j}\T \one_T$, by Lemma \ref{lem:logsobconcmean} we have
    \begin{align*}
        \P\Parent{\abs{(\hat \mu_u)_j - \mu_j} \leq \sqrt{\frac{2\rho_{uj} \log(2/\alpha)}{T}}}
        \geq 1 - \alpha. 
    \end{align*}

    Union bounding over $j \in [d]$ and setting $\alpha = \gamma/dn$ yields $(\tau, \gamma)_\infty$-concentration with
    \begin{align*}
        \tau 
        = \max_{u \in [n]} \max_{j \in [d]} \sqrt{\frac{2\rho_{uj} \log(2dn/\gamma)}{T}}
        \leq \sqrt{\frac{2\rho \log(2dn/\gamma)}{T}}. 
    \end{align*}

    The characterization $\bar X_{nT} = \frac{1}{n} (\hat \mu^n)\T \one_{n}$ allows us to invoke the $L_2$-norm result of Lemma \ref{lem:logsobconcmean}: 
   \begin{align*}
        \P\Parent{\Twonorm{\bar X_{nT} - \mu} \leq \sqrt{\frac{2\rho \log(2d/\alpha)}{nT}}}
        \geq 1 - \alpha. 
    \end{align*}
\end{proof}

\begin{proof}[Proof of Corollary \ref{cor:rMSEuserlvldes}] \hfill

    This result is a direct consequence of Theorem \ref{thm:rMSEwinsmeanHD} under Assumption \ref{ass:logsobdep}. By Lemma \ref{lem:concuserlvlmeanest} $\hat \mu^n$ is $(\tau, \gamma)_\infty^\mu$-concentrated with $\tau = \sqrt{2\rho\log(2dn/\gamma)/T}$ and with probability at least $1-\alpha$
    \begin{align*}
        \Twonorm{\bar X_{nT} - \mu}
        \leq \sqrt{\frac{2d\rho \log(2d/\alpha)}{nT}}.
    \end{align*}

    Setting $\alpha = d\gamma$ and combing this with Theorem \ref{thm:rMSEwinsmeanHD} via a union bound, with probability at least $1-3d\gamma-O(d^2 n/\delta \cdot e^{-\kappa n \varepsilon^\prime})$ we have 
    \begin{align*}
        \Twonorm{\mathcal A(\hat \mu^n) - \mu} 
        &\lesssim \Twonorm{\bar X_{nT} - \mu} + \tau \sqrt{\frac{d^2\log(1/\varrho)\log(3/(d\gamma))^2}{n^2\varepsilon^2}} \\
        &\lesssim \sqrt{\frac{2d\rho \log(2/\gamma)}{nT}} + \sqrt{\frac{2\rho\log(2dn/\gamma)}{T}} \sqrt{\frac{d^2\log(1/\varrho)\log(3/(d\gamma))^2}{n^2\varepsilon^2}} \\
        &\lesssim \sqrt{\frac{d\rho \log(2/\gamma)}{nT}} + \sqrt{\frac{d^2\rho\log(2dn/\gamma)\log(1/\varrho)\log(3/(d\gamma))^2}{Tn^2\varepsilon^2}}. 
    \end{align*}
\end{proof}

\begin{proof}[Proof of Corollary \ref{cor:consistwinsmeanHDuserlvl}] \hfill

    By definition of the per-user means collected in $\hat \mu^n$ we have
    \begin{align*}
        (\hat \mu_u)_j
        = \frac 1 T \Parent{X_u\T \one_T}_j
        = \frac 1 T (X_u)_{\cdot j}\T \one_T
        = \frac 1 T \sum_{i=1}^T (X_u)_{ij}.
    \end{align*}
    
    Under Assumption \ref{ass:iiddata} the entries $(X_u)_{ij}$ are $\rho$-sub-Gaussian and independent. Hence, the mean $(\hat \mu_u)_j$ is $\rho/T$-sub-Gaussian. Under Assumption \ref{ass:logsobdep} we can use that $(\hat\mu_u)_j$ is $1/\sqrt T$-Lipschitz in $(X_u)_{\cdot j}$ and thus by Lemma \ref{lem:lipcontlogsob} its distribution has log-Sobolev constant $\rho/T$. The same holds for the entries of $\hat \mu_Z^n$. The result then follows immediately from Theorem \ref{thm:inexpub} with $\rho$ replaced by $ \rho/T$. 
\end{proof}

\subsection{Proofs for Random Effects Location Model}
\label{app:randeffects}

\begin{proof}[Proof of Lemma \ref{lem:randeffectslogsobdes}]\hfill 

    As the noise terms $U_g$ and $\epsilon_{ut}$ are independent and have zero-mean,
    \begin{align*}
        \E[Y_{gut}] 
        &= \E[\mu + U_g + \epsilon_{ut}]
        = \mu, \\
        \var{Y_{gut}}
        &= \var{U_g + \epsilon_{ut}}
        = \var{U_g} + \var{\epsilon_{ut}}
        = \sigma_U^2 + (\Sigma_u)_{tt}. 
    \end{align*}

    To show that $\mathcal D$ fulfills a Log-Sobolev inequality, we show that it is Gaussian and derive a bound on the operator norm of its covariance matrix $\Sigma^n$. Entry-wise, $Y^n$ is a sum of Gaussian random variables. Since the random variables $U_g$, $\epsilon_u$ are mutually independent for all $u \in [n], g \in [G]$, the joint is Gaussian. It has mean $\mu \one_{nT}$ and its covariance matrix $\Sigma^n$ has the following entries: 
    \begin{align*}
        \cov{Y_{gut}, Y_{ijk}}
        &= \cov{U_g + \epsilon_{ut}, U_i + \epsilon_{jk}} \\
        &= \cov{U_g, U_i} + \cov{U_g, \epsilon_{jk}} + \cov{\epsilon_{ut}, U_i} + \cov{\epsilon_{ut}, \epsilon_{jk}} \\
        &= \cov{U_g, U_i} + \cov{\epsilon_{ut}, \epsilon_{jk}}
    \end{align*}

    Here, we index with tuples $(gut)$ and $(ijk)$. Depending on the indices there are then four cases: 
    \begin{enumerate}
        \setlength\itemsep{-0.3em}

        \item[] \textbf{Case 1}: If $(gut) \neq (ijk)$ we have $\cov{Y_{gut}, Y_{ijk}} = 0$. 

        \item[] \textbf{Case 2}: If $(gut) = (ijk)$ we have $\cov{Y_{gut}, Y_{ijk}} = \sigma_U^2 + (\Sigma_u)_{tt}$. 

        \item[] \textbf{Case 3}: If $g = i, u \neq j$ we have $\cov{Y_{gut}, Y_{ijk}} = \cov{U_g, U_g} = \sigma_U^2$. 

        \item[] \textbf{Case 4}: If $g = i, u = j, t \neq k$ we have
        \begin{align*}
            \cov{Y_{gut}, Y_{ijk}} 
            = \cov{U_g, U_g} + \cov{\varepsilon_{ut}, \varepsilon_{uk}}
            = \sigma_U^2 + (\Sigma_u)_{tk}.
        \end{align*}
    \end{enumerate}

    Hence, the covariance matrix $\Sigma^n$ and the submatrix $\Sigma^U_g$ for a group $g$ of size $n_g$ have the form
    \begin{align*}
        \Sigma^n =
        \begin{pmatrix}
            \Sigma^U_1 & 0 & \dotsm & 0 \\
            0 & \Sigma^U_2 & \dotsm & 0 \\
            0 & 0 & \ddots & \vdots \\
            0 & 0 & \dotsm & \Sigma^U_G \\
        \end{pmatrix}, 
        \quad
        \Sigma^U_g = 
        \sigma_U^2 \one_{n_gT}\one_{n_gT}\T + 
        \begin{pmatrix}
            \Sigma_1 & 0 & \dotsm & 0 \\
            0 & \Sigma_2 & \dotsm & 0 \\
            0 & 0 & \ddots & \vdots \\
            0 & 0 & \dotsm & \Sigma_{n_g} \\
        \end{pmatrix}. 
    \end{align*}

    Together with the mean $\mu \one_{nT}$ this explicit form of $\Sigma^n$ specifies the Gaussian distribution $\mathcal D := \mathcal N(\mu \one_{nT}, \Sigma^n)$. By Example \ref{exa:gausslogsob} the log-Sobolev constant of $\mathcal D$ is therefore $\opnorm{\Sigma^n}$ or an upper bound thereof. As the operator norm of a block-diagonal matrix is the maximum norm of its blocks, 
    \begin{align*}
        \opnorm{\Sigma^n}
        = \max_{g \in [G]} \opnorm{\Sigma^U_g}
        \leq \max_{g \in [G]} \Parent{\opnorm{\sigma_U^2 \one_{n_gT}\one_{n_gT}\T} + \max_{u \in \mathcal G_g} \opnorm{\Sigma_u}}
        \leq \max_{g \in [G]} \Parent{\sigma_U^2 \cdot Tn_g + \sigma^2}.
    \end{align*}

    Here, $\mathcal G_g$ is the collection of users in group $g$. Now, we show that the marginal densities are bounded. Above we derived $\E[Y_{gut}] = \mu$ and $\var{Y_{gut}} = \sigma_U^2 + (\Sigma_u)_{tt} \leq \sigma_U^2 + \sigma^2 =: \sigma_Y^2$. We further know that any marginal of a Gaussian is also Gaussian. Therefore, for all $x \in \R$ the density of any $Y_{gut}$ is
    \begin{align*}
        \frac{1}{\sqrt{2\pi\sigma_Y^2}} \exp\Parent{-\frac{(x-\mu)^2}{2\sigma_Y^2}}
        \leq \frac{1}{\sqrt{2\pi\sigma_Y^2}}
        =: M. 
    \end{align*}

    Hence, $\frac{d}{dx}\bar F(x) = \frac 1 {nT} \sum_{u=1}^n \sum_{t=1}^T \frac{d}{dx} \P(Y_{gut} \leq x) \leq M$ for all $x \in \R$ and Assumption \ref{ass:logsobdep} holds. As a consequence the matrix $Y^n \in \R^{nT}$ fulfills the user-level log-Sobolev design of Definition \ref{def:userlvldatamatrix} because it has the required block structure and its entries $Y_{gut}$ have identical mean $\mu$. 
\end{proof}

\begin{proof}[Proof of Corollary \ref{cor:rMSErandeffects}] \hfill 

    The result is a direct consequence of Corollary \ref{cor:rMSEuserlvldes} enabled by Lemma \ref{lem:randeffectslogsobdes}. 
\end{proof}

\subsection{Proofs for Longitudinal Linear Regression}
\label{app:linreg}

\begin{proof}[Proof of Lemma \ref{lem:userlevelest}]\hfill 

    First, we derive the bias and covariance matrix for a general weighted OLS estimator. Based on this, we show unbiasedness of $\hat\beta, \hat\beta_u,$ and $\hat\beta_w, \hat\beta_{wu}$. Second, we prove the claims for the variances. 

    We begin reminding the reader of some properties of generalized least squares estimators. Note that for observations $\Upsilon = Z\theta + \epsilon$ with noise $\epsilon \in \R^l$ where $\epsilon \sim \mathcal N(0, \Lambda)$, the generalized least squares estimator $\hat \theta := (Z\T W Z)^{-1} Z\T W \Upsilon$ with weight matrix $W\in \R^{p\times p}$ and design $Z \in \R^{l\times p}$ is unbiased for $\theta$. Indeed, for fixed $W$ and $Z$ we have that
    \begin{equation}
        \label{eq:GLS_bias}
        \E[\hat\theta]
        = (Z\T W Z)\inv Z\T W \E[Z\theta + \epsilon] 
        = (Z\T W Z)\inv Z\T W Z\theta
        = \theta.
    \end{equation}

    Its covariance matrix can be computed as
    \begin{equation}
        \label{eq:GLS_variance}
         \cov{\hat \theta}
         = \cov{(Z\T W Z)\inv Z\T W \Upsilon}
        = (Z\T W Z)\inv Z\T W \Lambda W Z (Z\T W Z)\inv. 
    \end{equation}

    We note that all the estimators we are interested in can be written in this form, so they are all unbiased. It remains to compute their variance. 

    Write $\epsilon^n = [\epsilon_1,\dots ,\epsilon_n]\T \in \R^{nT}$. Since $\hat\beta_{u}$ is an OLS estimator with weight $W = I_T$, design $X_u$, observations $Y_u$ and noise $\epsilon_u \sim \mathcal N(0, \Sigma_u)$ we have 
    \begin{align*}
        \E[\hat\beta_u] = \beta, 
        \quad \text{and} \quad 
        \E[\hat\beta]
        = \E\Bracket{\frac 1 n \sum_{u=1}^n \hat \beta_u}
        = \frac 1 n \sum_{u=1}^n \E[\hat\beta_u]
        = \frac 1 n \sum_{u=1}^n \beta
        = \beta. 
    \end{align*}

    Similarly, $\hat\beta_{wu}$ is a weighted OLS estimator with weight $W = \Sigma_u$ and therefore
    \begin{align*}
        \E[\hat\beta_{wu}] = \beta 
        \quad \text{and} \quad 
        \E[\hat\beta]
        = \E\Bracket{\frac 1 n \sum_{u=1}^n \hat \beta_{wu}}
        = \frac 1 n \sum_{u=1}^n \E[\hat\beta_{wu}]
        = \frac 1 n \sum_{u=1}^n \beta
        = \beta. 
    \end{align*}

    We can again apply the general result to obtain the variance of $\hat\beta_u$ and $\hat\beta_{wu}$, 
    \begin{align*}
        \cov{\hat\beta_u}
        &= (X_u\T X_u)\inv X_u\T \Sigma_u X_u (X_u\T X_u)\inv, \\
        \cov{\hat\beta_{wu}}
        &= (X_u\T \Sigma_u \inv X_u)\inv X_u\T \Sigma_u\inv \Sigma_u \Sigma_u \inv X_u (X_u\T \Sigma_u \inv X_u)\inv
        = (X_u\T \Sigma_u \inv X_u)\inv. 
    \end{align*}

    For $u, v \in [n]$, by the independence of $\epsilon_u, \epsilon_v$ due to the block-diagonal structure of $\Sigma^n$ we have $\cov{\hat\beta_u, \hat\beta_v} = 0$. Hence, we can compute the variance of the estimator $\hat\beta$ as 
    \begin{align*}
        \cov{\hat\beta} 
        &= \Cov{\frac 1 n \sum_{u=1}^n \hat \beta_u}
        = \frac{1}{n^2} \sum_{u=1}^n \cov{\hat \beta_u}
        = \frac{1}{n^2} \sum_{u=1}^n (X_u\T X_u)\inv X_u\T \Sigma_u X_u (X_u\T X_u)\inv. 
    \end{align*}

    A similar variance calcualtion for $\hat\beta_w$ leads to
    \begin{align*}
        \cov{\hat\beta_w} 
        &= \Cov{\frac 1 n \sum_{u=1}^n \hat \beta_{wu}}
        = \frac{1}{n^2} \sum_{u=1}^n \Cov{\hat \beta_{wu}}
        = \frac{1}{n^2} \sum_{u=1}^n \Parent{X_u\T \Sigma_u \inv X_u}\inv. 
    \end{align*}

    Invoking the general result for the variance of $\hat\beta_{OLS}$
    \begin{align*}
        \cov{\hat \beta_{OLS}}
        &= \Parent{\sum_{u=1}^n X_u\T X_u}\inv \Parent{\sum_{u=1}^n X_u \T\Sigma_u X_u} \Parent{\sum_{u=1}^n X_u\T X_u}\inv.
    \end{align*}

    With weight $W = \Parent{\Sigma^n}\inv$ the same result yields 
    \begin{align*}
        \cov{\hat\beta_{wOLS}}
        &= \Parent{\sum_{u=1}^n X_u\T \Sigma_u\inv X_u}\inv. 
    \end{align*}

    We are now ready to find the desired variance relations. 
    \begin{enumerate}[(i)]
        \item Generally, for $\Sigma_u \neq \sigma^2I_d$ the Loewner ordering $\cov{\hat\beta_{OLS}} \preceq \cov{\hat\beta}$ does not hold, i.e., 
        \begin{align*}
            \Parent{\sum_{u=1}^n X_u\T X_u}\inv \Parent{\sum_{u=1}^n X_u \T\Sigma_u X_u} \Parent{\sum_{u=1}^n X_u\T X_u}\inv
            \not\preceq \frac{1}{n^2} \sum_{u=1}^n \Parent{X_u\T X_u}\inv X_u\T \Sigma_u X_u \Parent{X_u\T X_u}\inv.
        \end{align*}

        Already when $n = 2, T = 3$ and $p = 2$, the inequality above fails for the following example: 
        \begin{align*}
            X_1 = 
            \begin{pmatrix}
                1 & 0 \\
                0 & 1 \\
                0 & 0
            \end{pmatrix}, \quad 
            X_2 = 
            \begin{pmatrix}
                \sqrt 2 & 0 \\
                0 & \sqrt 2 \\
                0 & 0
            \end{pmatrix}, \quad 
            B_1 = I_3, \quad 
            B_2 = 4\cdot I_3.
        \end{align*}

        We have that $X_1\T X_1 = I_2$, $X_2 \T X_2 = 2 \cdot I_2$, $X_1\T B_1 X_1 = I_2$ and $X_2\T B_2 X_2 = 8 \cdot I_2$. Thus, 
        \begin{align*}
            \cov{\hat\beta_{OLS}}
            &=(X_1\T X_1 + X_2\T X_2)\inv (X_1\T B_1 X_1 + X_2\T B_2 X_2) (X_1\T X_1 + X_2\T X_2)\inv \\
            &= (3 \cdot I_2)\inv (I_2 + 8 \cdot I_2) (3 \cdot I_2)\inv \\
            &= (1/3 \cdot I_2) (9 \cdot I_2) (1/3 \cdot I_2) = I_2. 
        \end{align*}
    
        Moreover, we can explicitly compute 
        \begin{align*}
            \cov{\hat\beta}
            &= 1/2^2 \cdot ((X_1\T X_1)\inv X_1\T B_1 X_1 (X_1\T X_1)\inv + (X_2\T X_2)\inv X_2\T B_2 X_2 (X_2\T X_2)\inv) \\
            &= 1/4 \cdot (I_2 I_2 I_2 + (1/2 \cdot I_2)(8 \cdot I_2) (1/2 \cdot I_2)) \\
            &= 1/4 \cdot (I_2 + 2 \cdot I_2)
            = 3/4 \cdot I_2. 
        \end{align*}

        Overall, we have $I_2 = \cov{\hat\beta_{OLS}} \not\preceq \cov{\hat\beta} = 3/4 \cdot I_2$.

        \item When restricting to the case $\Sigma_u = \sigma^2 I_d$, Jensen's inequality shows that
        \begin{align*}
            \cov{\hat\beta_{OLS}}
            &= \Parent{\sum_{u=1}^n X_u\T X_u}\inv \Parent{\sum_{u=1}^n X_u \T\Sigma_u X_u} \Parent{\sum_{u=1}^n X_u\T X_u}\inv
            = \sigma^2 \Parent{\sum_{u=1}^n X_u\T X_u}\inv \\
            &\preceq
            \frac{1}{n^2} \sum_{i=1}^n \sigma^2 \Parent{X_u\T X_u}\inv
            = \sum_{u=1}^n \Parent{X_u\T X_u}\inv X_u\T \Sigma_u X_u \Parent{X_u\T X_u}\inv
            = \cov{\hat\beta}. 
        \end{align*}
    
        \item 
        When $X_u\T X_u=D \succeq 0$ for all users, we see that
        \begin{align*}
            \cov{\hat \beta_{OLS}}
            = (nD)\inv \Parent{\sum_{u=1}^n X_u \T\Sigma_u X_u} (nD)\inv
            = \frac{1}{n^2} \sum_{u=1}^n D\inv X_u \T\Sigma_u X_u D\inv
            = \cov{\hat \beta}. 
        \end{align*}
         \item Applying the Gauss-Markov Theorem 
        to $\hat\beta_{wu}$ as the weighted OLS and $\hat\beta_u$ as the other linear unbiased estimator, we have $\cov{\hat\beta_{wu}} \preceq \cov{\hat\beta_u}$. Thus, 
        \begin{align*}
            \cov{\hat\beta_w}
            = \frac{1}{n^2} \sum_{u=1}^n \cov{\hat \beta_{wu}} 
            \preceq \frac{1}{n^2} \sum_{u=1}^n \cov{\hat \beta_u}
            = \cov{\hat\beta}. 
        \end{align*}
        Moreover, Jensen's inequality shows that
    
        \begin{align*}
            \cov{\hat\beta_{wOLS}}
            = \Parent{\sum_{u=1}^n X_u\T \Sigma_u\inv X_u}\inv
            \preceq \frac{1}{n^2} \sum_{u=1}^n \Parent{X_u\T \Sigma_u\inv X_u}\inv
            = \cov{\hat\beta_w}.
        \end{align*}
    \end{enumerate}
\end{proof}

\begin{proof}[Proof of Lemma \ref{lem:concuserlvlreg}]\hfill 

    We first prove that $\hat\beta_u$ is $(\tau, \gamma)^\beta_\infty$-concentrated. Second, we derive the concentration of $\hat\beta$. 

    Since $\hat\beta_u$ is a linear map of $Y_u \sim \mathcal N(X_u\beta, \Sigma_u)$, it follows from \eqref{eq:GLS_bias} and \eqref{eq:GLS_variance} that
    $\hat\beta_u \sim \mathcal N(\beta, \Sigma_{\hat\beta_u})$ where $ \Sigma_{\hat\beta_u}=(X_u\T X_u)\inv X_u\T \Sigma_u X_u (X_u\T X_u)\inv$. We now bound the operator norm of $ \Sigma_{\hat\beta_u}$. 
    \begin{align*}
        \opnorm{\Sigma_{\hat\beta_u}}
        &= \opnorm{(X_u\T X_u)\inv X_u\T \Sigma_u X_u (X_u\T X_u)\inv}
        \leq \opnorm{(X_u\T X_u)\inv}^2 \cdot \opnorm{X_u\T \Sigma_u X_u} \\
        &\leq \opnorm{(X_u\T X_u)\inv}^2 \cdot \opnorm{X_u\T X_u} \cdot \opnorm{\Sigma_u}
        \leq \lambda_{\max}(X_u\T X_u)/\lambda^2_{\min}(X_u\T X_u) \cdot {\lambda_{\max}(\Sigma_u)}.
    \end{align*}

    Writing this in terms of the empirical second moment $\hat M_u = \frac 1 T X_u\T X_u$ and using the bounds $\theta \preceq \hat M_u \preceq \vartheta$ and $\Sigma_u \preceq \sigma^2$ we obtain an upper bound on the operator norm
    \begin{align*}
        \opnorm{\Sigma_{\hat\beta_u}}
        \leq \frac 1 T \frac{\lambda_{\max}(\hat M_u)\lambda_{\max}(\Sigma_u)}{\lambda^2_{\min}(\hat M_u)}
        \leq \frac{\vartheta\sigma^2}{T\theta^2}
        =: \rho.
    \end{align*}

    To show $(\tau, \gamma)^\beta_\infty$-concentration we use that $\hat\beta_u$ is Gaussian. Example \ref{exa:gausslogsob} then implies that its distribution is $\LSI{\rho}$. Hence, using the upper bound on the operator norm, $\hat\beta_u\T \in \R^{1\times d}$ fulfills Assumption \ref{ass:logsobdep} with constants $(\rho, \infty)$ and by Lemma \ref{lem:logsobconcmean} we have $(\tau, \gamma)_\infty^\beta$-concentration as
    \begin{align*}
        \P\Parent{\Infnorm{\hat\beta_u - \beta} \leq \sqrt{\frac{2\vartheta\sigma^2\log(2dn/\gamma)}{m\theta^2}}}
        \leq \gamma/n. 
    \end{align*}

    The desired result follows from Lemma \ref{lem:gaussconciidempmean} applied to $\hat\beta_1,\dots ,\hat\beta_n$ combined with $\opnorm{\Sigma_{\hat\beta_u}} \leq \rho$. Concretely, we use that $\frac 1 n \sum_{u=1}^n \Sigma_{\hat\beta_u} \preceq \rho$. On top, we exploit that $\trace(\Sigma_{\hat \beta_u}) \leq p \opnorm{\Sigma_{\hat \beta_u}}$ to obtain
    \begin{align*}
        \P\Parent{\Twonorm{\hat\beta - \beta} \geq \sqrt{\frac{p\vartheta\sigma^2}{nT\theta^2}} + \sqrt{\frac{2\vartheta\sigma^2\log(1/\alpha)}{nT\theta^2}}}
        \leq \P\Parent{\Twonorm{\hat\beta - \beta} \geq \sqrt{\frac{\trace\parent{\Sigma_{\hat\beta_u}}}{n}} + \sqrt{\frac{2\rho\log(1/\alpha)}{n}}}
        \leq \alpha. 
    \end{align*}
\end{proof}

\begin{proof}[Proof of Corollary \ref{cor:MSEuserlvlreg}]\hfill 

    By Lemma \ref{lem:concuserlvlreg} $\hat\beta^n$ is the matrix $(\tau, \gamma)_\infty^\beta$-concentrated with $\tau = \sqrt{2\vartheta\sigma^2\log(2pn/\gamma)/(T\theta^2)}$ and with probability at least $1-\alpha$ it holds that
    \begin{align*}
        \Twonorm{\hat\beta - \beta} \leq \sqrt{\frac{p\vartheta\sigma^2}{nT\theta^2}} + \sqrt{\frac{2\vartheta\sigma^2\log(1/\alpha)}{nT\theta^2}}.
    \end{align*}

    Furthermore, assumption \ref{ass:iiddata} holds for $\hat\beta^n$ as the rows $\hat\beta_u$ are independent of each other due to the block structure of $\Sigma^n$. Thus by the concentration derived above with $\alpha = p\gamma$ combined with Theorem \ref{thm:rMSEwinsmeanHD} via a union bound, with probability at least $1-3p\gamma-O(p^2 n/\delta \cdot e^{-\kappa n \varepsilon^\prime})$, 
    \begin{align*}
        \Twonorm{\mathcal A(\hat\beta^n) - \beta} 
        &\lesssim \Twonorm{\hat \beta - \beta} + \tau \sqrt{\frac{p^2\log(1/\varrho)\log(3/(p\gamma))^2}{n^2\varepsilon^2}} \\
        &\lesssim \sqrt{\frac{p\vartheta\sigma^2}{nT\theta^2}} + \sqrt{\frac{\vartheta\sigma^2\log(1/(p\gamma))}{nT\theta^2}} + \sqrt{\frac{p^2\vartheta\sigma^2\log(2dn/\gamma)\log(1/\varrho)\log(3/(p\gamma))^2}{\theta^2Tn^2\varepsilon^2}}.
    \end{align*}
\end{proof}

\section{Proofs of Section \ref{sec:LDP}}

\subsection{Finding Private Data-Driven Projection Intervals (Local)}
\label{app:mean_LDP}

\begin{algorithm}
    \caption{ProjectionInterval$(X^n,\tau,\varepsilon,\delta, \mathsf B)$:}
    \label{alg:privrangerandhist}
    \begin{algorithmic}[1]
        \REQUIRE $X^n \in \R^n$, $\tau, \varepsilon > 0, \delta = 0$
        \STATE $S \gets \curly{k \in \Z : \abs{2\tau k} \leq \mathsf B}$
        \STATE $B_k^+ \gets (2\tau k \pm \tau]$ \hfill $\forall k \in S$
        \STATE $\boldsymbol{\tilde p} \gets \operatorname{RandomizedHistogram}(X, (B_k^+)_{k \in S}, \varepsilon/2, 0)$
        \STATE $\hat k \gets \arg\max_{k \in S} \tilde p_k$
        \STATE $\hat m \gets 2\tau \hat k$
        \RETURN $[\hat m \pm 3\tau]$
    \end{algorithmic}
\end{algorithm}

Our main application of the randomized histogram is to compute private data-driven projection intervals that we use in our local Winsorized mean estimator. Our approach hereby resembles the one in the central model. In Algorithm \ref{alg:privrangerandhist} we again choose the projection interval as the bin of the histogram that has the highest estimated mass and its two neighboring bins. Because the randomized histogram requires finitely many bins as input, we cannot search for the bin with highest mass over the entirety of $\R$ anymore. Instead we restrict ourselves to the interval $\bracket{- \mathsf B, \mathsf B} \subset \R$. For mean estimation, this will impose the assumption that the mean $\mu \in \R^d$ to be estimated fulfills $\infnorm{\mu} \leq \mathsf B$. This boundedness assumption relaxes boundedness assumptions on the single observations $X_i$ as they are made, e.g., by \citet{kent:berrett:yu2024}. Yet, it is more restrictive than the boundedness we assume for our in-expectation bound in the central model, because the constant $\mathsf B$ actually has to be known for the construction of the algorithm and not just during its analysis in expectation. Hence, this peculiarity of the private projection interval estimator shows up in the finite sample MSE-risk bounds for our local estimators. These build on top of the estimation guarantee for the center of the projection interval that is given in Lemma \ref{lem:privmidpointrandhist} below whose proof we defer for now. 

\begin{lemma}
    \label{lem:privmidpointrandhist}
    Let $\varepsilon > 0$, $\gamma \in (0, 1 \wedge \frac n 4)$. Suppose $X^n \in \R^n$ is $(\tau, \gamma)_\infty$-concentrated around $x_0 \in [-\mathsf B, \mathsf B] \subset \R$ and fulfills Assumption \ref{ass:iiddata} or Assumption \ref{ass:logsobdep} with $\rho M^2 \lesssim 1$. Then, Algorithm \ref{alg:privrangerandhist} is $(\varepsilon, 0)$-DP and with probability at least $1-O\parent{\mathsf B/\tau + 1} \cdot e^{-\kappa n \tanh^2(\varepsilon/4)} \vee e^{-\kappa^\prime n}$ with $\kappa, \kappa^\prime > 0$: 
    \begin{align*}
        \hat m \in \Big[x_0 \pm 2\tau \Big]. 
    \end{align*}
\end{lemma}

\subsection{Proofs for Randomized Histogram}
\label{app:hist_LDP}

\begin{proof}[Proof of Lemma \ref{lem:randhist}] \hfill 

    The randomized histogram is private by the reasoning in Lemma \ref{lem:privlossrandhist}. 

    For the proof of utility, define $\hat p_k := \frac 1 n \sum_{i=1}^n \indicator{X_i \in B_k}$. An application of the triangle inequality $\abs{\tilde p_k - p_k} \leq \abs{\tilde p_k - \hat p_k} + \abs{\hat p_k - p_k}$ and the total law of probability yield: 
    \begin{align*}
        \P\Parent{\max_{k \in \Z} \abs{\tilde p_k - p_k} \geq \eta}
        &\leq \E\Bracket{\P\Parent{\max_{k \in \Z} \abs{\tilde p_k - \hat p_k} \geq \frac \eta 2\Big|X^n}} + \P\Parent{\max_{k \in \Z} \abs{\hat p_k - p_k} \geq \frac \eta 2}. 
    \end{align*}

    The above isolates the privacy loss of the randomized histogram due to conditioning on $X^n$. This makes Lemma \ref{lem:privlossrandhist} applicable and we obtain
    \begin{align*}
        \E\Bracket{\P\Parent{\max_{k \in \Z} \abs{\tilde p_k - \hat p_k} \geq \frac \eta 2 \Big | X^n}}
        \leq 2 N_{\mathsf{bins}} \cdot \exp\Parent{-\frac{n\eta^2}{4} \tanh^2\Parent{\frac{\varepsilon}{4}}}. 
    \end{align*}

    The estimation error can now be bounded using Lemma \ref{lem:estlossemphist}, which gets us 
    \begin{align*}
        \P\Parent{\max_{k \in \Z} |\hat p_k - p_k| \geq \frac \eta 2}
        \leq 
        \begin{cases}
            2 \exp\Parent{-\frac{n\eta^2}{8}} & \text{if $X_1,\dots ,X_n$ are \iid{}}, \\
            \frac{16}{\eta} \exp\Parent{-\frac{2}{27} \frac{n\eta^3}{4^3 \rho M^2}} & \text{if $X$ fulfills Assumption \ref{ass:logsobdep}}.
        \end{cases}
    \end{align*}

    The statement is then recovered by collecting terms. 
\end{proof}

\begin{lemma}
    \label{lem:privlossrandhist}
    Let $\eta, \varepsilon > 0$. Moreover, let $(B_k)_{k \in [N_{\mathsf{bins}}]}$ be disjoint bins covering $[-\mathsf B,\mathsf B] \subset \R$. Then, on input of a fixed vector $x \in \R^n$, Algorithm \ref{alg:randhist} is $(\varepsilon, 0)$-DP and its output $\boldsymbol{\tilde p}$ fulfills: 
    \begin{align*}
        \P\Parent{\max_{k \in \Z} \abs{\tilde p_k - \hat p_k} \geq \eta}
        \leq 2N_{\mathsf{bins}} \cdot \exp\Parent{-n\eta^2 \tanh^2\Parent{\frac{\varepsilon}{4}}}. 
    \end{align*}
    
\end{lemma}

\begin{proof}

    The privacy follows from that of the randomized response mechanism. Concretely, for $x, x^\prime \in \R^n$ \st{} $d(x,x^\prime) \leq 1$ at most two bins can be affected. Basic composition and the $(\varepsilon, 0)$-\DP{} guarantee of Algorithm \ref{alg:randresponse} in Theorem \ref{thm:randresponse} yield the claimed privacy. The debiasing step does not affect the privacy guarantee by post-processing. 

    For utility, we first show that the $\tilde p_k^{\operatorname{debiased}}$ are indeed unbiased for $\hat p_k$. For this, let $\pi$ be the probability that the randomization keeps the data unchanged in Algorithm \ref{alg:randresponse}. Then, 
    \begin{align*}
        \E\Bracket{\tilde p_k^{\operatorname{debiased}}}
        = \E\Bracket{\frac{\tilde p_k - (1-\pi)}{2\pi-1}}
         =\frac{\E\Bracket{\tilde p_k} - (1-\pi)}{2\pi-1}
        \overset{(a)}{=} \frac{(2\pi-1) \hat p_k + (1-\pi) - (1-\pi))}{2\pi-1}
        = \hat p_k. 
    \end{align*}

    Above, equality $(a)$ holds by linearity of the expectation since
    \begin{align*}
        \E\Bracket{\tilde p_k}
        = \E\Bracket{\frac 1 n \sum_{i=1}^n \tilde c_{ik}}
        &= \E\Bracket{\frac 1 n \sum_{i=1}^n c_{ik} \cdot \indicator{U_i \leq \pi} + (1-c_{ik}) \cdot \indicator{U_i > \pi}} \\
        &= \frac 1 n \sum_{i=1}^n\Big[ c_{ik} \cdot \P\Parent{U_i \leq \pi} + (1-c_{ik}) \cdot \P\Parent{U_i > \pi}\Big] \\
        &= \frac 1 n \sum_{i=1}^n \Big[c_{ik} \cdot \pi + (1-c_{ik}) \cdot (1-\pi)\Big] \\
        &= \frac 1 n \sum_{i=1}^n c_{ik} (2 \pi-1) + (1 - \pi) \\
        &= (2\pi-1) \hat p_k + (1-\pi).
    \end{align*}

    Note also that 
    \begin{align*}
        \tilde p_k^{\operatorname{debiased}} - \hat p_k
        = \tilde p_k^{\operatorname{debiased}} - \E\Bracket{\tilde p_k^{\operatorname{debiased}}}
        = \frac{\tilde p_k - (1-\pi))}{2\pi-1} - \frac{\E\Bracket{\tilde p_k} - (1-\pi))}{2\pi-1} 
        = \frac{\tilde p_k - \E[\tilde p_k]}{2\pi-1}. 
    \end{align*}
    Here, the numerator is a sum of independent terms $\tilde c_{ik} \in [0,1]$ minus their expectation since 
    \begin{align*}
        \tilde p_k - \E[\tilde p_k]
        = \frac 1 n \sum_{i=1}^n \tilde c_{ik} - \E\Bracket{\tilde c_{ik}}
        = \frac 1 n \sum_{i=1}^n c_{ik} \cdot \indicator{U_i \leq \pi} + (1-c_{ik}) \cdot \indicator{U_i > \pi} - \E[\tilde c_{ik}].
    \end{align*}

    As the $U_i$ and the $\tilde c_{ik}$ are \iid{}, an application of Hoeffding's inequality yields
    \begin{align*}
        \P\Parent{\Abs{\tilde p_k^{\operatorname{debiased}} - \hat p_k} \geq \eta}
        = \P\Parent{\Abs{\frac{\tilde p_k - \E[\tilde p_k]}{2\pi - 1}} \geq \eta}
        =\P\Parent{\Abs{\frac{\sum_{i=1}^n \tilde c_{ik} - \E[\tilde c_{ik}]}{2\pi-1}} \geq n\eta} \leq 2e^{-n\eta^2(2\pi-1)^2}. 
    \end{align*}

    Using that $(2\pi-1)^2 = \tanh^2(\varepsilon/4)$ and union bounding over the $N_{\mathsf{bins}}$ bins recovers the statement. 
\end{proof}

\subsection{Proofs for Private Projection Interval (Local)}

\begin{proof}[Proof of Lemma \ref{lem:privmidpointrandhist}] \hfill 

    The privacy of Algorithm \ref{alg:privrangerandhist} follows directly from that of Algorithm \ref{alg:randhist} shown in Lemma \ref{lem:privlossrandhist}. 

    Algorithm \ref{alg:randhist} uses bins $B_k := (2\tau k \pm \tau]$ of width $2\tau$ to cover the interval $[- \mathsf B, \mathsf B]$. Hence, the collection of bins $(B_k)_{k \in S}$ used is of size $N_{\mathsf{bins}} \leq \lceil \mathsf B/\tau \rceil + 1 \leq \mathsf B/\tau + 2$. With this, under Assumption \ref{ass:logsobdep} Lemma \ref{lem:randhist} yields the following guarantee: 
    \begin{align*}
        \P\Parent{\max_{k \in \Z} \abs{\tilde p_k - p_k} \geq \frac{1}{16}} 
        &\leq \Parent{\frac{\mathsf B}{\tau} + 2} 2\exp\Parent{-\frac{n}{4\cdot16} \tanh^2\Parent{\frac{\varepsilon}{4}}} - 16^2 \exp\Parent{-\frac{1}{864} \frac{n}{16^3\rho M^2}} \\
        &\lesssim \Parent{\frac{\mathsf B}{\tau}+1} e^{-\kappa n \tanh^2\Parent{\frac \varepsilon 4}} \vee e^{-\kappa^\prime n}. 
    \end{align*}

    Above, we suppress the constant $\rho M^2 \lesssim 1$. Given that $X^n$ is $(\tau, \gamma)_\infty^{x_0}$-concentrated, Lemma \ref{lem:midpoint} translates this guarantee into one for the midpoint $\hat m$: 
    \begin{align*}
        \P\Parent{\hat m \in \Big[x_0 \pm 2\tau \Big]}
        \geq 1 - O\Parent{\Parent{\frac{\mathsf B}{\tau}+1} e^{-\kappa n \tanh^2\Parent{\frac \varepsilon 4}} \vee e^{-\kappa^\prime n}}. 
    \end{align*}

    The proof is analogous and instead uses Lemma \ref{lem:randhist} under Assumption \ref{ass:iiddata}. 
\end{proof}

\subsection{Proofs for Mean Estimators (Local)}

\begin{lemma}
    \label{lem:winsmean1Dlocal}
    Algorithm \ref{alg:winsmean1Dlocal} denoted by $\mathcal A$ is $(\varepsilon, 0)$-\DP{} with $\varepsilon > 0$. Assume $X^n \in \R^n$ is $(\tau, \gamma)^{x_0}_\infty$-concentrated with $\infnorm{x_0} \leq \mathsf B$ and fulfills Assumption \ref{ass:iiddata} or \ref{ass:logsobdep} \st{} $\rho M^2 \lesssim 1$. Then, the event $\mathcal E := \curly{\forall i \in [n] : \proj{\hat I}{X_i} = X_i}$ has probability at least $1 - \gamma - O\parent{\mathsf B/\tau + 1} \cdot e^{-\kappa n \tanh^2(\varepsilon/4)} \vee e^{-\kappa^\prime n}$ and 
    \begin{align*}
        \mathcal A(X^n) \cdot \indicator{\mathcal E} = \Parent{\bar X_n + \bar \xi_n} \cdot \indicator{\mathcal E}, \quad \text{almost surely.}
    \end{align*}
\end{lemma}

\begin{proof}

    The private midpoint estimator of Algorithm \ref{alg:privrangestabhist} is $(\frac \varepsilon 2, 0)$-DP by Lemma \ref{lem:privmidpointrandhist} as we call it with $\frac \varepsilon 2$ in Algorithm \ref{alg:winsmean1Dlocal}. Further, the Laplace mechanism therein is $(\frac{\varepsilon}{2},0)$-DP by Theorem \ref{thm:lapmech}. Therefore, algorithm $\mathcal A$ is $(\varepsilon, 0)$-DP through basic composition in Theorem \ref{thm:compthms}. 

    For utility, by Lemma \ref{lem:privmidpointrandhist} under Assumption \ref{ass:iiddata} or \ref{ass:logsobdep} we have 
    \begin{align*}
        \P\Parent{\hat m \in [x_0 \pm 2\tau]}
        \geq 1-O\Parent{\mathsf B/\tau + 1} \cdot e^{-\kappa n \tanh^2(\varepsilon/4)} \vee e^{-\kappa^\prime n}.
    \end{align*}
    
    Hence, by an application of Lemma \ref{lem:winsmean1Dgeneral} with $C=2$, the event $\mathcal E := \curly{\forall i \in [n] : \proj{[\hat m \pm 3\tau]}{X_i} = X_i}$ has probability at least $1 - \gamma - O\parent{\mathsf B/\tau + 1} \cdot e^{-\kappa n \tanh^2(\varepsilon/4)} \vee e^{-\kappa^\prime n}$ and 
    \begin{align*}
        \mathcal A(X^n) \cdot \indicator{\mathcal E} 
        = \Parent{\bar X_n + \bar \xi_n} \cdot \indicator{\mathcal E}, 
        \quad \text{almost surely}. 
    \end{align*}
\end{proof}

\begin{remark}
    If we assume $\varepsilon \in (0,1)$ we can make the bound above exponential in the quadratic of $\varepsilon$ instead of the hyperbolic tangent. This holds as $\tanh(1/2) x \leq \tanh(x/2)$ for $x \in [0,1]$ and thus, 
    \begin{align*}
        \tanh^2\Parent{\frac \varepsilon 4}
        \leq -\tanh^2\Parent{\frac 1 4}\varepsilon^2
        =: -\kappa^{\prime \prime} \varepsilon^2. 
    \end{align*}

    As a result, for $\varepsilon \in (0,1)$ and $\mathcal E$ as above, with probability $1 - \gamma - O(\mathsf B/\tau \cdot e^{-\kappa^{\prime\prime\prime} n \varepsilon^2})$ for $\kappa^{\prime\prime\prime} > 0$, 
    \begin{align*}
        \mathcal A(X^n) \cdot \indicator{\mathcal E} 
        = \Parent{\bar X_n + \bar \xi_n} \cdot \indicator{\mathcal E}, 
        \quad \text{almost surely}. 
    \end{align*}
\end{remark}

\begin{lemma}
    \label{lem:winsmeanHDlocal}
    Let $\varepsilon, \varrho \in (0,1)$ and $\gamma \in (0, 1 \wedge \frac n 4)$. Algorithm \ref{alg:winsmeanHD} denoted by $\mathcal A$ calling Algorithm \ref{alg:winsmean1Dlocal} in Line 3 is $(\varepsilon, \varrho)$-\LDP{}. Suppose $X^n \in \R^{n \times d}$ is $(\tau, \gamma)^{x_0}_\infty$-concentrated with $\infnorm{x_0} \leq \mathsf B$. Further, let $X^n$ fulfill Assumption \ref{ass:iiddata} or \ref{ass:logsobdep} \st{} $\rho M^2 \lesssim 1$. Then, the event $\mathcal E := \curly{\forall i \in [n], j \in [d] : \proj{R_j}{X_{ij}} = X_{ij}}$ has probability at least $1 - d\gamma - O\parent{\mathsf B/\tau +1} \cdot de^{-\kappa n (\varepsilon^\prime)^2}$ with $\varepsilon'=\varepsilon/\sqrt{8d\log(1/\varrho)}$. Moreover, a.s.\,
    \begin{align*}
        \mathcal A(X^n) \cdot \indicator{\mathcal E}
        = \Parent{\bar X_n + \bar \Xi_n} \cdot \indicator{\mathcal E}, \quad \text{where} \quad \Xi_i \sim \operatorname{Lap}\Parent{0, \frac{12\tau}{\varepsilon^\prime}I_d}.
    \end{align*}
\end{lemma}

\begin{proof}

    Since the $d$ one-dimensional mean estimators fulfill $(\varepsilon, 0)$-DP, the coordinate-wise multi-dimensional estimator is private by composition. We use the privacy parameters $\varepsilon^\prime, \delta^\prime = 0, \varrho$ defined in Algorithm \ref{lem:winsmeanHD} governed by advanced composition in Theorem \ref{thm:compthms}. This introduces $\varrho$.

    $X^n$ is $(\tau,\gamma)_\infty^{x_0}$-concentrated. Applying Algorithm \ref{alg:winsmean1Dlocal} denoted by $\mathcal A_1$ with $(X^n_{\cdot j}, \tau, \varepsilon^\prime, \delta^\prime)$ per dimension, by Lemma \ref{lem:winsmean1Dlocal}, $\mathcal E_j := \curly{\forall i \in [n] : \proj{\hat I}{X_i} = X_i}$ has $\P(\mathcal E_j) \geq 1 - \gamma - O\parent{\mathsf B/\tau + 1} \cdot e^{-\kappa n \tanh^2(\varepsilon)}$ and 
    \begin{align*}
        \mathcal A_1(X_{\cdot j}) \cdot \indicator{\mathcal E} = \Parent{\frac 1 n \sum_{i=1}^n X_{ij} + \xi_{ij}} \cdot \indicator{\mathcal E}, \quad \text{almost surely.}
    \end{align*}

    Using the remark above and by Lemma \ref{lem:winsmeanHDgeneral} with $h=n$ and $b = \frac{12\tau}{\varepsilon^\prime}$, on the event $\mathcal E = \bigcap_{j \in [d]} \mathcal E_j$ that has probability at least $1-d\gamma-O\parent{\mathsf B/\tau + 1} \cdot de^{-\kappa n (\varepsilon^\prime)^2}$ it holds that, almost surely
    \begin{align*}
        \mathcal A(X) \cdot \indicator{\mathcal E}
        = \Parent{\frac 1 n \sum_{i=1}^n X_{i\cdot}\T + \Xi_{i}} \cdot \indicator{\mathcal E}, 
        \quad \text{with $\Xi_i = (\xi_{i1},..., \xi_{id})\T$}.
    \end{align*}

    Note that by definition it holds that $\bar X_n + \bar \Xi_n = \frac 1 n \sum_{i=1}^n X_{i\cdot}\T + \Xi_{i}$. The statement is then recovered as the the $\Xi_{ij}$ are \iid{} and therefore $\Xi_i \sim \operatorname{Lap}(0, \frac{12\tau}{\varepsilon^\prime})$. 
\end{proof}

\subsection{Proofs of Theoretical Guarantees (Local)}
\label{app:theorems_mean_LDP}

\begin{proof}[Proof of Theorem \ref{thm:rMSEwinsmeanHDlocal}] \hfill 

    This proof directly follows from Lemma \ref{lem:concwinsmeanhd}. By Lemma \ref{lem:winsmeanHDlocal} there is an event $\mathcal E$ that has probability at least $1-\gamma-O\parent{\mathsf B/\tau +1} \cdot de^{-\kappa n (\varepsilon^\prime)^2}$ \st{} almost surely, 
    \begin{align*}
        \mathcal A(X^n) \cdot \indicator{\mathcal{E}}
        = \Parent{\bar X_n + \bar\Xi_n} \cdot \indicator{\mathcal{E}}, \quad \text{where $\Xi \sim \operatorname{Lap}\Parent{0, \frac{12\tau}{\varepsilon^\prime}I_d}$}.
    \end{align*}

    Applying Lemma \ref{lem:concwinsmeanhd} with $h=n$, with probability at least $1-\alpha-d\gamma-O\parent{\mathsf B/\tau +1} \cdot de^{-\kappa n (\varepsilon^\prime)^2}$, 
    \begin{align*}
        \Twonorm{\mathcal A(X) - \mu} 
        &< \Twonorm{\bar X_n - \mu} + \frac{12\tau}{\varepsilon^\prime} \Parent{\sqrt{\frac{2d}{n}} + \sqrt{\frac{4\log(3/\alpha)^2}{n}}} \\
        &\leq \Twonorm{\bar X_n - \mu} + 12\tau \Parent{\sqrt{\frac{16d^2\log(1/\varrho)}{n\varepsilon^2}} + \sqrt{\frac{32d\log(1/\varrho)\log(3/\alpha)^2}{n\varepsilon^2}}}. 
    \end{align*}

    The statement follows by omitting absolute constants in $\lesssim$-notation and bounding
    \begin{align*}
        \sqrt{16d^2\log(1/\varrho)} + \sqrt{32d\log(1/\varrho)\log(3/\alpha)^2}
        \lesssim \sqrt{d^2\log(1/\varrho)\log(3/\alpha)^2}. 
    \end{align*}

    Finally, the statement is obtained by setting $\alpha = d\gamma$. 
\end{proof}

\begin{proof}[Proof of Theorem \ref{thm:inexpublocal}] \hfill 

    Define the noiseless estimator $T_0 := \frac 1 n \sum_{i=1}^n \proj{\hat I^d}{X_i}$ where $\proj{\hat I^d}{X_i} := [\Pi_{\hat I_1}(X_{i1}), \dots, \Pi_{\hat I_1}(X_{ij})]\T$ stems from the coordinate-wise application of Algorithm \ref{alg:winsmean1Dlocal}. Then, by independence of $\bar \Xi_n$ and $T_0$
    \begin{align*}
        \E\Bracket{\Twonorm{\mathcal A(X^n) - \mu}^2}
        = \E\Bracket{\Twonorm{T_0 + \bar \Xi_n - \mu}^2}
       =\E\Bracket{\Twonorm{\bar \Xi_n}^2} + \E\Bracket{\Twonorm{T_0 - \mu}^2}. 
    \end{align*}

    Here, $\E[\twonorm{\bar \Xi_n}^2] \leq 2db^2/n$ as $\bar \Xi_n$ is an average of $\Xi_1,\dots,\Xi_n \overset{iid}{\sim} \operatorname{Lap}(0, b I_d)$. We now bound the second term. For this, define the event on which the projections in Algorithm \ref{alg:winsmeanHD} don't have an effect: 
    \begin{align*}
        \mathcal E 
        := \curly{\forall i \in [n], j \in [d] : \proj{\hat I_j}{X_{ij}} = X_{ij}}. 
    \end{align*}

    By Lemma \ref{lem:winsmeanHDlocal} we have then have $\P\Parent{T_0 = \bar X_n} \geq \P\Parent{\mathcal E} \geq 1 - d\gamma - O\parent{\mathsf B/\tau + 1} \cdot de^{-\kappa n (\varepsilon^\prime)^2}$. Thus, on $\mathcal E$ it holds that $T_0 = \bar X_n$ and we may split the expectation as follows
    \begin{align*}
        \E\Bracket{\Twonorm{T_0 - \mu}^2}
        &= \E\Bracket{\Twonorm{T_0 - \mu}^2 \cdot \indicator{\mathcal E}} + \E\Bracket{\Twonorm{T_0 - \mu}^2 \cdot \indicator{\mathcal E^c}} \\
        &\leq \E\Bracket{\Twonorm{\bar X_n - \mu}^2} + 2\E\Bracket{\Parent{\twonorm{T_0}^2 + \twonorm{\mu}^2} \cdot \indicator{\mathcal E^c}}.
    \end{align*}

    By Assumption we have $\twonorm{\mu}^2 \leq d\infnorm{\mu}^2 \leq d\mathsf B^2$. Further, by Jensen's inequality
    \begin{align*}
        \Twonorm{T_0}^2 
        \leq \sum_{j=1}^d \frac 1 n \sum_{i=1}^n \proj{R_j}{X_{ij}}^2
        \leq \sum_{j=1}^d \frac 1 n \sum_{i=1}^n (\hat m_j + \tau)^2
        \leq 2d (\mathsf B^2 + \tau^2). 
    \end{align*}

    Combining both now yields the following upper bound
    \begin{align*}
        \E\Bracket{\Parent{\twonorm{T_0}^2 + \twonorm{\mu}^2} \cdot \indicator{\mathcal E^c}}
        &\lesssim d(\mathsf B^2 + \tau^2) \cdot \P\Parent{\mathcal E^c} \\
        &\lesssim d(\mathsf B^2 + \tau^2) \cdot \Parent{d\gamma + \Parent{\frac{\mathsf B}{\tau}+1} \cdot de^{-\kappa n (\varepsilon^\prime)^2}} \\
        &\lesssim d^2\gamma(\mathsf B^2 + \tau^2) + \frac{d^2}{e^{\kappa n (\varepsilon^\prime)^2}}\Parent{\frac{\mathsf B^3}{\tau} + \mathsf B^2 + \mathsf B\tau + \tau^2}. 
    \end{align*}

    Collecting terms and plugging in $b \asymp \tau/(\varepsilon^\prime)$ we then obtain
    \begin{align*}
        \E\Bracket{\Twonorm{\mathcal A(X^n) - \mu}^2}
        &\leq \E\Bracket{\Twonorm{\bar \Xi_n}^2} + \E\Bracket{\Twonorm{T_0 - \mu}^2} \\
        & \lesssim \E\Bracket{\Twonorm{\bar X_n - \mu}^2} + \frac{d\tau^2}{n(\varepsilon^\prime)^2} + d^2(\mathsf B^2 + \tau^2) \cdot \Parent{\gamma + \frac{\mathsf B/\tau+1}{e^{\kappa n (\varepsilon^\prime)^2}}}. 
    \end{align*}

    The statement follows as $\varepsilon^\prime \asymp \varepsilon / \sqrt{d\log(1/\varrho)}$ and considering the cases $\mathsf B \geq \tau$ and $\mathsf B \leq \tau$. 
\end{proof}

\begin{proof}[Proof of Corollary \ref{cor:inexpubuserlvllocal}] \hfill 

    Given that Assumption \ref{ass:logsobdep} holds, by Corollary \ref{cor:varlogsobdes} we have $\E\bracket{\twonorm{\bar X_{nT} - \mu}^2} \leq d\rho / (nT)$. Combining this with Theorem \ref{thm:inexpublocal} we thus obtain
    \begin{align*}
        \E[\twonorm{\mathcal A(X^n) - \mu}^2]
        & \lesssim \E\Bracket{\Twonorm{\bar X_n - \mu}^2} + d^2 \Parent{\frac{\tau^2\log(1/\varrho)}{n\varepsilon^2} + (\mathsf B^2 + \tau^2) \cdot \Parent{\gamma + \frac{\mathsf B/\tau+1}{e^{\kappa n (\varepsilon^\prime)^2}}}} \\
        & \leq \frac{d\rho}{nT} + \frac{d^2\tau^2\log(1/\varrho)}{n\varepsilon^2} + d^2(\mathsf B^2 + \tau^2) \cdot \Parent{\gamma + \frac{\mathsf B/\tau+1}{e^{\kappa n (\varepsilon^\prime)^2}}}
        =: T_1 + T_2 + T_3. 
    \end{align*}

    Again using Assumption \ref{ass:logsobdep}, $X^n$ is $(\tau, \gamma)^\mu_\infty$-concentrated with $\tau = \sqrt{2\rho \log(2dn/\gamma)/T}$ by Lemma \ref{lem:logsobconcmean}. We specify $\gamma \asymp (Tn\varepsilon^2)\inv$ which yields $\tau^2 \asymp \rho \log(2dTn^2\varepsilon^2)/T$. The term $T_2$ then becomes
    \begin{align*}
        T_2
        = \frac{d^2\tau^2\log(1/\varrho)}{n\varepsilon^2}
        = \frac{d^2\rho\log(2dTn^2\varepsilon^2)\log(1/\varrho)}{Tn\varepsilon^2}. 
    \end{align*}

    Note that $T_1$ is always dominated by $T_2$ since $\varepsilon \lesssim 1$ by assumption. For the third term $T_3$ we get
    \begin{align*}
        T_3
        &= d^2(\mathsf B^2 + \tau^2) \cdot \Parent{\gamma + \frac{\mathsf B/\tau+1}{e^{\kappa n (\varepsilon^\prime)^2}}} \\
        &= d^2\Parent{\mathsf B^2 + \frac{\rho\log(2dTn^2\varepsilon^2)}{T}} \cdot \Parent{\frac{1}{Tn\varepsilon^2} + \Parent{\sqrt{\frac{\mathsf B^2T}{\rho\log(2dTn^2\varepsilon^2)}} + 1}\frac{1}{e^{\kappa n (\varepsilon^\prime)^2}}}. 
    \end{align*}

    In the setting when $\mathsf B \geq \tau$ and hence $\mathsf B/\tau \geq 1$ , collecting terms yields
    \begin{align*}
        \E[\twonorm{\mathcal A(X^n) - \mu}^2]
        & \lesssim \frac{d^2\rho\log(2dTn^2\varepsilon^2)\log(1/\varrho)}{Tn\varepsilon^2} + d^2\mathsf B^2 \cdot \Parent{\frac{1}{Tn\varepsilon^2} + \sqrt{\frac{\mathsf B^2T}{\rho\log(2dTn^2\varepsilon^2)}}\frac{1}{e^{\kappa n (\varepsilon^\prime)^2}}} \\
        & \lesssim \frac{d^2\rho\log(2dTn^2\varepsilon^2)\log(1/\varrho)}{Tn\varepsilon^2} + d^2\mathsf B^2 \cdot \Parent{\frac{1}{Tn\varepsilon^2} + \frac{\mathsf B}{\sqrt \rho}\frac{\sqrt T}{e^{\kappa n (\varepsilon^\prime)^2}}}. 
    \end{align*}

    Instead if $\mathsf B \leq \tau$ it holds that $\mathsf B/\tau \leq 1$ and therefore
    \begin{align*}
        \E[\twonorm{\mathcal A(X^n) - \mu}^2]
        & \lesssim \frac{d^2\rho\log(2dTn^2\varepsilon^2)\log(1/\varrho)}{Tn\varepsilon^2} + \frac{d^2\rho\log(2dTn^2\varepsilon^2)}{T} \cdot \Parent{\frac{1}{Tn\varepsilon^2} + \frac{1}{e^{\kappa n (\varepsilon^\prime)^2}}} \\
        & \lesssim \frac{d^2\rho\log(2dTn^2\varepsilon^2)}{T} \cdot \Parent{\frac{\log(1/\varrho)}{n\varepsilon^2} + \frac{1}{e^{\kappa n (\varepsilon^\prime)^2}}} \\
        & \lesssim \frac{d^2\rho\log(2dTn^2\varepsilon^2)\log(1/\varrho)}{Tn\varepsilon^2}.
    \end{align*}
    
    Under Assumption \ref{ass:iiddata} we exploit that the rows of $X^n$ are \iid{} with $\rho$-sub-Gaussian entries. This yields $\E[\twonorm{\bar X_n - \mu}^2] \leq d\rho /n$ and $(\tau, \gamma)_\infty^\mu$-concentration with the same $\tau$. The rest is analogous.
\end{proof}

\subsection{Proofs for Nonparametric Regression (Local)}
\label{app:nonparametricslocal}

\begin{proof}[Proof of Corollary \ref{cor:rMSEpriestchaoestlocal}] \hfill

    This proof is analogous to the proof of Corollary \ref{cor:rMSEpriestchaoest} covering the central model. 
    
    We want to combine the concentration results of Lemma \ref{lem:concpriestchaoest} with Theorem \ref{thm:rMSEwinsmeanHDlocal} for $d=1$. Note $\hat f^n(x)$ is concentrated around $f(x)$ and $\sup_{x \in [0,1]}\abs{f(x)} \leq \infnorm{f}$ and thus the theorem's boundedness condition holds. Further, by Lemma \ref{lem:logsobdeppriestchaoest} we may call the theorem under Assumption \ref{ass:logsobdep}. 

    In Lemma \ref{lem:concpriestchaoest} we derived $(\tau, \gamma)_\infty$-concentration of $\hat f^n(x)$ around $f(x)$ with
    \begin{align*}
        \tau 
        &\asymp b + \frac{1}{b} \Parent{1 + \sqrt{\sigma^2_{\max} \infnorm{K} \log(2n/\gamma)}}
        \lesssim \frac{1}{b} \sqrt{\sigma^2_{\max} \infnorm{K} \log(2n/\gamma)}
        =: \frac{C}{b}.
    \end{align*}
    
    There, we also derived a statistical rate for the rMSE, meaning that with probability at least $1-\alpha$ 
    \begin{align*}
        \Abs{\hat f_n(x) - f(x)}
        &\lesssim 
        b + \frac{1}{nb} + 
        \sqrt{\frac{\sigma_{\max}^2 \infnorm{K} \log(2/\alpha)}{nb}}. 
    \end{align*}

    Setting $\alpha = \gamma$ and combining this with Theorem \ref{thm:rMSEwinsmeanHDlocal} via a union bound, with probability at least $1-3\gamma-O(M/\tau \cdot e^{-\kappa n \varepsilon^2})$ we have 
    \begin{align*}
        \Abs{\mathcal A(\hat f^n(x)) - f(x)} 
        &\lesssim \Abs{\hat f_n(x) - f(x)} + \tau \sqrt{\frac{\log(1/\varrho)\log(3/\gamma)^2}{n^2\varepsilon^2}} \\
        &\lesssim \Abs{\hat f_n(x) - f(x)} + \sqrt{\frac{C^2\log(1/\varrho)\log(3/\gamma)^2}{b^2n\varepsilon^2}} \\
        &\lesssim b + \frac{1}{nb} + \sqrt{\frac{\sigma_{\max}^2 \infnorm{K} \log(2/\gamma)}{nb}} + \sqrt{\frac{C^2\log(1/\varrho)\log(3/\gamma)^2}{nb^2\varepsilon^2}}.
    \end{align*}

    Note that by the expression for $\tau$ above we have $\tau \gtrsim b$, which lets us bound the big-O in the probability. Modulo logarithms, and using $1\leq 1/b$, $1\lesssim 1/\varepsilon$, $1/n \leq 1$ and $1 \lesssim \sigma^2_{\max}$ to merge the second and fourth term yields
    \begin{align*}
        \Abs{\mathcal A(\hat f^n(x)) - f(x)} 
        &\lessapprox b + \frac{1}{nb} + \sigma_{\max} \Parent{\sqrt{\frac{1}{nb}} + \sqrt{\frac{1}{n^2b^2\varepsilon^2}}}
        \lesssim b + \sigma_{\max} \Parent{\sqrt{\frac{1}{nb}} + \sqrt{\frac{1}{nb^2\varepsilon^2}}}.
    \end{align*}

    We call the three terms on the right hand side above $T_1, T_2$ and $T_3$. Finally, to balance $T_1, T_3$ and $T_2, T_3$, we perform the following case distinction: 
    \begin{enumerate}
        \item[] \textbf{Case 1}: If $T_2 \gtrsim T_3$, i.e., when $\varepsilon \gtrsim 1/\sqrt{b}$ we balance $T_1$ and $T_2$ as follows: 
        \begin{align*}
            T_1 \asymp T_2 
            \quad \Leftrightarrow \quad 
            b \asymp \frac{\sigma_{\max}}{\sqrt{nb}}
            \quad \Leftrightarrow \quad 
            b \asymp \Parent{\frac{\sigma^2_{\max}}{n}}^{1/3}
            \quad \Rightarrow \quad 
            \varepsilon \gtrsim \Parent{\frac{n}{\sigma^2_{\max}}}^{1/6}.
        \end{align*}

        \item[] \textbf{Case 2}: If $T_3 \gtrsim T_2$, i.e., when $\varepsilon \lesssim 1/\sqrt{b}$ we balance $T_1$ and $T_3$ as follows: 
        \begin{align*}
            T_1 \asymp T_3
            \quad \Leftrightarrow \quad
            b \asymp \frac{\sigma_{\max}}{\sqrt{nb^2\varepsilon^2}}
            \quad \Leftrightarrow \quad 
            b \asymp \sqrt{\frac{\sigma_{\max}}{\sqrt n \varepsilon}}
            \quad \Rightarrow \quad
            \varepsilon \lesssim \Parent{\frac{n}{\sigma^2_{\max}}}^{1/6}. 
        \end{align*}
    \end{enumerate}

    We therefore know that $T_2$ dominates when $\varepsilon \lesssim \varepsilon_c := (n/\sigma^2_{\max})^{1/6}$ and otherwise $T_3$ does. Choosing $b \asymp (\sigma^2_{\max}/n)^{1/3} \vee (\sigma_{\max}/(\sqrt n \varepsilon))^{1/2}$ incorporates both cases. We then recover the statement by plugging in $b$ and bounding the resulting maximum by a sum of both terms.
\end{proof}

\begin{proof}[Proof of Corollary \ref{cor:rMSEuserlvldeslocal}] \hfill 

    The proof is analogous to that of Corollary \ref{cor:rMSEuserlvldeslocal}. As in the central model, Assumption \ref{ass:logsobdep} holds and we also assume $\infnorm{\mu} \leq \mathsf B$. We then combine Lemma \ref{lem:concuserlvlmeanest} with Theorem \ref{thm:rMSEwinsmeanHDlocal} instead of Theorem \ref{thm:rMSEwinsmeanHD}. The statement follows by plugging $\tau \asymp \sqrt{\rho \log(2dn/\gamma)/T}$ into the bound on the probability.
\end{proof}

\begin{proof}[Proof of Corollary \ref{cor:rMSErandeffectslocal}] \hfill 

    The result is a direct consequence of Corollary \ref{cor:rMSEuserlvldeslocal} enabled by Lemma \ref{lem:randeffectslogsobdes}. 
\end{proof}

\begin{proof}[Proof of Corollary \ref{cor:MSEuserlvlreglocal}] \hfill 

    The proof is analogous to that of Corollary \ref{cor:MSEuserlvlreg}. Because of the block structure of $\Sigma^n$ and since $\infnorm{\beta} \leq \mathsf B$ by assumption, we use Theorem \ref{thm:rMSEwinsmeanHDlocal} under Assumption \ref{ass:iiddata} instead of Theorem \ref{thm:rMSEwinsmeanHD}. 
\end{proof}

\section{Variance Estimation Algorithms}
\label{app:plugin}

Lastly, we present the plug-in variance estimators used in Subsection \ref{subsubsec:pluginvar}. They both take $\sigma_{\min}^2, \sigma_{\max}^2$ and $\hat m$ as input, i.e., a lower and upper bounds on the variance as well as a potentially very rough estimate for the first moment. The algorithms iteratively update the upper bound and the first moment estimate, and eventually output an estimate of the variance. 

\subsection{Variance Estimation by Bisection}

Our first plug-in variance estimator repeatedly halves the provided upper bound on the variance as long as the coverage of the resulting confidence interval has high enough empirical coverage.

\begin{algorithm}[H]
    \caption{VarianceBisection$(X^n, \varepsilon, \delta, \varrho, \sigma_{\min}^2, \sigma_{\max}^2, \hat m, \gamma)$:}
    \label{alg:bisection}
    \begin{algorithmic}[1]
        \REQUIRE $X^n \in \R$, $\varepsilon, \delta, \varrho > 0$, $\sigma_{\min}^2 \sigma_{\max}^2 \in (0,\infty)$, $\hat m \in \R$, $\gamma > 0$    
        \STATE $N_{\mathsf{iter}} \gets \lceil\log_2(\sigma_{\max}^2/\sigma_{\min}^2)\rceil$
        \STATE $\varepsilon^\prime \gets \varepsilon/\sqrt{16N_{\mathsf{iter}}\log(1/\varrho)}$
        \WHILE{$\operatorname{Coverage}(X^n, \varepsilon^\prime, \sigma_{\max}, \hat m) \geq 1-\gamma$}
            \STATE $\hat m \gets \operatorname{RefineMidpoint}(X^n, \varepsilon^\prime, \sigma_{\max}, \hat m)$
            \STATE $\sigma_{\max}^2 \gets \sigma_{\max}^2/2$
        \ENDWHILE
        \RETURN $\sigma_{\max}^2$
    \end{algorithmic}
\end{algorithm}

The intervals coverage is computed privately in Algorithm \ref{alg:coverage} using a Laplace mechanism and will by design tend to give a conservative large variance estimate. Algorithm \ref{alg:firstmoment} is used to refine the mean estimator using the current variance upper bounds and privacy is ensured by the Laplace mechanism. 

\begin{algorithm}[H]
    \caption{Coverage$(X^n, \varepsilon, \sigma, \hat m)$:}
    \label{alg:coverage}
    \begin{algorithmic}[1]
        \REQUIRE $X^n \in \R$, $\varepsilon > 0$, $\sigma \in (0,\infty)$
        \STATE $\hat \tau \gets \sqrt{2}\sigma$
        \STATE $\hat I \gets \bracket{\hat m \pm \hat \tau}$
        \STATE $\hat p \gets \frac 1 n \sum_{i=1}^n \indicator{X_i \in \hat I}$
        \RETURN $\hat p + \xi$ with $\xi \sim \operatorname{Lap}(0, \frac{2}{n\varepsilon})$
    \end{algorithmic}
\end{algorithm}

\begin{algorithm}[H]
    \caption{RefineMidpoint$(X^n, \varepsilon, \sigma, \hat m)$:}
    \label{alg:firstmoment}
    \begin{algorithmic}[1]
        \REQUIRE $X^n \in \R$, $\varepsilon > 0$, $\sigma \in (0,\infty)$, $\hat m \in \R$
        \STATE $\hat \tau \gets \sqrt{2}\sigma$
        \STATE $\hat I \gets \bracket{\hat m \pm \hat \tau}$
        \RETURN $\frac{1}{n} \sum_{i=1}^n \proj{\hat I}{X_i} + \xi$ with $\xi \sim \operatorname{Lap}(0, \frac{2\tau}{n\varepsilon})$
    \end{algorithmic}
\end{algorithm}

\subsection{Variance Estimation via CoinPress}

The second plug-in variance estimator we employ is an adjusted version of Algorithms 4 and 8 (CoinPress) by \citet{biswas:dong:kamath:ullman2020}. Their original implementation assumes a zero mean and argues that under \iid{} samples non-zero means this is not restrictive since one can always work with an \iid{} sample of size $n/2$ symmetrized observations $Y_i := X_{i+1} - X_i$ and just double the variance. Under dependence, however, differencing observations biases the variance estimate. Algorithm \ref{alg:coinpress} avoids this issue by centering observations using the current estimate of the first moment $\hat m$.

\begin{algorithm}[H]
    \caption{AdaptiveCoinPress$(X^n, \varepsilon, \delta, \varrho, \sigma_{\min}^2, \sigma_{\max}^2, \hat m, \gamma)$:}
    \label{alg:coinpress}
    \begin{algorithmic}[1]
        \REQUIRE $X^n \in \R$, $\varepsilon, \delta, \varrho > 0$, $\sigma_{\min}^2 \sigma_{\max}^2 \in (0,\infty)$, $\hat m \in \R$, $\gamma > 0$ 
        \STATE $N_{\mathsf{iter}} \gets \lceil\log_2(\sigma_{\max}^2/\sigma_{\min}^2)\rceil$
        \STATE $\varepsilon^\prime \gets \varepsilon/\sqrt{16N_{\mathsf{iter}}\log(1/\varrho)}$
        \STATE $\sigma \gets \sigma_{\max}$
        \FOR{$t \in [N_{iter}]$}
            \STATE $\hat m \gets \operatorname{RefineMidpoint}(X^n, \varepsilon^\prime, \sigma_{\max}, \hat m)$
            \STATE $(\sigma, \sigma_{\max}^2) \gets \operatorname{RefineVariance}(X^n, \varepsilon^\prime, \sigma, \hat m)$
        \ENDFOR
        \RETURN $\sigma_{\max}^2$
    \end{algorithmic}
\end{algorithm}

The variance estimation \`a la CoinPress is then performed on centered observations in Algorithm \ref{alg:coinpressiter}. The algorithm implements a one-dimensional version of \citet[Algorithm 3]{biswas:dong:kamath:ullman2020}. 

\begin{algorithm}[H]
    \caption{RefineVariance$(X^n, \varepsilon, \sigma, \hat m, \gamma)$:}
    \label{alg:coinpressiter}
    \begin{algorithmic}[1]
        \REQUIRE $X^n \in \R$, $\varepsilon > 0$, $\sigma \in (0,\infty)$, $\hat m \in \R$, $\gamma > 0$ 
        \STATE $\beta \gets \parent{1+2\sqrt{\log(1/\gamma)} + 2 \log(1/\gamma)}^{1/2}$
        \STATE $w \gets \proj{[\pm\beta]^n}{(X^n - \hat m \cdot \one_n)/\sigma}$
        \STATE $z \gets 0 \vee (\frac 1 n \twonorm{w}^2 + \xi)$ with $\xi \sim \operatorname{Lap}(0, \frac{2\beta^2}{n\varepsilon})$
        \STATE $\sigma \gets \sigma \cdot \parent{z + \sqrt{1/n} + 1/(2n)}^{1/2}$
        \RETURN $(\sigma, z \cdot \sigma^2)$
    \end{algorithmic}
\end{algorithm}

\bibliography{bibliography}

\end{document}